\documentclass[conference,compsoc]{IEEEtran}

\ifCLASSOPTIONcompsoc
  \usepackage[nocompress]{cite}
\else
  \usepackage{cite}
\fi

\ifCLASSINFOpdf
\else
\fi

\usepackage[hidelinks]{hyperref}
\usepackage{microtype}
\usepackage{graphicx}
\usepackage{subcaption}
\usepackage{booktabs} %

\usepackage[utf8]{inputenc} %
\usepackage[T1]{fontenc}    %
\usepackage{url}            %
\usepackage{booktabs}       %
\usepackage{amsfonts}       %
\usepackage{nicefrac}       %
\usepackage{wrapfig}
\usepackage{xcolor}         %
\usepackage{cuted} %
\usepackage{pifont}
\usepackage{afterpage}

\newcommand*\colourcheck[1]{%
  \expandafter\newcommand\csname #1check\endcsname{\textcolor{#1}{\ding{52}}}%
}

\usepackage{tikz}

\usetikzlibrary{shapes.misc, positioning}
\usetikzlibrary{positioning, backgrounds, fit, calc}

\usepackage{siunitx}

\newcommand{\splitcell}[1]{%
  \begin{tabular}{@{}c@{}}\strut#1\strut\end{tabular}%
}

\usepackage{amsmath,amsfonts,bm}

\def\eqref#1{equation~\ref{#1}}

\def\1{\bm{1}}

\def\va{{\bm{a}}}
\def\vb{{\bm{b}}}

\def\vx{{\bm{x}}}

\def\vz{{\bm{z}}}

\DeclareMathAlphabet{\mathsfit}{\encodingdefault}{\sfdefault}{m}{sl}
\SetMathAlphabet{\mathsfit}{bold}{\encodingdefault}{\sfdefault}{bx}{n}

\def\sR{{\mathbb{R}}}

\newcommand{\E}{\mathbb{E}}

\newcommand{\R}{\mathbb{R}}

\DeclareMathOperator*{\argmin}{arg\,min}

\usepackage{bbm}

\usepackage{listings}
\usepackage{stackengine}
\usepackage{scalerel}

\usetikzlibrary{matrix}
\usetikzlibrary{positioning}
\usepackage{tikzsymbols}
\usepackage[normalem]{ulem}
\usetikzlibrary{arrows.meta}
\usetikzlibrary{math} %

\usepackage{amsmath}
\usepackage{mathtools}
\usepackage{amsthm}

\usepackage[capitalize,noabbrev]{cleveref}

\usepackage{letltxmacro}
\LetLtxMacro\oldttfamily\ttfamily
\DeclareRobustCommand{\ttfamily}{\oldttfamily\csname ttsize\endcsname}
\newcommand{\setttsize}[1]{\def\ttsize{#1}}%

\usepackage{booktabs}
\newcommand{\ras}[1]{\renewcommand{\arraystretch}{#1}}

\newcommand{\citet}[1]{\cite{#1}} 
\newcommand{\citep}[1]{\cite{#1}} 

\newcommand{\eg}{e.g., }
\newcommand{\ie}{i.e., }

\usepackage{makecell}
\usepackage{twoopt}

\makeatletter
\renewcommand{\paragraph}{\textbf} 
    \def\@IEEEsectpunct{.\ }

    \def\para{\@startsection%
        {paragraph}%
        {4}%
        {0\parindent}%
        {0.6ex plus 0.1ex minus 0.1ex}%
        {0ex}%
        {\normalfont\normalsize\itshape\bfseries}%
        *%
    }%
\makeatother

\newcommand\distOrig{\ensuremath{\mathcal{D}_{\text{orig}}}}
\newcommand\distMin{\ensuremath{\mathcal{D}_{\text{min}}}}

\newcommand\setOrig{\ensuremath{{S}_{\text{orig}}}}
\newcommand\setMin{\ensuremath{{S}_{\text{min}}}}
\newcommand\advOrig{\ensuremath{{S'}_{\text{orig}}}}
\newcommand\advMin{\ensuremath{{S'}_{\text{min}}}}

\newcommand\tool{Privacy-aware Tree}
\newcommand\toolshort{PAT}

\definecolor{testcolor}{HTML}{6a00ff}
\definecolor{testcolor2}{HTML}{6affff}

\newcommand{\bsize}{0.15}
\newcommand{\boxyOrig}{%
  \begin{tikzpicture}[every node/.style={inner sep=0,outer sep=0}]
    \protect\draw[fill={rgb,255:red,250;green,0;blue,0}] (0,0) rectangle ++(\bsize, \bsize);
    \protect\draw[fill={rgb,255:red,250;green,103;blue,0}] (1*\bsize,0) rectangle ++(\bsize, \bsize);
    \protect\draw[fill={rgb,255:red,250;green,163;blue,0}] (2*\bsize,0) rectangle ++(\bsize, \bsize);
    \protect\draw[fill={rgb,255:red,250;green,195;blue,0}] (3*\bsize,0) rectangle ++(\bsize, \bsize);
    \protect\draw[fill={rgb,255:red,250;green,222;blue,0}] (4*\bsize,0) rectangle ++(\bsize, \bsize);
    \protect\draw[fill={rgb,255:red,250;green,250;blue,0}] (5*\bsize,0) rectangle ++(\bsize, \bsize);
  \end{tikzpicture}%
}

\newcommand{\boxyMin}{%
  \begin{tikzpicture}
    \protect\draw[fill={rgb,255:red,223;green,175;blue,160}] (0,0) rectangle ++(\bsize, \bsize);
    \protect\draw[fill={rgb,255:red,243;green,225;blue,191}] (\bsize,0) rectangle ++(\bsize, \bsize);
  \end{tikzpicture}%
}

\usetikzlibrary{shapes.geometric, calc}

\newcommand\score[2]{%
  \pgfmathsetmacro\pgfxa{#1 + 1}%
  \tikzstyle{scorestars}=[star, star points=5, star point ratio=2.25, draw, inner sep=0.15em, anchor=outer point 3]%
  \begin{tikzpicture}[baseline]
    \foreach \i in {1, ..., #2} {
      \pgfmathparse{\i<=#1 ? "yellow" : "gray"}
      \edef\starcolor{\pgfmathresult}
      \draw (\i*1em, 0) node[name=star\i, scorestars, fill=\starcolor]  {};
    }
    \pgfmathparse{#1>int(#1) ? int(#1+1) : 0}
    \let\partstar=\pgfmathresult
    \ifnum\partstar>0
      \pgfmathsetmacro\starpart{#1-(int(#1)}
      \path [clip] ($(star\partstar.outer point 3)!(star\partstar.outer point 2)!(star\partstar.outer point 4)$) rectangle 
      ($(star\partstar.outer point 2 |- star\partstar.outer point 1)!\starpart!(star\partstar.outer point 1 -| star\partstar.outer point 5)$);
      \fill (\partstar*1em, 0) node[scorestars, fill=yellow]  {};
    \fi
  \end{tikzpicture}%
}

\definecolor{genout}{RGB}{100, 176, 119}
\definecolor{genin}{RGB}{148, 227, 168}

\definecolor{bgreen}{RGB}{116, 186, 89}
\definecolor{oranyellow}{RGB}{245, 229, 113}
\definecolor{bred}{RGB}{222, 72, 42}

\newcommand*\emptycirc[1][1.25ex]{\tikz\draw[color=bred, thick] (0,0) circle (#1);} 
  
\newcommandtwoopt*\quartcirc[2][1.25ex][bred]{%
  \begin{tikzpicture}
  \draw[fill=#2, color=#2] (0,0) -- (#1,0) arc[start angle=0, end angle=360,radius=#1] -- (0,0);
  \draw[fill=#2, color=white] (0,0) -- (0,#1) arc[start angle=90, delta angle=270, radius=#1] -- (0,0);
  \draw[color=#2] (0,0) circle (#1);
\end{tikzpicture}}

\newcommandtwoopt*\halfcirc[2][1.25ex][oranyellow]{%
  \begin{tikzpicture}
  \draw[fill=#2, color=#2] (0,0) -- (#1,0) arc[start angle=0, end angle=360,radius=#1] -- (0,0);
  \draw[fill=#2, color=white] (0,0) -- (0,#1) arc[start angle=90, delta angle=180, radius=#1] -- (0,0);
  \draw[color=#2] (0,0) circle (#1);
  \end{tikzpicture}}

\newcommandtwoopt*\threequarcirc[2][1.25ex][bgreen]{%
  \begin{tikzpicture}
  \draw[fill=#2, color=#2] (0,0) -- (#1,0) arc[start angle=0, end angle=360,radius=#1] -- (0,0);
  \draw[fill=#2, color=white] (0,0) -- (0,#1) arc[start angle=90, delta angle=90, radius=#1] -- (0,0);
  \draw[color=#2] (0,0) circle (#1);
  \end{tikzpicture}}
  
\newcommand*\fullcirc[1][1.25ex]{\tikz\fill[color=genout] (0,0) circle (#1);}

\begin{document}
\title{From Principle to Practice:\\ Vertical Data Minimization for Machine Learning}

\setttsize{\small}%

\author{\IEEEauthorblockN{Robin Staab} 
\IEEEauthorblockA{ETH Zurich, Switzerland \\ robin.staab@inf.ethz.ch}
\and
\IEEEauthorblockN{Nikola Jovanović}
\IEEEauthorblockA{ETH Zurich, Switzerland \\ nikola.jovanovic@inf.ethz.ch}
\and
\IEEEauthorblockN{Mislav Balunović}
\IEEEauthorblockA{ETH Zurich, Switzerland \\ mislav.balunovic@inf.ethz.ch}
\and
\IEEEauthorblockN{Martin Vechev}
\IEEEauthorblockA{ETH Zurich, Switzerland \\ martin.vechev@inf.ethz.ch}}

\maketitle
\thispagestyle{plain}
\pagestyle{plain}

\captionsetup[figure]{labelfont=bf,textfont=rm}
\captionsetup[table]{labelfont=bf,textfont=rm}

\begin{abstract}  
  Aiming to train and deploy predictive models, organizations collect large amounts of detailed client data, risking the exposure of private information in the event of a breach. 
To mitigate this, policymakers increasingly demand compliance with the \emph{data minimization} (DM) principle, restricting data collection to only that data which is relevant and necessary for the task.
Despite regulatory pressure, the problem of deploying machine learning models that obey DM has so far received little attention.
In this work, we address this challenge in a comprehensive manner.
We propose a novel \emph{vertical DM} (vDM) workflow based on data generalization, which by design ensures that \emph{no full-resolution client data is collected} during training and deployment of models, benefiting client privacy by reducing the attack surface in case of a breach.
We formalize and study the corresponding problem of finding generalizations that both maximize data utility and minimize empirical privacy risk, which we quantify by introducing a diverse set of policy-aligned adversarial scenarios.
Finally, we propose a range of baseline vDM algorithms, as well as \emph{Privacy-aware Tree} (\emph{PAT}), an especially effective vDM algorithm that outperforms all baselines across several settings.
We plan to release our code as a publicly available library, helping advance the standardization of DM for machine learning.
Overall, we believe our work can help lay the foundation for further exploration and adoption of DM principles in real-world applications. 
 
\end{abstract}

\section{Introduction} \label{sec:intro}
Advances in machine learning (ML) have enabled organizations to automate tasks such as credit risk scoring~\cite{khandani2010consumer} or fraud detection~\cite{perols2011financial}. As ML models require large amounts of training samples, organizations increasingly collect different types of detailed client data, hoping to improve the models' performance. 
The deployment of such models, in turn, necessitates the collection of an even larger amount of highly-detailed client data for inference.
These developments have led to growing regulatory concerns regarding the effects of large-scale data collection on individuals' privacy.
 
\begin{figure}[t]
  \centering
  \resizebox{0.48\textwidth}{!}{
        \centering
  \includegraphics{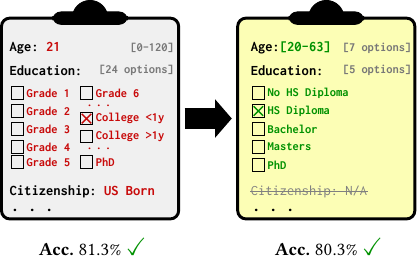}}
  \caption{Vertical data minimization (vDM) can greatly reduce the granularity of the data being collected, while not significantly impacting downstream ML models. We give a full overview of the chosen example in \cref{ssec:experiments:qualitative}.}
  \label{fig:bait} 
  \vspace{-0.5em}
\end{figure}

\subsection{Data Minimization}
In an attempt to address this issue, several authorities have developed regulations limiting data collection and processing. Most notably, such concerns are an important part of EU's General Data Protection Regulation (GDPR)~\citep{gdpr}, California's Privacy Rights Act (CPRA)~\citep{cpra}, and the recent Blueprint for a U.S. AI Bill of Rights~\cite{whitehouse2022bill}. The GDPR, for example, defines \emph{data minimization} (DM) in Article 5C as the principle of only collecting and using data that is \emph{``adequate, relevant and limited to what is necessary in relation to the purposes for which it is processed''}. Similarly, the AI Bill of Rights Blueprint dictates \emph{``ensuring that data collection conforms to reasonable expectations and that only data strictly necessary for the specific context is collected.''}
In the context of ML, this implies that the collection of any personal data (\eg citizenship) has to be justified by the increase in utility of the resulting model. Furthermore, these regulations also apply for \emph{model deployment}, a setting that many prior approaches cannot feasibly handle  (\cref{sec:setting}).

\para{DM in Practice}
DM regulations have already had real-world impact---the EU has so far issued at least 166 fines due to DM violations~\cite{gdprtracker}. An example is a fine of 2.75M euros issued in a case of fraud detection for childcare benefits~\cite{dutchgdpr}, where the tax agency collected applicants' citizenship as a feature for their fraud detection model, even though it has later been shown that a simpler feature (being a resident or not) would have sufficed to achieve the same utility. 
Similarly, the EU has fined a delivery company~\cite{gdpr790} for recording the GPS position of drivers every $12s$, despite less detailed information being sufficient for their purpose. 
Finally, there are various examples of fines issued when a security breach exposed sensitive data~\citep{gdprtracker}, which could have been avoided if such data was not unnecessarily collected to begin with.

On a positive note, there are examples of successful DM applications, \eg the Norwegian Tax Authority reported using only 30 out of 500 considered variables in their tax error detection system~\cite{norway2018ai}.
Our experiments in~\cref{sec:experimental} further indicate the feasibility of DM approaches in practice---we illustrate one such example in \cref{fig:bait}, where collecting only the age group (e.g., 20-63) and the highest obtained diploma (e.g., High School) instead of exact age, educational status, and citizenship, results in minimal utility loss.

\subsection{Principled Vertical DM for ML} \label{ssec:intro:vDM}
Reducing the \emph{number of data points} collected is an important and well-studied research problem~\cite{shanmu,mirzasoleiman,paul}, which can be interpreted as \emph{horizontal data minimization} (hDM). 
We argue that hDM is not a suitable solution to many of the privacy concerns behind DM, as it offers \emph{no privacy protection} for individual clients whose data was collected.
In contrast, we therefore focus on \emph{vertical data minimization} (vDM, formalized in \cref{sec:problem}), as the process of reducing the information collected \emph{within each data point}, which directly protects clients that provide data during model training and especially deployment.

Despite these advantages of vDM, there has so far not been a well-established framework for the principled application and evaluation of vDM for ML.
Without standardized evaluation, practitioners lack insight into how well their DM solutions protect client privacy in real-world settings.
For example, as we will elaborate on in~\cref{ssec:background:generalization}, one prior attempt to formalize vDM uses metrics that fail to capture the vulnerability to practical privacy attacks. 
This raises two key questions: 
\begin{enumerate}
    \setlength\itemsep{1em}
    \item \emph{What are the unique privacy challenges posed by DM regulations and why are existing approaches insufficient for addressing them?}
    \item \emph{Can we devise an approach for principled application and evaluation of vDM that addresses these concerns for practical machine learning use-cases?}
\end{enumerate}

\subsection{This Work}
In this work, we tackle these questions, taking several steps to lay the foundations of vDM for ML.

\para{The vDM Setting}
First, we differentiate the vDM setting from similar settings highlighting how they are insufficient for addressing the requirements set forth by DM regulations. Afterwards we formalize the vDM setting via the concept of \emph{generalization}, \ie replacing detailed attributes with less granular ones (\eg age with an age group).
We define a workflow that organizations can employ, which after choosing a generalization based on a small set of full-granularity data points, ensures that during future model training and deployment \emph{no full-granularity data is collected from the clients} (\eg in a data collection survey, as illustrated in~\cref{fig:bait}), reducing the attack surface in case of a data breach and aligning with regulatory requirements.

To answer the question of which generalizations are suitable, we define two key objectives: \emph{data utility}, which measures how useful the minimized data is for the downstream ML task (by training a classifier), and \emph{empirical privacy risk}, which measures the potential to compromise individual privacy by observing minimized data (\eg after a data breach).
To quantify the latter, we formalize a comprehensive set of diverse and policy-aligned (see \cref{sec:metrics}) \emph{adversaries} with different attack objectives and capabilities, \eg regarding their side information. 

\para{The vDM Algorithms}
We introduce a range of strong baseline vDM algorithms, as well as \emph{\tool} (\toolshort), a vDM algorithm inspired by prior work on tree-based fair encoders \cite{jovanovic2023fare}.
We perform an extensive experimental evaluation of PAT and our baselines in various settings, demonstrating that PAT generally achieves the most favorable utility-privacy tradeoffs compared to all baselines.
Our experimental findings highlight the importance of a principled evaluation of empirical privacy risks by illustrating how a naive evaluation would fail to capture key aspects of vDM.

\subsection{Key Contributions} Our main contributions are:

\begin{itemize}
  \item A formalization of the \emph{vertical data minimization (vDM)} setting, and the underlying problem of generalizing the data such that it remains useful for the downstream task (\emph{utility}) and exhibits low \emph{empirical privacy risk} (\cref{sec:problem}).
  \item A formulation and instantiation of a comprehensive set of adversaries with different attack capabilities as a tool for evaluating the empirical privacy risk of vDM generalizations (\cref{sec:metrics}).
  \item A diverse set of baseline vDM algorithms that can serve as a benchmark for future work (\cref{sec:baselines}).
  \item A novel vDM algorithm, \emph{\tool} (\toolshort), achieving state-of-the-art results across multiple datasets (\cref{sec:algorithm}).
  \item An extensive experimental evaluation of all introduced vDM algorithms and adversaries on several real-world datasets highlighting the practical applicability of vDM (\cref{sec:experimental}).
  \item A library with all our adversaries, baselines, and \toolshort~, advancing the standardization of vDM available at \href{https://github.com/eth-sri/datamin}{https://github.com/eth-sri/datamin}.
\end{itemize}

\section{Background} \label{sec:background}

\begin{figure*}
    \centering
    \resizebox{0.85\textwidth}{!}{
        \centering
        \input{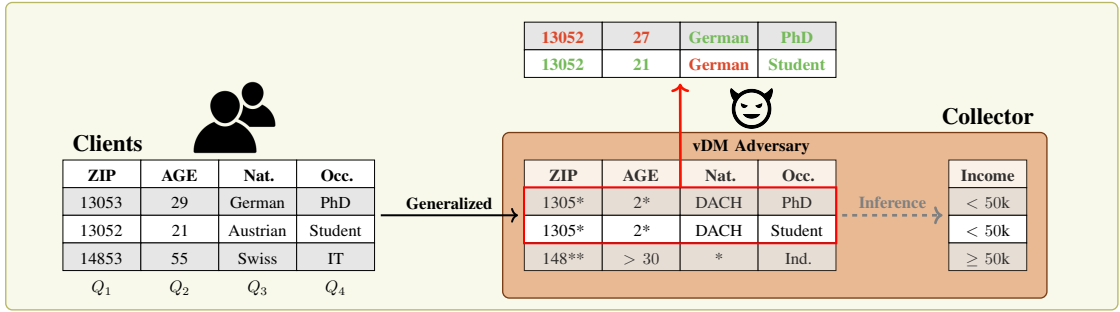}}
    \caption{Overview of the vDM deployment setting on an ACSIncome toy dataset (see \cref{sec:experimental}). Clients directly provide generalized data, \ie adversary and collector (which could be the same) only observe generalized records. The vDM adversary tries to reconstruct the original values of the generalized attributes (here all $Q_i$). The collector runs a downstream ML model on the generalized data for inference.}
    \label{fig:dmppdp}
\vspace{-0.5em}
\end{figure*}

In this section, we will introduce the background necessary for the subsequent parts of the paper.

\subsection{Machine Learning}
In recent years ML algorithms, specifically deep neural networks, have been applied in classification tasks in a wide variety of settings \cite{grigorescu2020survey, chen2018rise}. Let $\mathcal{D}$ be a data distribution over $ \mathcal{X} \times \mathcal{Y}$ and $(x,y) \sim \mathcal{D}$ be a data point consisting of a \emph{record} $x$ and its discrete label $y \in \mathcal{Y} = \{1,\ldots,c\}$.
The goal of an ML algorithm is to learn a mapping function $f_{\theta}\colon X \rightarrow Y$ (\eg a neural network) parameterized by $\theta$ (\eg neural network weights), which maps records $x \in X$ to $f_{\theta}(x) \in Y$ aiming to minimize some objective function $\mathcal{L}$. We will refer to $f = f_{\theta}$ as a \emph{model}.
For classification, we set \mbox{$\mathcal{L} = \mathbb{E}_{(x,y)\sim \mathcal{D}}[f(x) \neq y]$}, the expected rate of misclassification, which in training is approximated using finite samples $S_{train} \sim \mathcal{D}$ (\emph{training set}). We assume that $f$ is trained on $S_{train}$ and evaluated on a distinct \emph{test set} $S_{test} \sim \mathcal{D}$.

\subsection{Attribute Generalization and NCP} \label{ssec:background:generalization}

In ML the term \emph{generalization} commonly refers to \emph{model} generalization, \ie the model's performance on unseen data. For the scope of this work, we define \emph{attribute} generalization (hereafter referred to as \emph{generalization}) as a function $g\colon X \rightarrow Z$ on our data $X \subseteq \mathbb{R}^d$. This is referred to as a \emph{global} generalization as it fixes a single $g$ for all data points. \emph{Local} generalizations, on the other hand, allow different generalization functions for separate data points, enabling two data points with identical values to map to different generalized points.
As in \cite{lefevre2006mondrian} we say that $g$ is \emph{single-dimensional} when it can be decomposed into a set of $g_i$ such that $g(x) = (g_1(x_1),\ldots,g_d(x_d))$ and \emph{multidimensional} otherwise.
Finally, generalizations can be either \emph{strict}, ensuring that the image of each $g_i$ forms a proper partition of the respective attribute domain, or \emph{relaxed}, allowing elements of $g_i$'s image to have non-empty intersections (\eg \texttt{age} both to ranges 20-30 and 25-32). 
\para{NCP} We recall the definition of the \emph{normalized certainty penalty (NCP)}, a metric commonly used in data anonymization settings to quantify the information loss of a generalization~\cite{xu2006utility,Ghinita}.
For brevity, we focus only on categorical attributes.
Assume a generalization $g$ defined on $d$-dimensional data, and an attribute $i$ with domain $D_i$. For a generalized record $g(\vx) = \vz$ we set $NCP_i(\vz)=0$ if $\vz_i = \vx_i$, and $NCP_i(\vz) = |\vz_i| / |D_i|$ otherwise, where we define $|\vz_i|$ as $|g_i^{-1}(\vz_i)|$ (the size of the pre-image of $\vz_i$).
The per-attribute $NCP_i(\vz)$ is combined using pre-selected attribute weights $w_i$ to obtain the $NCP(\vz) = \sum_{i=0}^{d-1} w_i \cdot NCP_i(\vz)$. Given a dataset with $n$ points this is combined to a (normalized) Global Certainty Penalty $GCP = \frac{1}{n} \sum_{i=1}^{n} NCP(\vz^{(i)})$ \cite{IbmApt}. 

The only prior work~\cite{IbmApt} attempting to formalize vDM relied on NCP-based metrics to quantify the privacy risk of a generalization.
We argue, however, that NCP is a generic information loss metric incapable of accurately reflecting the adversarial vulnerability of the data when vDM is applied in real-life scenarios.
Assume, \eg an attribute $a$ with possible values $\{a_1,a_2,a_3,a_4\}$ used for medical diagnosis, where $a\in \{a_1,a_2\}$ implies that the patient requires medication, while $a\in \{a_3,a_4\}$ implies otherwise.
For an adversary with knowledge of the generalization, generalizing $a$ such that $\{a_1, a_2\} \to g_1$ and $\{a_3, a_4\} \to g_2$ (Gen. 1) reveals whether a patient needs medication. Generalizing $\{a_1, a_4\} \to g_1$ and $\{a_2, a_3\} \to g_2$ (Gen. 2) does not leak this information. 
Despite this, both generalizations have a GCP score of $0.5$, making it impossible to distinguish between them.
Motivated by this, we advocate for a more realistic measure of privacy risk that directly quantifies an adversary's success at compromising client privacy in the vDM setting. Based on existing legislature~\cite{GDPR3Attacks}, we will formalize a comprehensive set of relevant adversaries in~\cref{sec:metrics}. 

\subsection{Personal \& Sensitive Data} \label{ssec:background:data}

\para{Legal Definitions}
In the EU, the introduction of the GDPR has led to the establishment of a clear definition of \emph{personal} data. According to Article 4 of the GDPR ~\cite{gdpr}, personal data is defined as \emph{``any information relating to an identified or identifiable natural person ('data subject').''} 
This definition is more rigorous than the Personal Identifiable Information (PII) definitions commonly employed under U.S. jurisdiction. The U.S. Department of Labor defines PII as \emph{``any representation of information that permits the identity of an individual to\ldots be reasonably inferred''} \cite{piius}.
Both in GDPR and PII, there is a concept of (especially) \emph{sensitive} data. The GDPR details this in Article 9 as a set of sensitive (special category) personal data (\eg relating to race, sexual orientation, religion) for which special care needs to be taken.

\para{Technical Definitions} \label{ssec:background:technical}
As we elaborate on in \cref{sec:setting}, the well-established research areas of \emph{privacy-preserving data publishing (PPDP)}~\cite{fung2010privacy} and, within it, \emph{data anonymization (DA)}, operate with a narrower definition of sensitive data. 
The standard PPDP setting assumes that a \emph{data collector} wants to release a (fixed) table $T$ of data entries with attributes $(T_1, \ldots, T_d)$ ($T_i$ denoting the i-th attribute), out of which only \emph{one} (instead of all personal attributes) is sensitive. 

In particular, each attribute $T_i$ is \textit{at most one} of the following: \textit{Unique Identifier} $U_i$ which directly identifies a single entry, \textit{Quasi-Identifier} $Q =(Q_1,\ldots, Q_{d_q})$, which allows unique identification of at least one record in $T$ by examining the attributes in $Q$, or \textit{Sensitive} $S$ whose value we want to protect from being connected with any particular individual in $T$. 
Some PPDP works~\cite{han2013sloms,anjum2018efficient} attempt to relax the assumption of a single $S$. However, this usually happens at the cost of data utility \cite{anjum2018efficient} and often requires the data to be sliced into multiple tables, each containing only parts of the sensitive attributes.
This gap between stricter regulatory requirements (requiring the protection of all personal attributes) and existing work is one of the motivations for our vDM setting in \cref{sec:problem}.

As in PPDP/DA we will focus on tabular data. This both follows a long line of work in related areas \cite{jovanovic2023fare, balunovic2022fair} and also fits regulatory requirements, which are primarily focused on sensitive attributes in tabular format \cite{noauthor_prohibited_nodate, gdpr}. 

\afterpage{

\begin{table*}[t]

\centering
\ras{1.3}
\caption{Comparison of PETs to the vDM setting. PETs are rated in each category on a scale \{\emptycirc[0.75ex], \quartcirc[0.75ex], \halfcirc[0.75ex], \threequarcirc[0.75ex], \fullcirc[0.75ex]\} with more detail in \cref{ssec:PET}. We rate assumptions made on client capabilities and trust in the collector during collection/deployment from \emptycirc[0.75ex] (\emph{strong}) to \fullcirc[0.75ex] (\emph{none}). Further, we rate current technical feasibility and client privacy from \emptycirc[0.75ex] (\emph{none}) over \threequarcirc[0.75ex]  
 ~(\emph{empirical}) to \fullcirc[0.75ex] (\emph{guaranteed}). For inference, only vDM yields a feasible solution in an untrusted collector setting.}

\resizebox{0.92 \linewidth}{!}{
\begin{tabular}{@{}rccccccccccc@{}}\toprule
& \multicolumn{3}{c}{Collection} & \multicolumn{2}{c}{Training} & \multicolumn{4}{c}{Deployment}\\
\cmidrule(l{2pt}r{2pt}){2-4} \cmidrule(l{2pt}r{2pt}){5-6} \cmidrule(l{2pt}r{2pt}){7-10}
& \thead{Client\\Assumptions} & \thead{Trust in\\Collector} & \thead{Privacy \\ (Wire)} & \thead{Technical\\Feasibility} & \thead{Privacy \\ (Model)} & \thead{Client\\Assumptions} & \thead{Trust in\\Collector} & \thead{Technical\\Feasibility} & \thead{Privacy \\ (New Record)}\\ \midrule
Fed. Learning & \quartcirc & \halfcirc & \halfcirc & \halfcirc & \halfcirc &\quartcirc & - & \quartcirc & \fullcirc\\
DP (Central) & \fullcirc & \emptycirc & \emptycirc & \threequarcirc & \fullcirc & \fullcirc & \emptycirc & \fullcirc & \emptycirc\\
DP (Local) & \halfcirc & \fullcirc & \fullcirc & \quartcirc & \fullcirc & \halfcirc & - & - & - \\
E2E-Crypto & \halfcirc & \emptycirc & \fullcirc & \fullcirc & \quartcirc & \halfcirc & \emptycirc & \fullcirc & \emptycirc \\
FHE & \halfcirc & \fullcirc & \fullcirc & \quartcirc & \fullcirc & \halfcirc & \fullcirc & \quartcirc & \fullcirc \\
SMC  & \halfcirc & - & \fullcirc & \quartcirc & \fullcirc & \quartcirc & - & \quartcirc & \fullcirc \\
PPDP/DA & \fullcirc & \emptycirc & \emptycirc & - & - & - & - & - & -  \\
Synthetic Data & \fullcirc & \emptycirc & \emptycirc & \fullcirc & \fullcirc & \fullcirc & \emptycirc & \fullcirc & \emptycirc \\
\midrule
vDM & \fullcirc & \quartcirc & \halfcirc & \fullcirc & \halfcirc & \fullcirc & \fullcirc & \fullcirc & \threequarcirc \\
\bottomrule
\end{tabular}
}

\label{tab:comparison}
\vspace{-0.5em}
\end{table*} 
}

\section{Motivating a New vDM Setting} \label{sec:setting}

In this section, we motivate the need for a new vDM setting in two steps: In \cref{ssec:mot}, we derive concrete requirements for the vDM setting directly from regulations such as GDPR \cite{gdpr} and the U.S. AI Bill of Rights Blueprint \cite{whitehouse2022bill}. Using this, we show in \cref{ssec:PET} how current privacy-enhancing technologies (PETs) fail to address specific parts of these requirements (summarized in \cref{tab:comparison}), motivating our vDM formalization in \cref{sec:problem}.

\subsection{A Regulation-Guided vDM Setting} \label{ssec:mot}

As our vDM setting focuses on data minimization in ML, \emph{the specific purpose} of client data, as required by GDPR Article 5C, is to provide an accurate ML model for \emph{a specific} task (\eg for medical diagnosis) while protecting client privacy. To properly evaluate this via adversarial risk (as done in \cref{sec:metrics}), we first have to clarify assumptions made on the adversary and clients. %

\para{Protecting Data Collection}
The U.S. AI Bill of Rights dictates that \emph{``\textellipsis only data strictly necessary for the specific context is collected''}, putting the emphasis of the vDM setting on data collection not only for model training but also for model inference when (new) client data processing shall be ``limited to what is necessary'' (GDPR Article 5C). As depicted in~\cref{fig:dmppdp} on a toy example, this implies that the vDM adversary (formalized in \cref{sec:metrics}) is positioned between the client and the data collector or even is an honest-but-curious collector. 
The goal of the vDM adversary hereby is to reconstruct \emph{all} personal attribute values from an observed generalized record. This differs from the adversaries assumed in some other PETs (\cref{ssec:PET}). 

\para{Client Assumptions}
Inherent to this focus on data collection is the need to clarify client capabilities. To capture a wide range of use cases, we want to minimize the number of assumptions made about the client. In particular, the vDM setting does not assume any cryptographic capabilities or possibilities to interact with the collector---minimization should happen only on the data with a focus on the amount of data \emph{collected} from clients. This approach is orthogonal to many PETs that aim to protect full-resolution data collection (\cref{ssec:PET}).
To avoid assumptions on client capabilities in vDM, we will require our generalization $g$ to be (1) directly usable on the client side and (2) easily applicable to new data points for inference. 
Namely, clients should be able to enter values (\eg \texttt{age}) independently of other values (\ie $g$ is single-dimensional), and data entry should remain consistent across clients (\ie $g$ is global). Further, we must ensure that each original attribute value has exactly one corresponding generalized value (\ie $g$ is strict).

\subsection{Other Privacy-Enhancing Technologies}
\label{ssec:PET}

We now discuss other commonly applied PETs in ML, highlighting how they are unable to address the vDM setting outlined in \cref{ssec:mot}.
Throughout this section, we split a typical ML workflow into three stages: Initial data \emph{collection}, model \emph{training}, and \emph{deployment}, allowing for a more fine-grained analysis as shown in \cref{tab:comparison}. For data collection, we first distinguish whether the PET makes strong assumptions about client capabilities and whether clients assume the collector to be trusted. We then categorize the privacy of client records sent to the collector (\emph{Privacy (Wire)}). The collector uses this data during the training stage to train an ML model. For this, we rank both technical feasibility (\eg expensive computations, decreased performance) and the privacy of the resulting model against membership inference attacks (\emph{Privacy (Model)}) \cite{MIA,MIASOK}. With a trained model, we proceed to the deployment (inference) stage. Again we first categorize assumptions on (new) clients and respective trust in the collector. Additionally, we rate how technically feasible the PET is for inference on new records as well as the records' privacy protection (\emph{Privacy (New Record)}).

\para{Federated Learning}
Federated Learning aims to protect client privacy by letting them train models locally, only combining the resulting models/gradients on a server \cite{9084352}. This, however, comes with the strong technical requirement of clients being capable of training a model locally (denoted by \quartcirc[0.75ex]~in \cref{tab:comparison}) and also was shown to be vulnerable to gradient inversion attacks \cite{balunovic2021bayesian} leaking full resolution data.

\para{Differential Privacy}
Differential privacy (DP) \cite{dwork2006differential} has been widely recognized as the privacy standard in various data analytics applications due to its rigorous privacy assurances independent of the adversary's background knowledge.
We differentiate between local and central DP \cite{ye_local_2020}, both of which try to make it hard to determine whether data of a specific client is included in a training dataset (membership inference) while still allowing meaningful conclusions to be drawn from the aggregate data. 

In ML settings, central DP is commonly achieved by adding noise to gradients during training \cite{abadi2016deep}. 
vDM differs from central DP already in the setup, \ie by focusing on limiting the amount of personal data put into the system instead of the amount remaining in the model. 
In particular vDM is concerned with protecting privacy during deployment where centralized DP does not offer any privacy for new clients (\emptycirc[0.75ex]~for \emph{Privacy (New Record)} in \cref{tab:comparison}).

Local DP, on the other hand, requires clients to perturb their data locally before sending it to an (untrusted) collector. While closer to the vDM setting, it is infeasible for vDM for three reasons: (1) It requires active participation of clients for perturbation, (2) practical applications of local DP in ML are limited in scalability \cite{ye_local_2020}, and (3) it offers no privacy in an inference setting where clients want to receive results on their non-perturbed data (in \cref{tab:comparison} we therefore consider it non-applicable for deployment). 

\afterpage{
\begin{figure*}
  
  \centering
  
  \resizebox{0.77\textwidth}{!}{ 
    \centering
    \includegraphics{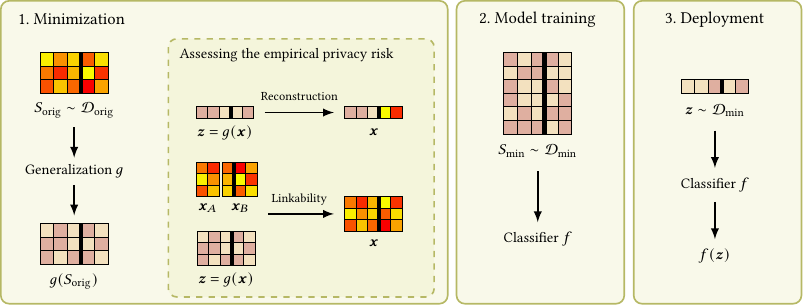}}

  \caption{Overview of our vertical data minimization (vDM) workflow. In the \emph{minimization} phase a minimizer proposes a generalization $g$ using a limited sample of full-granularity original data $\setOrig$, where each record (here, each row) consists of \emph{non-personal} (here, first 3) and \emph{personal} (here, last 2) attributes. The empirical privacy risk of $g$ is then assessed using a wide range of adversaries (here we show only two examples). In the \emph{model training} phase, we collect a large sample $\setMin$ of generalized data to train the classifier $f$, and use it for inference in the \emph{deployment} phase. We use saturated colors, \boxyOrig, to indicate full-granularity attributes, and desaturated colors, \eg \boxyMin, to indicate generalized attributes.}

  \label{fig:overview}

  \vspace{-0.5em}
  
\end{figure*} 
}

\para{Cryptographic Approaches}
For cryptography-based approaches, we differentiate between (1) Simple encryption between client and collector (E2E), (2) Fully-Homomorphic-Encryption (FHE) based schemes \cite{fhe_gentry} which allow ML on top of the encrypted data, and Secure Multi-party Computation (SMC). The first two approaches require clients to have a secure device capable of cryptographic capabilities (\halfcirc[0.75ex]~for \emph{Client Assumptions} in \cref{tab:comparison}). Furthermore, while E2E encryption protects data in transit, it offers no privacy guarantee in case of a curious collector. 

FHE, on the contrary, protects data during transit, training, and deployment. However, due to its heavy use of cryptographic primitives, it is limited in \emph{Technical Feasibility} (\quartcirc[0.75ex]~in \cref{tab:comparison}) \cite{chillotti_new_nodate} due to large overheads in memory and runtime \cite{hernandez_marcano_fully_2019, lee_privacy-preserving_2022} or requiring specific architectures \cite{lee_privacy-preserving_2022}.

SMC goes beyond this, avoiding a central collector by requiring active participation of (most) clients in the computation \cite{cramer2015secure}. While enabling strict privacy guarantees, this makes it infeasible for larger client sizes and many ML use-cases \cite{knott2021crypten}.
A recent line of work has combined SMC with versions of FHE for \emph{secure neural network inference} \cite{juvekar2018gazelle, mishra2020delphi}. While feasible for some scenarios, this approach still suffers from clients having to actively participate in the computation (\eg to evaluate activation layers \cite{juvekar2018gazelle}).

\para{PPDP/DA}
As mentioned in \cref{ssec:background:technical}, PPDP aims to protect individuals in a table $T$ from being connected to their sensitive attribute value by releasing only an anonymized data table $T'$ (alongside $S$). 
Records in $T'$ consist of generalized, suppressed, and perturbed quasi-identifiers of T, with the PPDP adversary aiming to map specific records in $T'$ to their respective sensitive attribute value. The degree of privacy protection through $T'$ is commonly formalized by ensuring one of the following: each record in $T'$ is indistinguishable from at least $k-1$ other records (k-anonymity~\cite{Sweene02}), the values for $S$ are well spread (l-diversity~\cite{machana}), or that the distribution of the sensitive attribute is close to the distribution over the entire dataset (t-closeness~\cite{li2006t}).
Note that these constraints, as they focus solely on re-identification risk, can often be significantly misaligned with our privacy notion of individual attribute protection.

Unlike vDM, PPDP does not consider inference, \ie data is released once, and the used generalizations are not required to be applicable to new data points as in vDM.
For this reason, most PPDP work focuses on non-strict, multidimensional~\cite{lefevre2006mondrian} generalizations, or relies on \emph{perturbing} and \emph{permuting} data in $T$~\cite{he2012permutation}.
While this broadens the space of solutions in the PPDP setting, it makes most algorithms directly non-applicable to the \emph{Deployment} setting in \cref{tab:comparison}.

\para{Synthetic Data Generation}
Recently, Synthetic Data Generation (SDG)~\cite{xu2018synthesizing,stadler2022synthetic} has gained popularity as another approach to address similar privacy concerns as PPDP. 
Unlike PPDP, where we generalize original data, SDG trains a model on the original dataset and uses it to generate entirely new data points. These synthetic data points resemble the original data but do not directly correspond to any individuals. In a deployment setting, SDG cannot protect the privacy of new clients sending their real data (\emptycirc[0.75ex] in \cref{tab:comparison}).

\section{Formalizing vDM for ML} \label{sec:problem}

Having established the key vDM requirements and highlighted shortcomings of existing PETs, we now formalize the vDM setting, describe the corresponding workflow, and discuss instantiations of utility and empirical privacy risk tailored to the context of ML.

\para{Overview}
Assume an organization aims to solve a prediction task by learning and deploying a classifier $f$ to predict labels $y \in \mathcal{Y}$ from records $x \in \mathcal{X} \subseteq \mathbb{R}^d$ with joint distribution $(\vx, y) \sim \distOrig$. 
To train $f$ well, the set of training pairs $(\vx, y)$ should be large and records $\vx$ as detailed as possible. At the same time, we want to protect the privacy of individuals.

To treat this tradeoff in a principled manner, we introduce the \emph{vDM workflow} depicted in \cref{fig:overview} and explained in detail below.
Importantly, after having chosen the generalization $g$ in the \emph{minimization} phase (transforming $\mathcal{D}_{\text{orig}}$ into the generalized $\mathcal{D}_{\text{min}}$), \emph{no} full-granularity data from $\mathcal{D}_{\text{orig}}$ is collected during the \emph{model training} or \emph{deployment} phases.

To accurately represent the empirical privacy risk of a generalization, we introduce a wide range of policy-aligned adversaries in \cref{sec:metrics}. 
In particular, we directly reference the EU Working Party \citet{GDPR3Attacks}, which outlines the threats of \emph{inference} (\ie \emph{reconstruction}), \emph{linkability}, and \emph{singling out} for anonymized data.
We formalize these concepts and define several adversaries that aim to \emph{reconstruct} the personal attributes of a client $\vx$ based on the leaked generalized record $\vz$, utilizing different degrees of side information, as well as adversaries that aim to use the generalized dataset to link two partial datasets $\vx_A$ and $\vx_B$ (\emph{linkability}) or isolate a single individual from the dataset (\emph{singling out}). 

\begin{table*}[t]
  \centering
  \caption{Summary of all adversaries introduced in \cref{sec:metrics} to evaluate the empirical privacy risk. All adversaries, as prior knowledge, have access to the generalization $g$ and a small set of full-granularity records $\advOrig$. As part of the breach, they additionally have observed a set of generalized records $\advMin$. We evaluate all adversaries in \cref{sec:experimental}.}
  \label{tab:adversaries}
  \resizebox{!}{6.25em}{
  \begin{tabular}{rllll}
    \toprule 
    Attack vector & & Adversary & Side information & Goal \\ 
    \midrule
    Reconstruction& A1 & Reconstruction & / & Reconstruct personal attributes $\vx_P$ \\ 
    &A2 & High-certainty Reconstruction & / & Reconstruct $\vx_P$ with high confidence \\ 
    &A3 & Non-personal Knowledge & $\vx_N$ & Reconstruct personal attributes $\vx_P$ \\ 
    &A4 & Leave-one-out & $\vx_N$ and $\vx_{P \setminus \{p\}}$ & Reconstruct attribute $x_p$ \\ 
    &A5 & Partial Personal Knowledge & $\vx_N$ and $\vx_{\{1, \ldots, k-1\}}$ & Reconstruct attribute $x_k$ \\
    &A6 & Multi-breach Reconstruction & Another set of $g$ and $\advMin$ & Reconstruct attributes $\vx_P$ \\ 
    Linkability&A7 & Linkability & $\vx_A$ and $\vx_B$ for $k$ individuals & Link each $\vx_A$ to a corresponding $\vx_B$ \\ 
    Singling Out&A8 & Singling Out & / & Isolate a single individual \\ 
    \bottomrule
  \end{tabular}}
  \vspace{-0.5em}
\end{table*}

\para{Minimization via Generalization} \label{ssec:problem:gen}
Formally, we propose to train $f$ on low-granularity \emph{generalized} records $\vz \in \mathcal{Z}$ instead of full-granularity records $\vx \in \mathcal{X}$.
To this end, we define a \emph{generalization} function $g \colon \mathcal{X} \to \mathcal{Z}$ (\emph{global}), which reduces data granularity, produced by a \emph{minimizer} (an algorithm or a human).
We set $\mathcal{Z} = \mathcal{Z}_1 \times \mathcal{Z}_2 \times \ldots \times \mathcal{Z}_d$ and $\mathcal{Z}_i = \{1, 2, \ldots, k_i\}$ such that $g(\vx) := (g_0(x_0), g_1(x_1), \ldots, g_d(x_d))$ generalizes each attribute independently (\emph{single-dimensional}). Further, we require the image of $g$ to be a proper partition of each dimension (attribute) of $\mathcal{X}$ (\emph{strict}).
We distinguish continuous attributes (\eg \texttt{salary}) and discrete ones (\eg \texttt{occupation}).
Continuous $x_i$ are scaled to $[0, 1]$ and transformed using a non-decreasing function $g_i\colon [0, 1] \rightarrow \mathcal{Z}_i$.
Discrete $x_i$ with values $\{1, 2, \ldots, c_i\}$, where $c_i \geq k_i$, are transformed using $g_i\colon \{1, 2, \ldots, c_i\} \rightarrow \mathcal{Z}_i$. We use $\distMin$ to denote the induced distribution of $(\vz = g(\vx), y)$.
Finally, we note that all our generalizations are both independent of respective downstream classifiers (results in \cref{app:architectures} indicate that they transfer well between different downstream architectures) and as depicted in \cref{ssec:experiments:qualitative} easy to apply for clients, justifying the \fullcirc[0.75ex] on \emph{Client Assumptions} in \cref{tab:comparison}.

\para{Proposed Workflow}
With this setup, the \emph{vDM workflow}, depicted in \cref{fig:overview}, consists of three distinct phases aiming to minimize the amount of full-resolution data that ever enters the system (\ie has to be collected from clients). In the \emph{minimization phase}, we collect a small (see \cref{ssec:experimental:sizes}), well-protected set $\setOrig$ of full-granularity pairs $(\vx, y) \sim \distOrig$ (\quartcirc[0.75ex]~for \emph{Trust in Collector}), which a minimizer uses to propose a set of generalizations $g^{(i)}$ . 
We evaluate those in terms of utility and empirical privacy risk and select the most suitable one ($g$).
Afterward, there is no more need to collect full-resolution data from clients. 
In the \emph{model training phase}, we collect a large set $\setMin$ of \emph{generalized} pairs $(\vz, y) \sim \distMin$ and train a classifier $f$ on it (\halfcirc[0.75ex]~for \emph{Privacy (Wire)} as we already collected $\setOrig$).
vDM imposes no restrictions on used architectures/algorithms and we show in \cref{sec:experimental} that the resulting loss in utility is small (\fullcirc[0.75ex]~for \emph{Technical Feasibility}).
Finally, in the \emph{deployment phase}, we deploy $f$ for inference and answer queries from clients, who only need to input their \emph{generalized} records $\vz$ (\fullcirc[0.75ex]~for \emph{Trust in Collector}, \threequarcirc[0.75ex]~for \emph{Privacy (New Record)}) and receive predictions $f(\vz)$. 

As shown in \cref{fig:overview}, during the training and deployment phases, \emph{no full-granularity data enters the system}, \eg individuals are never asked about their exact age.
This simplifies the security analysis, as regardless of the breach target (\eg collection, processing, or storage), the only data that can be leaked in the latter two phases are generalized records. 
This also entails that membership inference attacks can only recover generalized records (\halfcirc[0.75ex]~for \emph{Privacy (Model)}).
While the minimization phase still requires some full-granularity data, we argue that this phase can be protected against breaches more easily as opposed to a live deployment---data policies can be stricter, \eg via external auditing, ensured data deletion, or the use of commercial data clean rooms \cite{chen_training_2020, decentriq}, which can be impractical for a live deployment.

\para{Evaluating Generalizations}
As previously discussed, $g$ should produce generalized data with both \emph{high utility} and \emph{low empirical privacy risk}, two goals generally at odds.
The former means that generalized data should contain enough information to solve the original task---formally, to keep the utility risk $\mathcal{UR}({g})$ low, which is defined as the error rate of the best possible classifier $f$ predicting $y$ from $\vz = g(\vx)$:

\begin{equation} \label{eq:clfobj}
  \mathcal{UR}({g}) := \min_f \underset{(\vz, y) \sim \distMin}{\E}  \mathbbm{1} \left\{ f(\vz) \neq y \right\}.
\end{equation}

Evaluating the \emph{empirical privacy risk} of $g$ is more involved.
Namely, despite the advantages of the proposed workflow noted above which reduces the attack surface in case of a data breach, it remains unclear how much the generalized records leaked in the training or deployment phases reveal about individuals, to which extent their privacy is protected, and how this should be quantified.
We thoroughly study this question next.

\section{Assessing the Empirical Privacy Risk} \label{sec:metrics}

In this section, we formulate a comprehensive set of \emph{adversaries} with different attack capabilities, all aiming to use the data from a breach to compromise client privacy in various ways, often by training adversarial models.
Our adversaries can serve as a tool for evaluating vDM or provide insights to organizations in the minimization phase.
For this, we define a subset of attributes $P \subseteq \{1,2,\ldots,d\}$ as \emph{personal} (with the rest $N = \{1,2,\ldots,d\} \setminus P$ being non-personal).

\para{Threat Model} \label{ssec:metrics:threadmodel}
We start by defining the assumed prior knowledge of our adversaries.
First, we assume all adversaries know the generalization $g$ being attacked, as $g$ needs to be open to every party providing data in the training and deployment phases.
Second, all adversaries have a small set $\advOrig$ of full-granularity records $\vx$ (we ignore $y$ here for simplicity)---this can be obtained either by a breach in the minimization step ($\advOrig \equiv \setOrig$, the setting considered in \cref{sec:experimental}) or by obtaining other samples from $\distOrig$. 
Notably, this implies that the adversary knows $g(\advOrig)$, \ie the generalized records corresponding to $\advOrig$.

With this prior knowledge of $g$ and $\advOrig$, we define a \emph{breach} as an event where the adversary observes a set of generalized records $\advMin$, obtained by compromising the data collection or storage pipelines in the training or deployment phase, or even via model inversion \cite{ModelInv} on $f$. \emph{All} adversaries we will now introduce share this base threat model, reflecting different ways of utilizing $\advMin$ to compromise client privacy, often with additional side information.
Note that our base threat model is quite generous to the adversary (\eg in terms of knowledge of $\distOrig$), and thus even our weakest adversary models a relatively strong attacker.

\para{Overview of Adversaries}
Our set of adversaries is diverse, taking into account various levels of side information, but more importantly, directly aligned with policy, with each adversary corresponding to attacks described in \citet{GDPR3Attacks} (recently studied for synthetic data in \citet{giomi2023privacysynthetic}).
While practitioners can decide that some threats are more relevant, this broad set of adversaries enables a thorough evaluation of the empirical privacy risk.
We now describe the adversaries (labeled A1-A8 and summarized in \cref{tab:adversaries}).
We develop practical implementations of all adversaries and apply them in \cref{sec:experimental} evaluating our minimizers from \crefrange{sec:baselines}{sec:algorithm}.

\para{(A1-A2) Reconstruction}
The goal of the reconstruction adversary is to use its prior knowledge of $\advOrig$ to learn a function $h$, which can be used to recover personal attributes $h(\vz)_P \approx \vx_P$ of breached records $\vz \in \advMin$. Formally, the \emph{Reconstruction (A1)} adversary aims to find $h$ minimizing the error rate of predicting $\vx_P$ from $\vz$. 
We can use this to define a corresponding empirical privacy risk
\begin{equation*} \label{eq:advobj}
  \mathcal{PR}_{A1}(g) := \min_h \underset{(\vx, y) \sim \mathcal{D}_{\text{orig}}}{\E} \left[ \frac{1}{|P|} \sum_{p \in P} \mathbbm{1} \left\{ h(g(\vx))_p \neq x_p \right\}  \right],
\end{equation*}
which a practical implementation of A1 approximates by sampling from $\advOrig$.
When combining different personal attributes in $S$, we focus on mean aggregation as a reasonable choice for the generic case. We explore this choice for adversaries in \cref{ssec:experimental:individual} and minimizers in \cref{app:meanvsmax}.

However, leaking one personal attribute of a single individual with high certainty often has more direct privacy implications than an aggregate metric over many attributes and individuals.
Thus, we introduce the \emph{High-certainty Reconstruction (A2)} adversary, which has the same goal as A1 and is trained the same way but calculates the \emph{confidence} for each predicted attribute (\ie how certain it is that the prediction is correct) and outputs only predictions with the highest {confidence}. While we refer to \cref{sec:experimental} for details, intuitively, A2 uses logit magnitudes as a proxy for \emph{confidence}, identifying data points that have particularly high empirical privacy risk. %
We note that empirical privacy risk under A2 can be formalized similarly to $\mathcal{PR}_{A1}(g)$.
While we focus on A2 as a variant of A1, the idea of confidence can, in principle, be applied to any adversary.

\para{(A3-A6) Reconstruction with Side Information} 
A typical scenario is that an adversary has \emph{side information} about an individual (potentially from another breach) and is aiming to utilize this to boost the leakage of unknown personal attributes \cite{auxdata}.
To investigate this for vDM, we instantiate several reconstruction adversaries strictly stronger than A1.

The \emph{Non-personal Knowledge (A3)} adversary has knowledge of all non-personal attributes and thus aims to use $(\vz, \vx_N)$ to reconstruct $\vx_P$.  
The significantly stronger \emph{Leave-one-out (A4)} adversary models the worst case, knowing $(\vz, \vx_N, \vx_{P \setminus \{p\}})$, \ie all other attributes when predicting $x_p$.
The \emph{Partial Personal Knowledge (A5)} adversary models the intermediate cases between A3 and A4, giving us granular insight into how gracefully a generalization degrades under side information. 
Assuming $P=\{1, \ldots, p\}$, A5 predicts the personal attribute $\vx_k$
having knowledge of $(\vz, \vx_N, \vx_1, \ldots, \vx_{k-1})$. The cases $k=1$ and $k=p$ recover A3 and A4, respectively.  In \cref{sec:experimental}, we evaluate A5 by averaging its error over all choices of $k$.

Finally, the \emph{Multi-Breach Reconstruction (A6)} adversary focuses on the case where side information comes from the same source due to several breaches at different points in time.
Assume the case where an organization switches from a generalization $g^{(1)}$ to a different generalization $g^{(2)}$, and the adversary observes two breaches $\advMin$ and $S''_{\text{min}}$ corresponding to the same individuals.
Intuitively, observing two (or $k$) sufficiently different generalizations of the same individual can boost reconstruction. 
The goal of A6 is to evaluate the resilience of minimizers to repeated breaches.

\para{(A7-A8) Linkability and Singling Out} 
Next, we consider two non-reconstruction adversaries motivated by \citet{GDPR3Attacks}.
The increasing availability of data makes these attacks common in practice and often an essential first step towards mounting more powerful attacks ~\cite{linkability,giomi2023privacysynthetic,sideChannelFranziska,deepSinglingOut}.

The \emph{Linkability (A7)} adversary observes as side information, \eg from another organization using similar attributes, full-granularity \emph{partial records} (\ie records with a subset of attributes) for a set of individuals, and aims to use the set of generalized records $\advMin$ to connect partial records belonging to the same individual.
More formally, for disjoint subsets $A$ and $B$ of attributes $\{1, \ldots, d\}$, A7 has $k$ partial records of the form $\vx_A$ and $\vx_B$, and aims to predict for each $\vx_A$ which $\vx_B$ corresponds to the same individual. 

Finally, the goal of the \emph{Singling Out (A8)} adversary is to isolate a single individual from the dataset, a concept similar to having a small anonymity set which is a known issue in privacy-related areas \cite{dingledine2006anonymity}. 
Formally, it observes the breached generalized records $\advMin$ and outputs a predicate $\Pi$ that, when applied to the full-granularity records that produced $\advMin$, return exactly one individual.

\section{Baseline Minimizers for vDM} \label{sec:baselines}

Having introduced both the vDM setting and our adversaries, we now present several vDM algorithms that will establish baselines for our \toolshort~minimizer (\cref{sec:algorithm}).

\para{Uniform Minimizer} As an initial simple baseline, we consider a uniform minimizer. 
Let $k_i$ be a hyperparameter that denotes how many elements (buckets) $\mathcal{Z}_i$ should at most have.
Given a hyperparameter $k$, the uniform minimizer uses $k_i=k$ for all attributes and generalizes discrete attributes $x_i$ uniformly at random, and continuous attributes $x_i$ to $\lceil k x_i \rceil$. 

\para{Feature Selection Minimizer} 
As attribute suppression (\ie removal of an attribute) is a special case of generalization, we consider a feature selection minimizer which keeps $k$ of $d$ attributes based on ANOVA F-values, as this method supports both continuous and categorical attributes. In~\cref{app:details:moreresults} we further explore more methods and variants.

\para{Apt Minimizer}
We adapt the recently proposed \emph{Apt} method~\citep{IbmApt} (discussed in~\cref{ssec:related:op}) to our setting. \emph{Apt} uses information loss metrics and a decision tree to model the decisions of a classifier trained on original data. As the method was infeasible to run for dataset sizes we consider ($>24h$ where our minimizer \toolshort, requires $\approx1s$), we limit the tree-depth in \emph{Apt} to $10$ and each run to $2h$.

\para{Iterative Minimizer}
Given $k$, the \emph{Iterative} minimizer starts from a heuristic generalization $g$ with $k_i=k$, splitting continuous attributes based on $k$-quantiles and discrete ones based on their weight in a logistic regression model, minimizing the average variance of weights in each group.
It then iteratively improves $g$ by reducing $k_i$ using dynamic programming while keeping the resulting classifier error below some threshold. To determine the order in which attributes are generalized, it sorts all attributes based on their estimated impact on classification and adversarial error. We refer to~\cref{app:details:minimizers} for a more detailed description.

\para{Neural Minimizers}
Finally, we propose two vDM minimizers that model $g$ using neural networks.
We present a brief overview of how such modeling is done for continuous attributes and provide a corresponding description for discrete attributes and more details in \cref{app:details:minimizers}.

Both minimizers model $g$ as a set of $d$ independent neural networks $g^{(i)}$, with each network responsible for generalizing a single attribute.
Let $i$ denote the index of a continuous attribute of record $\vx$ and let $\vx_i$ be normalized to $[0,1]$. $g^{(i)}$ learns a monotonic and differentiable generalization by first learning a monotonic transformation $M\colon [0,1] \to [0,1]$.
Based on work on monotonic neural networks \cite{mono1,mono2}, $g$ ensures $M$'s monotonicity by constraining all linear layer weights to $W \odot W \geq 0$, using tanh activations and batch normalization. The output interval is then split uniformly into $k_i$ buckets (identified by center points $c_j$), and for a record $\vx$, the probability of generalizing attribute $i$ to bucket $j$ is taken as the softmax over the bucket-distances $(g^{(i)}(\vx_i)-c_j)^2$.

\para{AdvTrain Minimizer} 
The \emph{AdvTrain} minimizer, inspired by \citet{MadrasAdvTrain}, utilizes adversarial learning \cite{Gans} to jointly optimize the generalization $g_\psi: \mathcal{X} \to \mathcal{Z}$, classifier $f_\theta: \mathcal{Z} \to \mathcal{Y}$ and adversary $h_\phi: \mathcal{Z} \to \mathcal{X}_s$, all of which we model as neural networks. \emph{AdvTrain} then optimizes the following objective:
\newcommand{\smallminus}{\scalebox{0.7}[1.0]{$-$}} %
$$
  \min_{\theta, \psi} \max_{\phi} \E_{\vx, y} \left[ (1\smallminus\lambda) {L}_{\text{clf}}(f_\theta(g_\psi(\vx)), y) \smallminus \lambda {L}_{\text{adv}}(h_\phi(g_\psi(\vx)), \vx) \right],
$$
where $\lambda$ is a factor determining the tradeoff between the classification and adversarial error.
We instantiate ${L}_{\text{clf}}$ as the BCE loss between the predicted and true label, while ${L}_{\text{adv}}$ denotes the CE loss between the predicted and true personal attribute (averaged over all such attributes).
During training, we optimize $g_\psi$ and $f_\theta$ to reduce ${L}_{\text{clf}}$ and increase ${L}_{\text{adv}}$ and optimizing $h_\phi$ to reduce ${L}_{\text{adv}}$. For each step of optimizing $g_\psi$ and $f_\theta$, we take $N_{\text{inner}}$ steps optimizing $h_\phi$.

\para{MutualInf Minimizer}
The final baseline, \emph{MutualInf}, uses the same $g_\psi$, $f_\theta$, and training procedure as \emph{AdvTrain} but replaces the adversarial objective with one minimizing the mutual information between $\vx$ and its generalization $\vz=g_\psi(\vx)$:
$$
  \min_{\theta, \psi} \E_{\vx, y} \left[ (1 - \lambda) {L}_{\text{clf}}(f_\theta(g_\psi(\vx)), y) + \lambda {L}_{\text{inf}}(g_\psi(\vx), \vx) \right].
$$

To derive $L_{\text{inf}}$, we start from the definition of mutual information $I(\vz, \vx) = H(\vz) - H(\vz | \vx)$, and apply Jensen's inequality to independently derive upper bounds on $H(\vz)$ and $H(\vz | \vx)$, which can be approximated via sampling. In~\cref{app:details:minimizers} we describe this process in more detail.

\section{\tool~Minimizer} \label{sec:algorithm}
We now propose our minimizer, \emph{\tool} (\toolshort), that builds a generalization $g$ using a decision tree.

\para{Classification Trees}
We first recall key concepts related to classification trees.
Let $\mathcal{S}_{\text{root}}=(\vx,y)\in \R^{d}\times\{0,1\}$ be a dataset for binary classification. 
A decision tree $T$ repeatedly splits a leaf node $L$ with assigned dataset $\mathcal{S}_L$ into children nodes $L^\leq$ and $L^>$, by picking an attribute $j \in [1,d]$ and choosing a threshold value $v$ such that \mbox{$\mathcal{S}_{L^\leq} = \{(\vx,y) \in \mathcal{S}_L \mid x_j \leq v \}$} and \mbox{$\mathcal{S}_{L>} = \mathcal{S}_L \setminus \mathcal{S}_{L^\leq}$}. 
The goal is to select $v$ such that it splits samples with different $y$. 
A common criterion for selecting $j$ and $v$ is to minimize the \emph{Gini impurity} $Gini_y(\mathcal{S}) = 2\cdot p_y(1-p_y) \in [0,\frac{1}{2}]$ where $p_y = \sum_{(\vx,y')\in \mathcal{S}} \mathbbm{1}_{y'=y} / |\mathcal{S}|$ denotes the relative frequency of class $y$ in $\mathcal{S}$.
We build $T$ by repeating this procedure until we reach a predefined maximum number of leaf nodes $k^*$. 
At inference, we propagate a point $\vx'$ through $T$, reaching a leaf node $L'$, and returning the majority class of $\mathcal{S}_{L'}$.
 
\para{Categorical Splits} 
As splits in a decision tree generally are of the form $x_j \leq v$, this limits the tree's ability to partition one-hot-encoded categorical attributes effectively. Typical implementations of decision trees can only single out one category per split ($x_{j} \leq 0.5$). We avoid this issue in PAT by using the \emph{fairness-aware categorical splits} introduced in \cite{jovanovic2023fare}.
Namely, we represent categorical attributes not via one-hot encoded vectors but instead each category for an attribute $x_j$ with $|C|$ classes with a unique index in $\{1,\textellipsis ,|C|$\}. We explore multiple ways to sort the indices, and for each sorting, consider all possible prefix-postfix splits.

\begin{figure*}[!tbp]
    \newcommand{\relSubfigWidth}{0.5\textwidth}
    \newcommand{\innerWidth}{\textwidth}
    \centering
    \minipage{0.32\textwidth}
      \centering 
      \includegraphics[width=\innerWidth]{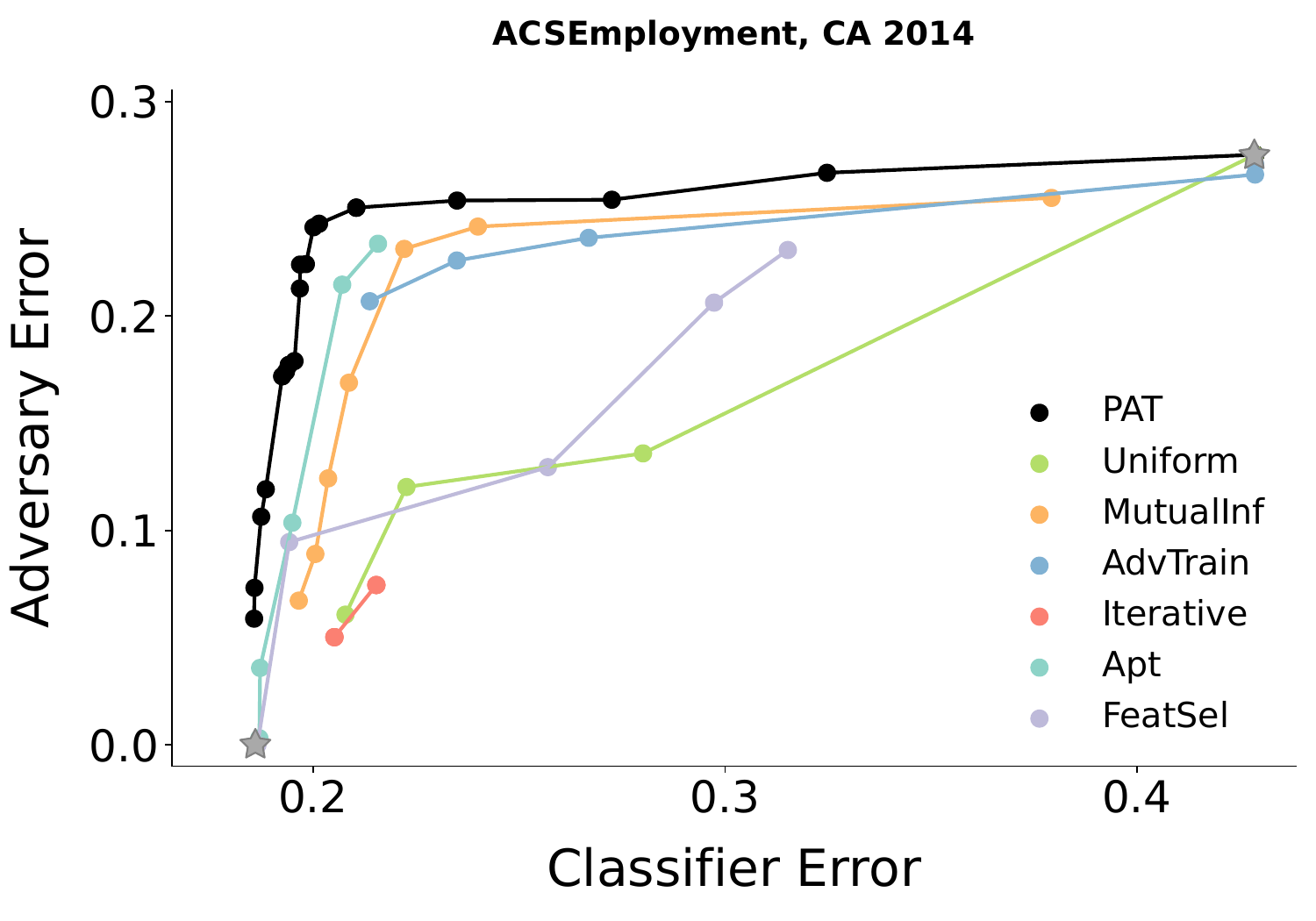}
    \endminipage
    \hfill 
    \minipage{0.32\textwidth}
        \centering 
        \includegraphics[width=\innerWidth]{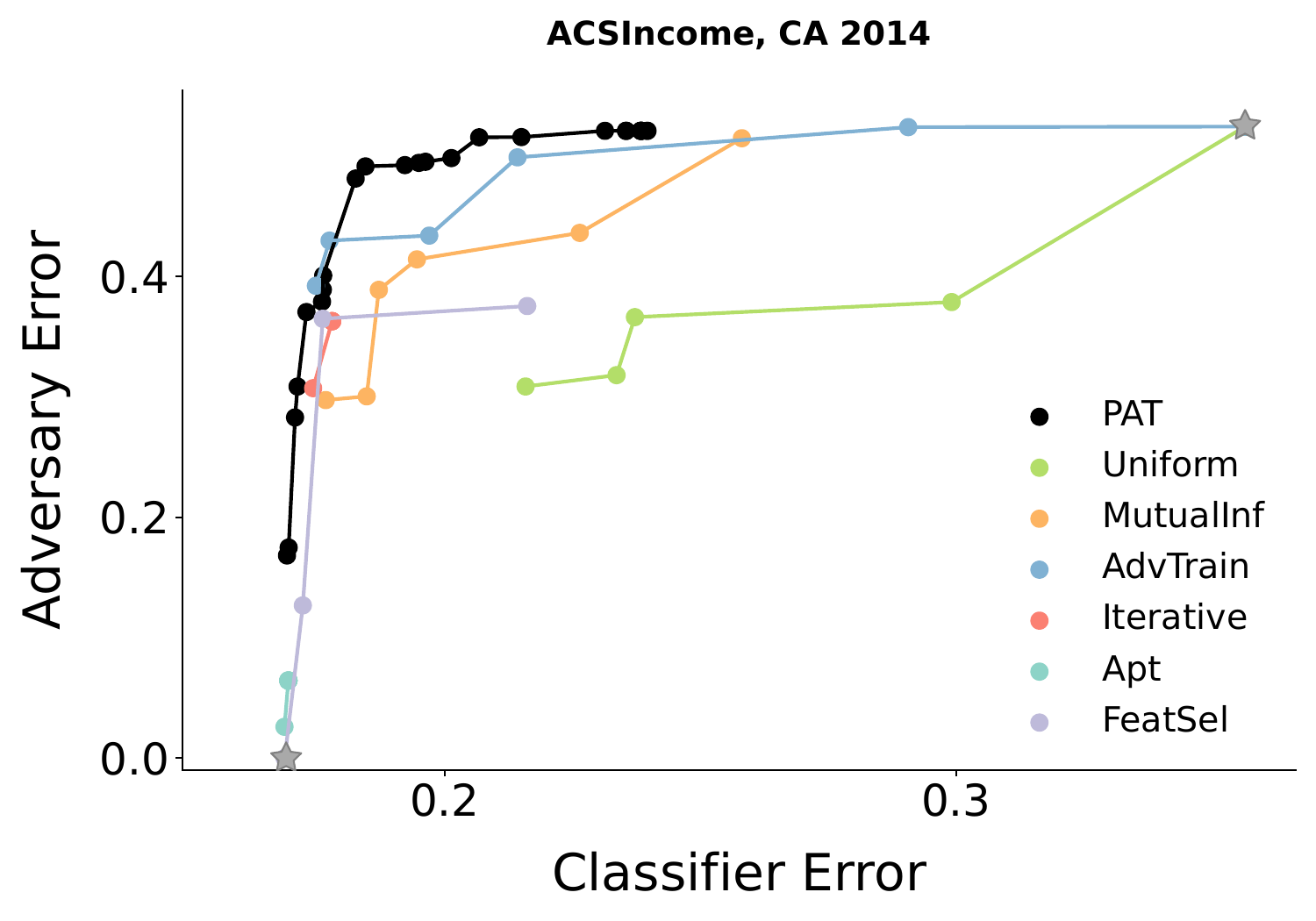}
    \endminipage
    \hfill
    \minipage{0.32\textwidth}
        \centering
        \includegraphics[width=\innerWidth]{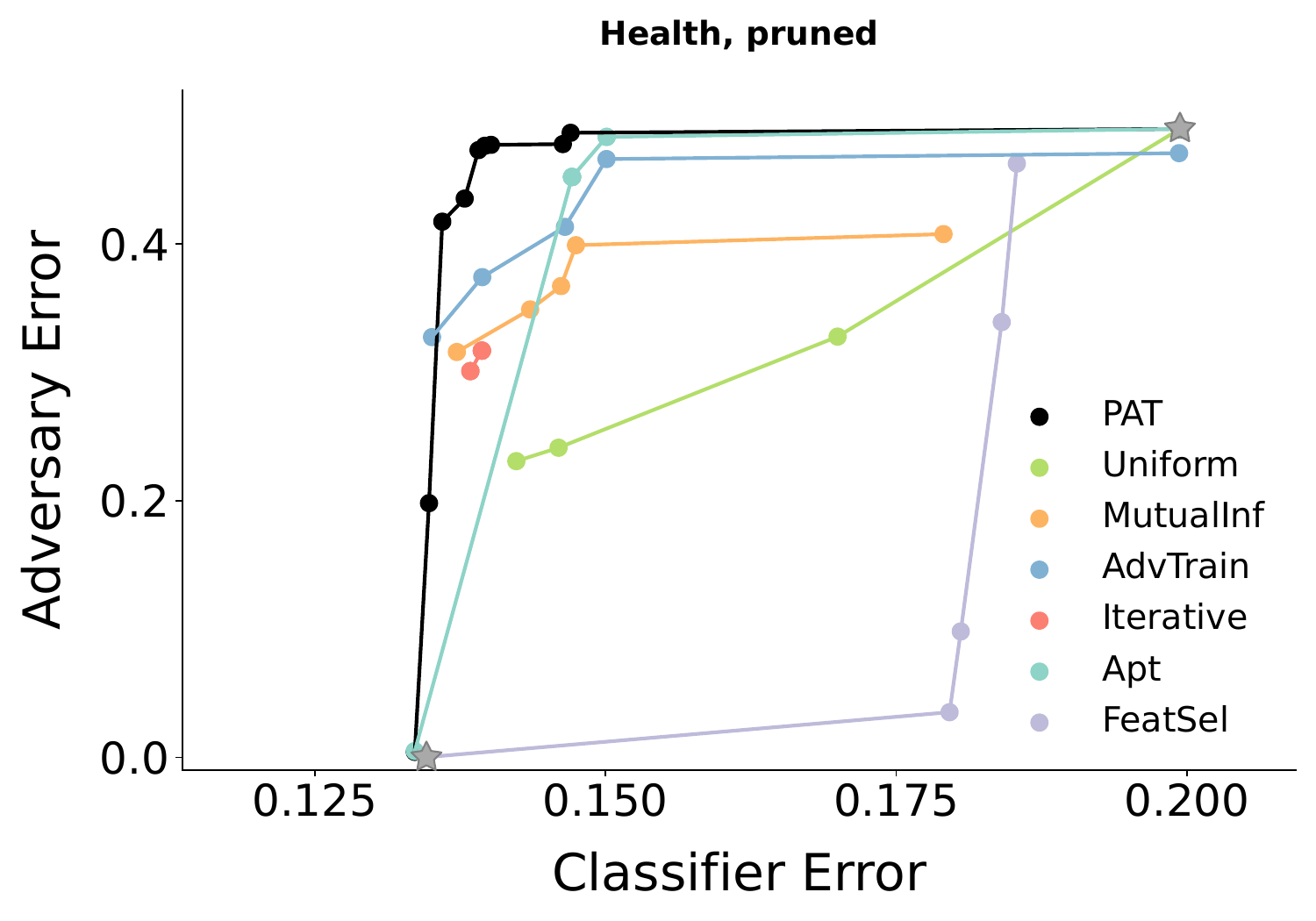}
    \endminipage
    \caption{Utility-privacy tradeoffs of candidate generalizations produced by minimizers on ACSEmployment, ACSIncome, and Health (pruned). Classifier and adversarial errors are reported on a held-out test set, while the candidate generalizations are selected on the validation set. Across all datasets, PAT generally achieves the most favorable utility-privacy tradeoff.}
    \label{fig:acs}
    \vspace{-0.5em}
\end{figure*}

\para{\emph{PGini} Criterion}
The next modification compared to standard classification trees is in the criterion, usually focused solely on utility.
We aim to include a privacy-aware component, where ideally one can explicitly control the utility-privacy tradeoff.
To this end, we extend prior work in fair trees \cite{zhang_faht_2019, jovanovic2023fare} to propose \emph{PGini}, a privacy-aware criterion that accounts for the distribution of multiple personal attributes.

Let $P$ denote our set of personal attributes, and $\mathcal{S}_D$ be a concrete dataset. We modify the multi-class Gini impurity, defined on an attribute $a$ with $V$ possible values as
$$Gini_a(\mathcal{S}_D) =\sum_{v=0}^{V-1} p_{a,v} \cdot (1-p_{a,v}) \in [0,1-\frac{1}{V}]$$ 
where $p_{a,v} = \sum_{(\vx,y)\in \mathcal{S}_D} \mathbbm{1}\{\vx_a=v\} / |\mathcal{S}_D|$, and define:
\begin{equation} \label{eq:multifairgini}
\begin{aligned}
    PGini(\mathcal{S}_D) =& (1-\alpha) \cdot 2\cdot Gini_y(\mathcal{S}_D)~+ \\
&\alpha \left( 1 - \frac{1}{|P|} \sum_{p \in P} \sigma_p \cdot Gini_{p}(\mathcal{S}_D) \right)
\end{aligned}
\end{equation}
where $\sigma_p = \frac{|p|}{|p|-1}$ and $2$ normalize the Gini values to $[0,1]$.

The utility term in \cref{eq:multifairgini}, $Gini_y(\mathcal{S}_D)$, is minimized as in the usual usage of Gini impurity.
Intuitively, this promotes partitions of the input space where each region is still predictive of the target label.
In contrast, the privacy term $Gini_{p}(\mathcal{S}_D)$ is \emph{maximized}, promoting partitions where each region is \emph{non-predictive} of personal attributes.
The parameter $\alpha \in [0, 1]$ allows for a smooth tradeoff between these two goals, where larger $\alpha$ implies more focus on privacy.
This separates our approach from prior tree-based vDM approaches, such as \cite{IbmApt}, which learn a decision tree without accounting for personal attributes and achieve the targeted utility-privacy tradeoff via pruning.
Further, $PGini$ can be easily adapted to weigh personal attributes individually in case some are deemed more sensitive than others.

\para{Obtaining a Generalization}
While the leaves of $T$ define a partition of the input space $\mathcal{X}$, the splits in each node $V$ of $T$ depend on prior splits in $V$'s ancestors $\mathbb{A}(V)$. 
As the splits in $\mathbb{A}(V)$ might be on different attributes than the split in $V$, the partition defined by $T$ is not by itself a strict generalization as defined in \cref{sec:problem}. 
Hence, to construct a generalization $g$, PAT post-processes $T$, partitioning each attribute into ranges by taking the union of all split thresholds for that attribute encountered in $T$.
Notably, while $g$ partitions $\mathcal{X}$ into strictly more granular subsets than $T$, it holds for all $\vx \in \mathcal{X}$ that $T(\vx) = T(g(\vx))$, \ie $g$ fully preserves the utility of $T$.

\section{Experimental Evaluation} \label{sec:experimental} 

\begin{figure}
	\centering
	\includegraphics[width=0.35\textwidth]{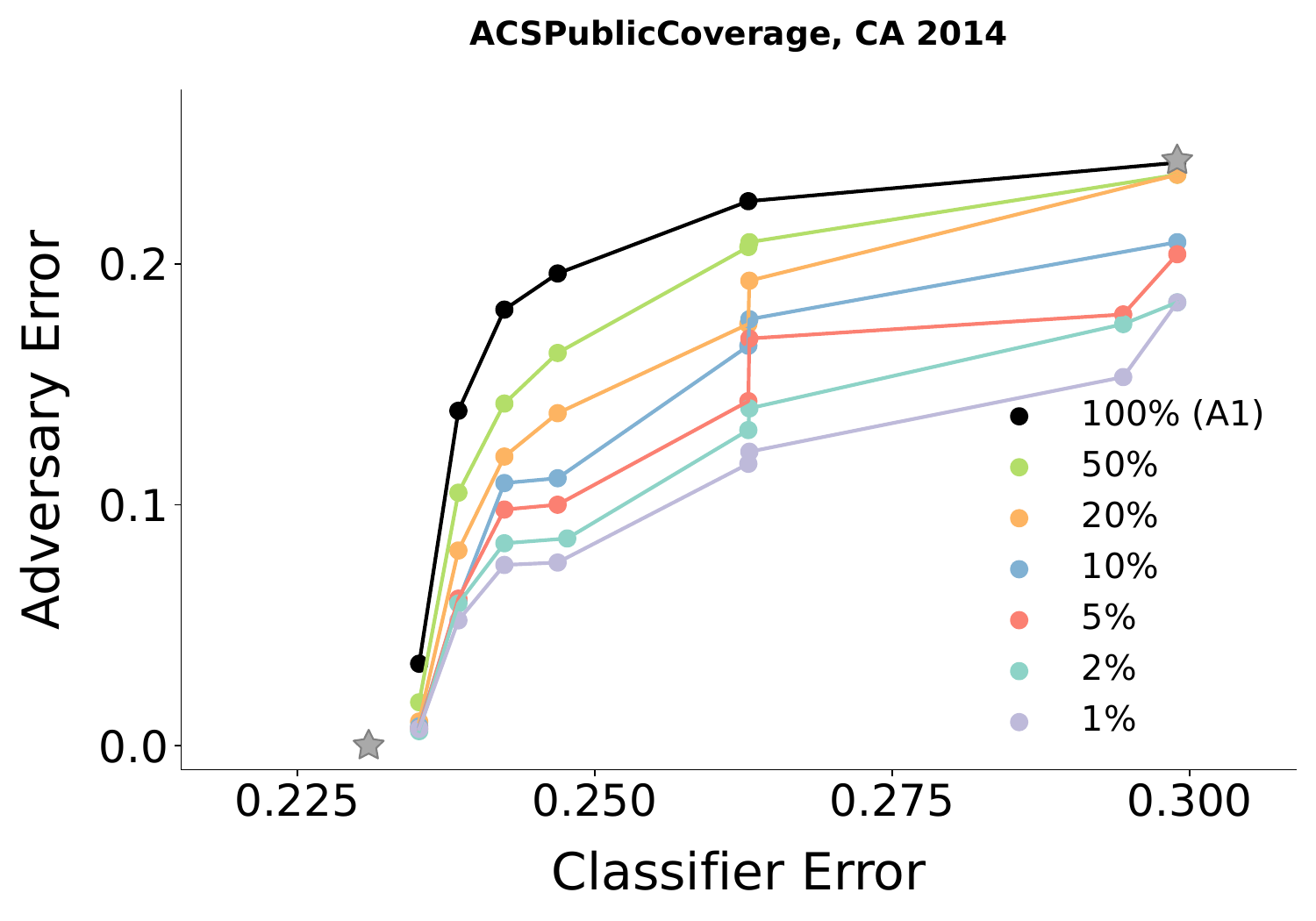}
    \caption{A2 reconstruction error for several generalizations with \emph{AdvTrain} on the ACSPublicCoverage test set.
    }
    \vspace{-0.5em}
  \label{fig:highcert}
\end{figure}

In this section, we present an extensive experimental evaluation of our vDM setting utilizing all adversaries from~\cref{sec:metrics} and minimizers introduced in~\crefrange{sec:baselines}{sec:algorithm}. 

In \cref{ssec:mainres} we present our main results, followed by a study of individual attribute reconstruction in \cref{ssec:experimental:individual}, minimizer training sizes in~\cref{ssec:experimental:sizes} and a qualitative study in~\cref{ssec:experiments:qualitative}.
In our supplementary material, we provide additional details on adversaries (\cref{app:adversaries} and \cref{app:architectures}) and experimental parameters (\cref{app:details:params}).%

\subsection{Main Results} \label{ssec:mainres}
Generalizations $g$ are learned on a fixed training set $\setOrig$, which we here assume to be observed by the adversary ($\advOrig \equiv \setOrig$). 
We set $\setMin = g(\setOrig)$, and split the data into three disjoint parts: training, validation and test. 
We train a classifier $f$ using training and validation splits, reporting the accuracy on the test split.
Finally, we set $\advMin$ to the generalized test split, assuming it is breached.

\para{(A1) Reconstruction} 
The A1 adversary tries to reconstruct the personal attributes of the generalized records from $\advMin$.
We run each minimizer with different parameters to produce a diverse set of generalizations and report the utility-privacy tradeoff on two datasets from the ACS suite \cite{RetiringAdult}, derived from US census, predicting individuals' employment and income, respectively.
Additionally, we evaluate on a preprocessed Health \cite{noauthor_heritage_nodate} dataset, predicting the Charlson Comorbidity Index. We provide more details on our datasets and their preprocessing in~\cref{app:details:dataset}.  

\cref{fig:acs} shows the classifier error $\mathcal{UR}({g})$ and mean adversary error $\mathcal{PR}_{A1}(g)$ on test data for each type of minimizer. We plot all points, marking those that remain on the Pareto front when using the test set.
The \score{0}{1} markers represent the two \emph{limits of generalization}: (i) fully-generalized data (all $k_i=1$), where both classifier and the adversary predict solely based on the class frequency of each personal attribute, and (ii) non-generalized data (all $k_i=c_i$), giving a lower bound on classification error with trivial adversary error of $0$.
We observe that PAT consistently achieves the most favorable utility-privacy tradeoffs across all datasets. Additionally, our new baselines provide a wide range of generalizations and can serve as benchmarks for future work.
In~\cref{app:details:moreresults} we show similar results on several additional datasets, and in~\cref{app:distshift} further demonstrate the robustness of PAT to temporal distribution shift.

Across all experiments, most minimizers show clear inflection points in the utility-privacy tradeoff. Taking PAT on ACSEmployment as an example, we find that there is almost no decrease in adversarial error until the classifier error reaches around $0.2$. Any decrease in classifier error after this comes with a significantly larger decrease in adversarial error. In addition to having pre-defined utility targets (for example, from requirements), these inflection points can assist practitioners in identifying generalizations that yield favorable utility-privacy tradeoffs.

\para{(A2) High-Certainty Reconstruction}
A2 operates in the same setting as A1, \ie tries to reconstruct samples from $\advMin$. Let $h(z)_a$ denote the logits that an A1 adversary predicts for attribute $a$. Further, let $\mathbbm{h}(z)_a = \max(h(z)_a)$ denote the maximum of the logits. We say that an A2 adversary is $k\%$-confident in its prediction for $a$ if $\mathbbm{h}(z)_a$ is in the $k$-th highest percentile of $\mathbbm{h}(\mathcal{Z},a) = \{\mathbbm{h}(z)_a \mid z \in \mathcal{Z}\}$, measuring confidence relative to other predictions for $a$.

\cref{fig:highcert} shows the results of A2 on the ACSPublicCoverage dataset predicting individuals' insurance coverage. We fix the minimizer to \emph{AdvTrain} (with similar results for other minimizers), and for each point in the Pareto front of the A1 adversary ($k=100\%$), we report the mean reconstruction error for several A2 adversaries with different $k$. We observe a significant decrease in the error rate for more confident adversaries, showing that A2 adversaries can recover specific individuals more accurately and relying solely on A1 underestimates the privacy risk.

\begin{figure}
	\centering
	\includegraphics[width=0.35\textwidth]{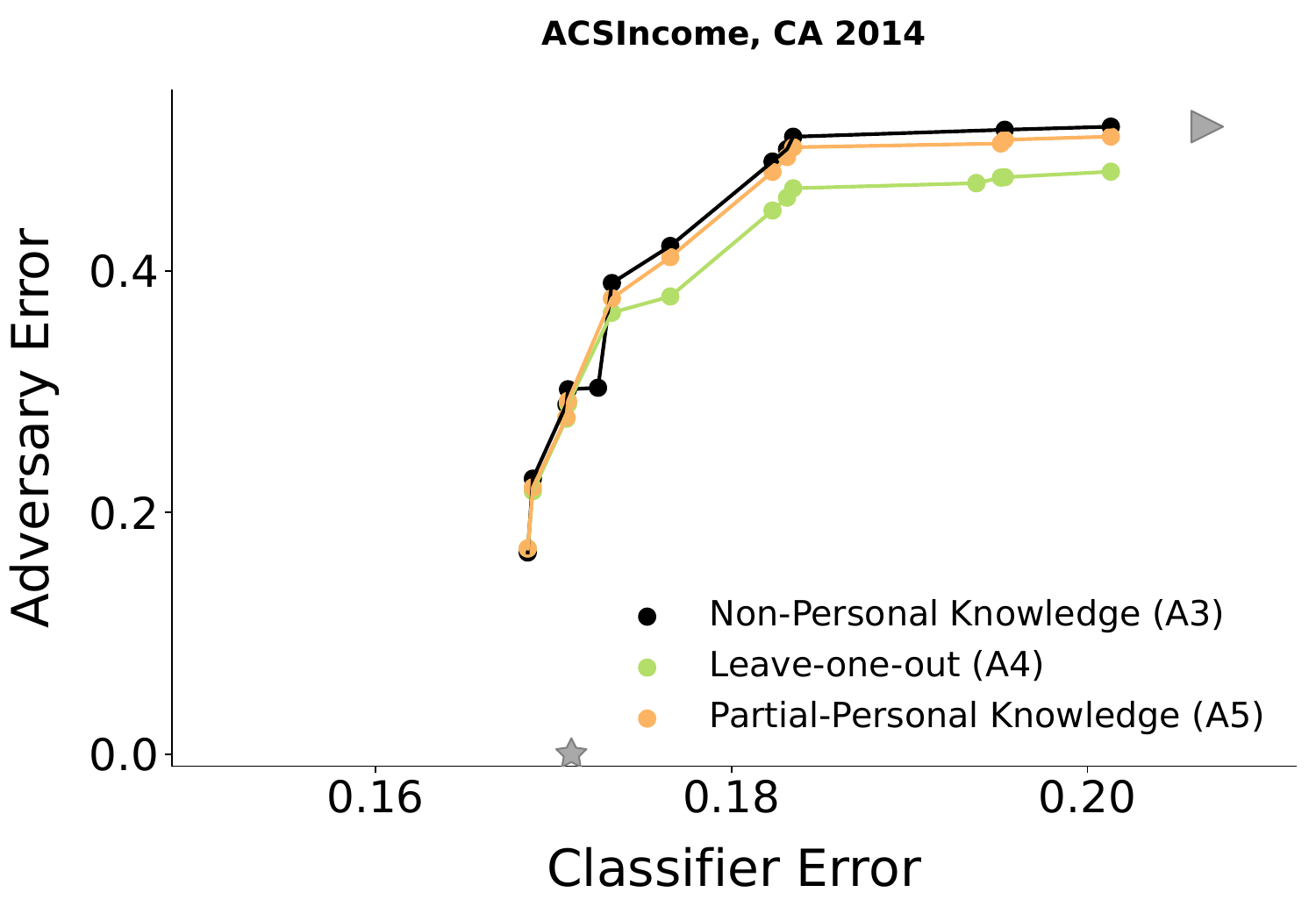}
  \caption{Pareto fronts of A3-A5 adversaries using PAT. In the low error regime, adversaries perform equally, as reconstruction is easier. In higher error regimes, A4 outperforms A3 and A5.}
  \label{fig:part_leakage}
  \vspace{-0.5em}
\end{figure}

\para{(A3-A5) Reconstruction with Side Information} 
Using the PAT minimizer, we evaluate A3-A5 via their Pareto fronts in the utility-privacy tradeoff on the ACSIncome dataset (\cref{fig:part_leakage}). As predicted in \cref{sec:metrics}, the \textit{Partial Personal Knowledge} (A5) adversary lies between the \textit{Non-Personal Knowledge} (A3)  and \textit{Leave-one-out} (A4) adversary. Looking closer at A4, we find that even though all but a single attribute have been leaked in full granularity, having this attribute generalized still hinders its reconstruction noticeably. This indicates that vDM offers valuable privacy protection even under heavy side information. Further, we find that the difference between the adversaries decreases in lower error regions. This is not unexpected, as the generalizations achieving lower classifier error are increasingly fine-grained.

\para{(A6) Multi-breach Reconstruction}
\cref{Tab:cont_release} reports the mean reconstruction error for several (C1-C4) multi-breach scenarios, each consisting of two breaches under different generalizations. To this end, we implement an A6 adversary by training two A1 adversaries and averaging their predictions. We find that the A6 adversary has a lower error than any A1 adversary on their respective individual breaches. This shows that {minimizing the same data multiple times can increase the privacy risk} warranting extra care. %

\para{(A7) Linkability}
We evaluate our A7 adversary on the ACSIncome dataset, picking two attributes, \texttt{OCCP} (\texttt{Occupation}) and \texttt{POBP} (\texttt{Place-of-Birth}), that we want to link. The A7 adversary can now use the minimized data $S'_{min}$ to approximate the distribution $p(\vx_{OCCP}\mid \vx_{POBP}=j)$ (see \cref{app:adversaries}) and predict the value for $\vx_{OCCP}$. \cref{Tab:linkage} shows how the probability of correctly linking entries significantly increases with access to higher granularity records. %

\begin{table}[t]
    \centering
    \tabcolsep=0.04cm
    \caption{Mean reconstruction error of A6 adversary in four MBR scenarios, using the \emph{Uniform} minimizer and PAT.}
    \begin{tabular}{
      S[table-format=1.2]
      S[table-format=1.2]
      S[table-format=1.2]
      S[table-format=1.2]
      S[table-format=1.2]
      S[table-format=1.2]
      S[table-format=1.2]
      S[table-format=1.2]
      S[table-format=1.2]
    }
     & \multicolumn{2}{c}{C1}
     & \multicolumn{2}{c}{C2}
     & \multicolumn{2}{c}{C3}
     & \multicolumn{2}{c}{C4}\\
    \cmidrule(lr){2-3}\cmidrule(lr){4-5}\cmidrule(lr){6-7}\cmidrule(lr){8-9}
    {\splitcell{$g$}}
     & {\small\splitcell{PAT \\ $\alpha {=} 0.1$}}
     & {\small\splitcell{PAT \\ $\alpha {=} 0.7$}}
     & {\small\splitcell{PAT \\ $\alpha {=} 0.3$}}
     & {\small\splitcell{PAT \\ $\alpha {=} 0.7$}}
     & {\small\splitcell{Unif. \\ $k{=}3$}}
     & {\small\splitcell{PAT \\ $\alpha {=} 0.7$}}
     & {\small\splitcell{Unif. \\ $k{=}4$}}
     & {\small\splitcell{PAT \\ $\alpha {=} 0.7$}} \\
    \midrule 
    {\splitcell{$\mathcal{PR}_{A1}$}}
    &0.43 &0.48&0.47&0.48&0.37&0.48&0.32&0.48 \\
    {\splitcell{$\mathcal{PR}_{A6}$}}
     & \multicolumn{2}{c}{\bfseries\splitcell{0.42}}
     & \multicolumn{2}{c}{\bfseries\splitcell{0.45}}
     & \multicolumn{2}{c}{\bfseries\splitcell{0.33}}
     & \multicolumn{2}{c}{\bfseries\splitcell{0.27}}\\
    \bottomrule
    \end{tabular}
    \label{Tab:cont_release}
    \vspace{-0.5em}
  \end{table}

\para{(A8) Singling Out}
Unlike the synthetic data setting \citep{xu_modeling_2019, giomi2023privacysynthetic}, in vDM, each $\vx \sim D_{orig}$ is directly mapped to exactly one $\vz=g(\vx)$. This makes it considerably easier for an A8 adversary to find a predicate $\Pi$, which singles out an individual from $\advMin$.
Intuitively, an adversary with access to $\advMin$ (and multiplicities of entries) can target $\vz$ which only rarely occur. In particular, if there exists a $\vz = g(\vx)$ which is only observed once, the adversary can single out the corresponding $\vx$ out by defining $\Pi_{\vx}$ based on the attribute ranges implied by $\vz$ (more detail in \cref{app:adversaries}). 

In \cref{Tab:singling_out}, we report how many $\vx$ get generalized to the \emph{rarest} $\vz$ in ACSEmployment (reported as \emph{Utilization}). When multiple $\vz$ are the rarest, we report their number in parentheses. For example, for PAT with $k^*=20$ (max. \# of leaves) and $\alpha=0.3$, we find $44$ different $\vz$, with each having only one $\vx$ getting generalized to it. As expected, more granular generalizations significantly increase the adversary's chance of finding low utilization $\vz$ to single out.

\para{Summary}
Our experiments have shown that the adversaries defined in \cref{sec:metrics} capture a diverse set of attacks on different aspects of vDM, establishing them as a comprehensive way to evaluate generalizations.
We have further demonstrated that \toolshort~consistently outperforms new and prior baseline minimizers across different settings.

\begin{table}[t]
    \centering
    \tabcolsep=0.05cm
    \caption{Percentage of entries correctly linked by A7 on the OCCP and POBP attributes (for different generalizations produced by PAT, \emph{Uniform}, and \emph{AdvTrain} minimizers). \emph{Rand.} denotes a baseline method that does not utilize $S'_{min}$. 
    }
    \begin{tabular}{
      S[table-format=1.2]
      S[table-format=1.2] 
      S[table-format=1.2]
      S[table-format=1.2]
      S[table-format=1.2]
      S[table-format=1.2]
      S[table-format=1.2]
      S[table-format=1.2]
      S[table-format=1.2]
    }
    {\bfseries\splitcell{$g$}}
     & {\small\splitcell{Rand. \\ \\}}
     & {\small\splitcell{PAT \\ $k^* {=} 8$ \\ $\alpha {=} 0.3$}}
     & {\small\splitcell{PAT \\ $k^* {=} 20$ \\ $\alpha {=} 0.3$}}
     & {\small\splitcell{PAT \\ $k^* {=} 50$ \\ $\alpha {=} 0.3$}}
     & {\small\splitcell{PAT \\ $k^* {=} 50$ \\ $\alpha {=} 0.9$}}
     & {\small\splitcell{Unif. \\ $k{=}3$\\}}
     & {\small\splitcell{Adv. \\ $\alpha {=} 0$\\}}\\
    \midrule
    {\splitcell{\#Buckets}}
    & {--} 
    &{\splitcell{21}} 
    &{\splitcell{43}}
    &{\splitcell{348}} 
    &{\splitcell{200}} 
    &{\splitcell{30}} 
    &{\splitcell{33}} \\
    {\splitcell{\% Linked}}
    & 0.24 & 0.25 &0.30& {\bfseries\splitcell{0.46}} &0.27&0.36&0.33& \\
    \bottomrule
    \end{tabular}
    \label{Tab:linkage}
\end{table}

\begin{table}[t]
  \centering
  \tabcolsep=0.05cm
  \caption{Utilization of the least common $\vz$ in multiple generalizations, for 223,531 ACSEmployment records.}
  \vspace{0.5em}
  \begin{tabular}{
    S[table-format=1.2]
    S[table-format=1.2]
    S[table-format=1.2]
    S[table-format=1.2]
    S[table-format=1.2]
    S[table-format=1.2]
    S[table-format=1.2]
    S[table-format=1.2]
  }
  {\splitcell{$g$}}
    & {\small\splitcell{PAT \\ $k^* {=} 2$ \\ $\alpha {=} 0.3$}}
    & {\small\splitcell{PAT \\ $k^* {=} 4$ \\ $\alpha {=} 0.3$}}
    & {\small\splitcell{PAT \\ $k^* {=} 10$ \\ $\alpha {=} 0.3$}}
    & {\small\splitcell{PAT \\ $k^* {=} 20$ \\ $\alpha {=} 0.3$}}
    & {\small\splitcell{PAT \\ $k^* {=} 20$ \\ $\alpha {=} 0.0$}}
    & {\small\splitcell{PAT \\ $k^* {=} 20$ \\ $\alpha {=} 0.8$}}\\
  \midrule
  {\splitcell{\#Buckets}}
  &{\splitcell{17}} 
  &{\splitcell{19}}
  &{\splitcell{25}} 
  &{\splitcell{33}} 
  &{\splitcell{39}} 
  &{\splitcell{33}} \\
  {\splitcell{Utilization (\#)}}
  & {52k (1)} & {1.9k (1)} & {1 (1)} & {1 (44)} & {1 (120)}& {1 (10)} \\
  \bottomrule
  \end{tabular}
  \label{Tab:singling_out}

  \vspace{-0.5em}
  
\end{table}

\subsection{Individual Attribute Reconstruction} \label{ssec:experimental:individual}

The results for reconstruction (A1-A6) so far only report the mean over all personal attributes.
We now investigate how well an adversary can reconstruct \emph{individual} attributes.

\begin{figure}[t]
  	\centering
	\includegraphics[width=0.35\textwidth]{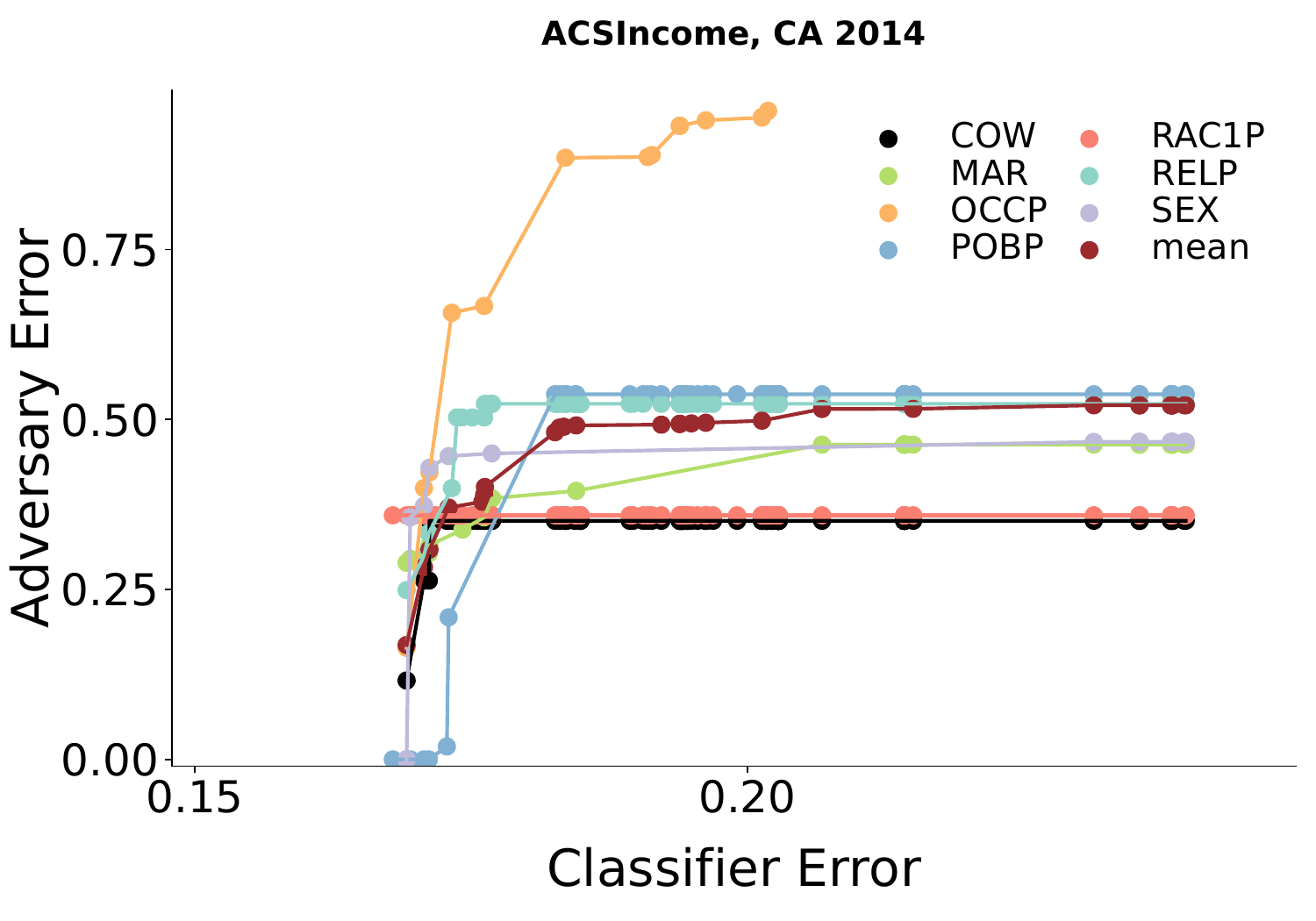}
  	\caption{Individual attribute reconstruction of the A1 adversary on ACSIncome, for points from the A1 PAT Pareto front. For many attributes the adversary can only improve over the naive baseline in very low error regimes.}
  	\label{fig:individual_plots_main}
   \vspace{-0.5em}
\end{figure}

\para{Results}
In particular, for A1, we report in \cref{fig:individual_plots_main} the individual attribute reconstruction error for all points on the PAT Pareto front for ACSIncome (we show another example in \cref{app:architectures}).
This more detailed view provides a better insight into A1's capabilities, allowing us to read out the graphs corresponding to $max$ (outer, left Pareto front) or $min$ (inner, right Pareto front) aggregators.

Overall, we observe how the adversary can only accurately reconstruct many personal attributes in low classification error regimes. For higher classification errors, the reported mean adversarial accuracy is dominated by a few attributes, which have many classes (\texttt{OCCP} has 477 classes). 

For practitioners, this can be especially interesting when their goal is to protect only specific attributes. For example, a practitioner who focuses on the place of birth (\texttt{POBP}) can choose the generalization with a classifier error of $\approx 0.18$, before the \texttt{POBP} adversarial error decreases rapidly.

\subsection{Minimizer Training Sizes} \label{ssec:experimental:sizes}
\begin{figure}[t]
  	\centering
	\includegraphics[width=0.35\textwidth]{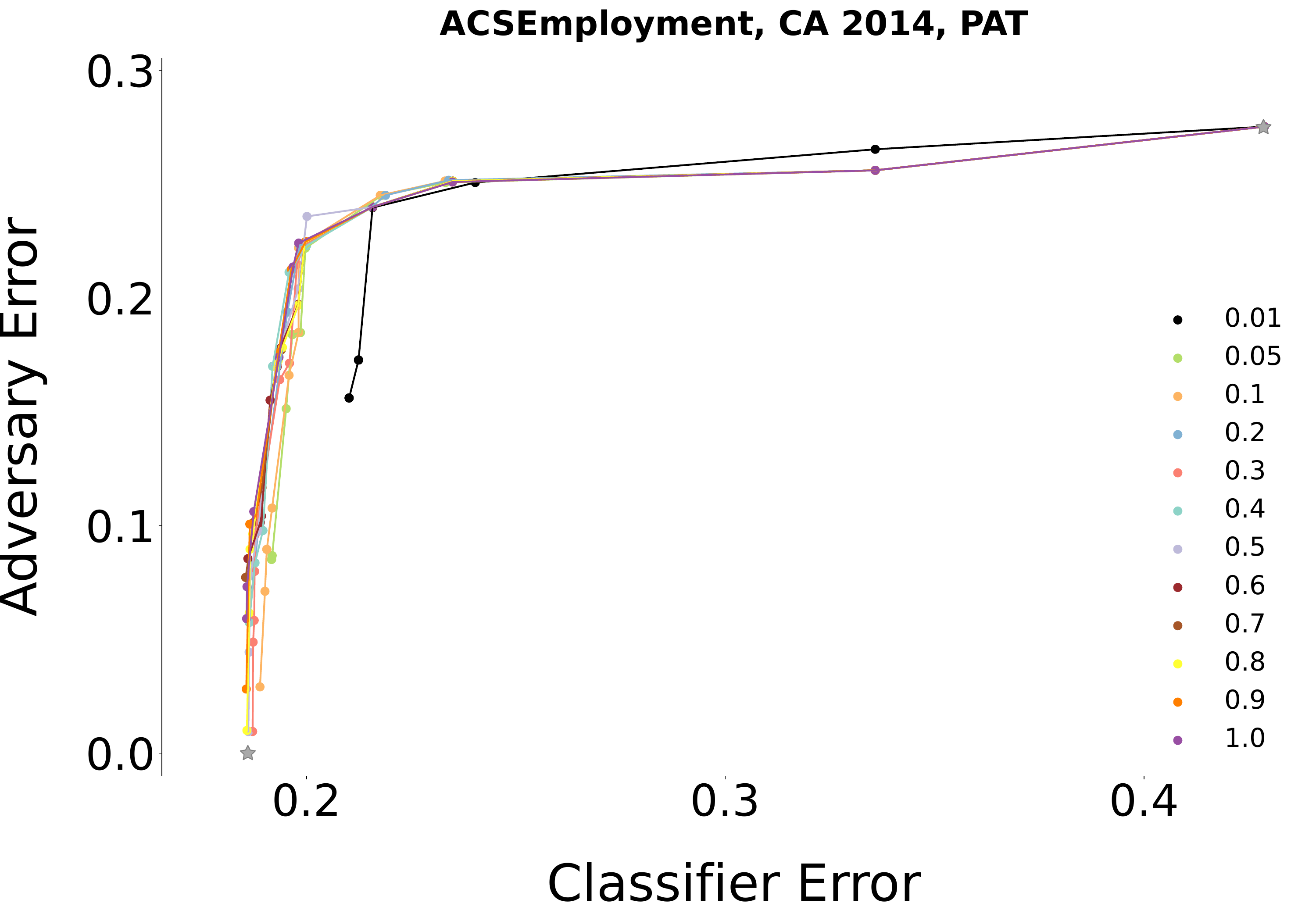}
  	\caption{Privacy-Utility Tradeoff for PAT on ACSEmployment with varying size of $S_{orig}$ (as fractions of the training set). We observe worse tradeoffs only for very small $S_{orig}$.}
  	\label{fig:min_train_sizes}
   \vspace{-0.5em}
\end{figure}

As mentioned in \cref{sec:problem}, we further observe that a small $\setOrig$ of full-granularity data is sufficient for training a minimizer. Taking, \eg \toolshort ~on ACSEmployment in \cref{fig:min_train_sizes}, we find that using only $3\%$ of the total data ($5\%$ of the training data) yields almost identical minimizer results, with performance only deteriorating for very small $\setOrig$. 

From a practitioner's point of view, this gives the advantage of only requiring small (well-protected) amounts of data to train $g$ before being able to collect minimized samples. In order to evaluate a larger set of generalizers, one might, however, increase $\setOrig$ to sizes commonly used for other PETs. Our framework then provides sensible ranges for all possible minimizer parameters, automatically tuning all adversarial hyperparameters. 
 
\begin{figure}[t]
	\centering
	\centering
	\includegraphics[width=0.4\textwidth]{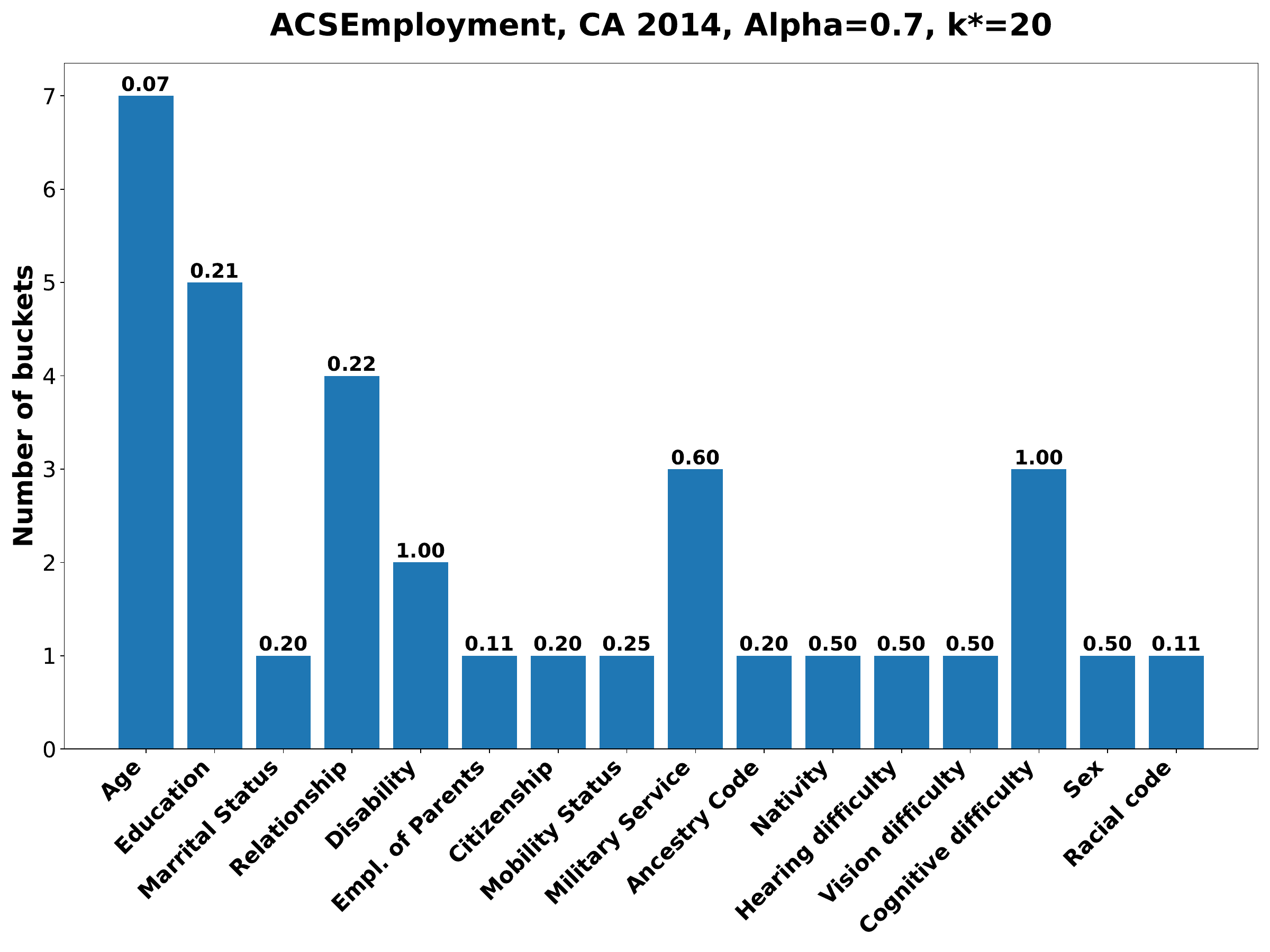}
	\vskip -0.1in
	\caption{Number of buckets for each attribute in a PAT generalization of ACSEmployment with $\alpha=0.7,k^*=20$. Above each bar, we denote the relative size of the attribute w.r.t. its original size, \eg \texttt{age} has been reduced by $93$\%.}
	\label{fig:q2}
	\vskip -0.5em
\end{figure}

\subsection{Qualitative Study} \label{ssec:experiments:qualitative}

We end our experimental evaluation with a qualitative study on one generalization. Namely, we run PAT on ACSEmployment with $k^*=20$ and $\alpha=0.7$, the same parameters as we used for our example in~\cref{fig:bait}. 

ACSEmployment has $16$ attributes with a total of $196$ attribute values (assuming \texttt{age} has integer range $[0-99]$), which are minimized to just $34$ buckets (across all attributes). This increases the classifier error by only $0.01$ while increasing the adversarial error from $0$ to $0.23$. A naive adversary which would only predict the majority value (over $\mathcal{S}'_{orig}$) for each attribute, constituting an upper bound, would have an adversarial error of $0.275$. Looking at individual attributes, \eg \texttt{SEX}, we find that we increase the adversarial error from $0$ to a nearly random $0.48$.

\para{Per-attribute Generalizations} In \cref{fig:q2}, we explore the values of $k_i$ for each attribute. We find that (1) the attributes that one would generally relate with the employment status (\eg \texttt{Age}, \texttt{Educational Status}) are less generalized than other attributes (\eg \texttt{Sex}). $10$ out of $16$ attributes (including highly sensitive ones as \texttt{Ancestry} and \texttt{Race}) can be fully generalized and hence would not even need to be collected. This suggests that client privacy could be greatly improved by deploying vDM in real-world scenarios.

Looking closer at some attributes we find that PAT groups all persons with \texttt{age} 20-63 into a single bucket, while the generalization of the \texttt{Educational Status} attribute can be roughly summarized as [No high school diploma, Highschool diploma, Assoc. Degree, Masters, Ph.D.] instead of 24 individual categories.
\section{Future Work} \label{sec:limitations}

We believe that due to an increase in regulation, data minimization will be a highly relevant topic for future research. We see four key areas for future work on vDM.

\para{Combining vDM and Other ML Privacy Topics}
This includes combining vDM with topics such as differential privacy \cite{AbadiCGMMT016} and secure computation \cite{GiladDLLNW16, LiuJLA17, DathathriSCLLMMM19}. As vDM stands orthogonal to many of these techniques, we believe their combination can be particularly interesting. Some of our early investigations already indicate that minimized data might be well-suited for common DP training procedures. An additional area of interest is the interplay between vertical and horizontal data minimization.
 
\para{Other Domains}
With most personal attributes (\eg religion, age, political affiliation) in tabular form, the application of generalizations for privacy protection has so far been focused on tabular data. Adapting such methods to, \eg images or text is a promising avenue for future work.

\para{Privacy Guarantees}
Further enhancing our privacy risk assessment with formal guarantees on the utility-fairness tradeoffs is another important direction that would benefit the practitioners. 
This is a challenging problem, and it is unclear what kind of approach would be able to provide such guarantees.
One potentially promising direction could be to enforce a distribution-wide lower bound on the utilization of each generalization bucket, and use this to bound the adversarial risk following the approach of \citet{jovanovic2023fare} that offer similar guarantees in the setting of fair representation learning. 

\para{New vDM Algorithms}
While PAT outperforms all baselines across a variety of settings, it does not come with an optimality guarantee. This leaves the field of vDM open for future algorithms with better utility-privacy tradeoffs.
  
\section{Related Work} \label{sec:related}

\subsection{Regulations and Policy}
The requirement of data minimization, which limits the collection and processing of personal data to the minimum necessary for a specific purpose, was introduced in 2016 in the E.U.s GDPR ~\cite{gdpr}. Similar principles have been adapted and integrated into other regulations, such as CPRA~\cite{cpra} and the recent Blueprint for a U.S. AI Bill of Rights~\cite{whitehouse2022bill}, emphasizing its relevance. While data protection authorities in certain countries (\eg the U.K.'s ICO~\cite{icouk} and the Norwegian Data Protection Authority~\cite{norway2018ai}) have proposed some guidelines for complying with DM in ML contexts, and there exists policy literature analyzing similar issues~\cite{finck2021reviving}, to date, no concrete evaluation tools have been proposed.

\subsection{Generalizations in ML}
Besides the already mentioned PPDP use-case, the idea of generalizing attributes to coarser representations has been applied in several ML settings under different names: \emph{Discretization} is used to combine similar attribute values in the context of recommender systems \cite{Guo,discret2}. \emph{Binning} is used in, \eg \citet{binning1,Palencia} to derive concrete attribute values out of histogram data. It is worth noting that suppression and feature selection are special cases of generalization.

\subsection{Operationalization of DM for ML} \label{ssec:related:op}
We now discuss prior attempts to operationalize DM for ML.
\citet{raste} discusses a \emph{need-to-know} principle which is similar to DM, but focuses primarily on fairness and does not touch on the generalization of attributes beyond attribute selection.
\citet{Biega} explicitly considers a formalization of DM tailored to recommender systems with techniques that are not applicable to our vDM for ML setting.
\citet{raste2} proposes algorithms for black-box auditing of DM compliance based on model instability. This setting differs from ours as it focuses on auditing an already trained model.
\citet{shanmu} focuses on minimizing training data size (\emph{hDM}), an important concern even outside DM \cite{mirzasoleiman, paul} that stands orthogonal to \emph{vDM}.%

The most closely related work to ours is \citet{IbmApt}, which applies concepts from data anonymization (discussed in \cref{sec:setting}) to propose a vDM minimizer. However, as shown in \cref{ssec:background:generalization}, the metrics from data anonymization \cite{Ghinita} do not translate well into the vDM setting. Additionally, PAT outperforms the \citet{IbmApt} approach both in terms of speed ($100$x) and in the utility-privacy tradeoff.
Crucially, no prior work defines comprehensive evaluation procedures, baselines, and ways to evaluate empirical privacy risks under different adversaries, which we argue is a key issue for vDM.

\subsection{Fair Representation Learning}

Fair representation learning (FRL)~\cite{zemel2013learning,MadrasAdvTrain,gupta2021controllable,balunovic2022fair} transforms data into a new representation useful for downstream tasks while ensuring that one cannot recover the sensitive attribute. The most common approaches are typically based on adversarial training~\cite{MadrasAdvTrain}, VAE~\cite{louizos2016vae}, mutual information~\cite{song2019controllable, gupta2021controllable} and normalizing flows~\cite{balunovic2022fair}.
While FRL partly inspired our approach for PAT in \cref{sec:algorithm}, FRL and DM consider different scenarios and have two key technical differences: (i) In FRL, it is always necessary to collect full-granularity data to produce the representation, as the transformation is non-interpretable (\ie cannot be easily applied by a client), and (ii) similar to work in data anonymization (see \cref{ssec:background:data}) FRL usually considers a single sensitive attribute, while we relax this constraint.

\subsection{Causal Feature Selection}
Another line of work studies causal feature selection~\cite{aliferis10a, YuLLDL20, tople, hasanfritz}.
Some works in particular investigate directions related to vDM such as connections to privacy~\cite{tople} or the design of FRL-inspired methods that offer interpretability~\cite{hasanfritz}. 
While suitable for certain scenarios, feature selection methods only have a binary choice for each feature. This severely limits their available transformations in comparison to tailor-made minimizers in the vDM setting, as we demonstrate in \cref{sec:experimental} and \cref{app:details:moreresults}.

\section{Conclusion} \label{sec:conclusion}

We have addressed the challenge of formalizing and achieving vertical data minimization, a highly relevant privacy requirement, in the context of ML.
We formalized the vDM setting and workflow via generalizations, defined two key requirements of utility and empirical privacy risk, and proposed a set of diverse adversaries as tools for empirical privacy risk evaluation.
We introduced several baseline vDM minimizers, as well as the \emph{Privacy-aware Tree} (PAT) minimizer, whose effectiveness we demonstrated on several real-world datasets.
Our hope is that our work and public release of the vDM toolbox will enable organizations to more effectively enforce and evaluate data minimization, ultimately helping reduce the privacy risks for individuals.

\section*{Acknowledgments}
We thank Matthew Jagielski for helpful discussions in early stages of the project. We are grateful to anonymous reviewers for their constructive feedback, and the S\&P organizing committee and our shepherd for facilitating a quality review process. This work has received funding from the Swiss State Secretariat for Education, Research and Innovation (SERI) (SERI-funded ERC Consolidator Grant).

\IEEEpeerreviewmaketitle

\bibliographystyle{IEEEtran}
\bibliography{references}

\begin{thebibliography}{10}
\providecommand{\url}[1]{#1}
\csname url@samestyle\endcsname
\providecommand{\newblock}{\relax}
\providecommand{\bibinfo}[2]{#2}
\providecommand{\BIBentrySTDinterwordspacing}{\spaceskip=0pt\relax}
\providecommand{\BIBentryALTinterwordstretchfactor}{4}
\providecommand{\BIBentryALTinterwordspacing}{\spaceskip=\fontdimen2\font plus
\BIBentryALTinterwordstretchfactor\fontdimen3\font minus
  \fontdimen4\font\relax}
\providecommand{\BIBforeignlanguage}[2]{{%
\expandafter\ifx\csname l@#1\endcsname\relax
\typeout{** WARNING: IEEEtran.bst: No hyphenation pattern has been}%
\typeout{** loaded for the language `#1'. Using the pattern for}%
\typeout{** the default language instead.}%
\else
\language=\csname l@#1\endcsname
\fi
#2}}
\providecommand{\BIBdecl}{\relax}
\BIBdecl

\bibitem{khandani2010consumer}
A.~E. Khandani, A.~J. Kim, and A.~W. Lo, ``Consumer credit-risk models via
  machine-learning algorithms,'' \emph{JBF}, 2010.

\bibitem{perols2011financial}
J.~Perols, ``Financial statement fraud detection: An analysis of statistical
  and machine learning algorithms,'' \emph{Auditing: A Journal of Practice \&
  Theory}, 2011.

\bibitem{gdpr}
\BIBentryALTinterwordspacing
E.~U. EU, ``General data protection regulation,'' 2016. [Online]. Available:
  \url{https://gdpr-info.eu/}
\BIBentrySTDinterwordspacing

\bibitem{cpra}
\BIBentryALTinterwordspacing
C.~S. o.~S. CAGOV, ``California privacy rights act,'' 2020. [Online].
  Available: \url{https://vig.cdn.sos.ca.gov/2020/general/pdf/topl-prop24.pdf}
\BIBentrySTDinterwordspacing

\bibitem{whitehouse2022bill}
\BIBentryALTinterwordspacing
USGOV, ``\BIBforeignlanguage{en-US}{Blueprint for an {AI} {Bill} of {Rights}
  {\textbar} {OSTP}},'' 2022. [Online]. Available:
  \url{https://www.whitehouse.gov/ostp/ai-bill-of-rights/}
\BIBentrySTDinterwordspacing

\bibitem{gdprtracker}
\BIBentryALTinterwordspacing
CMS, ``Gdpr enforcement tracker,'' 2018. [Online]. Available:
  \url{https://www.enforcementtracker.com/}
\BIBentrySTDinterwordspacing

\bibitem{dutchgdpr}
GDPRhub, ``Ap (the netherlands) - tax administration fined for discriminatory
  and unlawful data processing,'' 2023,
  \url{https://gdprhub.eu/index.php?title=AP_(The_Netherlands)_-_Tax_Administration_fined_for_discriminatory_and_unlawful_data_processing}.

\bibitem{gdpr790}
\BIBentryALTinterwordspacing
CMS, ``Gdpr enforcement tracker: Etid-790,'' 2021. [Online]. Available:
  \url{https://www.enforcementtracker.com/ETid-790}
\BIBentrySTDinterwordspacing

\bibitem{norway2018ai}
\BIBentryALTinterwordspacing
N.~D. P.~A. DPA, ``Artificial intelligence and privacy,'' 2018. [Online].
  Available:
  \url{https://www.datatilsynet.no/globalassets/global/english/ai-and-privacy.pdf}
\BIBentrySTDinterwordspacing

\bibitem{shanmu}
D.~{Shanmugam}, S.~{Shabanian}, F.~{Diaz}, M.~{Finck}, and A.~{Biega},
  ``Learning to limit data collection via scaling laws: Data minimization
  compliance in practice,'' \emph{ACM FAccT}, 2022.

\bibitem{mirzasoleiman}
B.~Mirzasoleiman, J.~A. Bilmes, and J.~Leskovec, ``Coresets for data-efficient
  training of machine learning models,'' in \emph{ICML}, 2020.

\bibitem{paul}
M.~Paul, S.~Ganguli, and G.~K. Dziugaite, ``Deep learning on a data diet:
  Finding important examples early in training,'' 2021.

\bibitem{jovanovic2023fare}
N.~Jovanović, M.~Balunović, D.~I. Dimitrov, and M.~Vechev, ``Fare: Provably
  fair representation learning with practical certificates,'' \emph{ICML},
  2023.

\bibitem{grigorescu2020survey}
S.~Grigorescu, B.~Trasnea, T.~Cocias, and G.~Macesanu, ``A survey of deep
  learning techniques for autonomous driving,'' \emph{JFR}, 2020.

\bibitem{chen2018rise}
H.~Chen, O.~Engkvist, Y.~Wang, M.~Olivecrona, and T.~Blaschke, ``The rise of
  deep learning in drug discovery,'' \emph{Drug discovery today}, 2018.

\bibitem{lefevre2006mondrian}
K.~LeFevre, D.~J. DeWitt, and R.~Ramakrishnan, ``Mondrian multidimensional
  k-anonymity,'' in \emph{IEEE ICDE}, 2006.

\bibitem{xu2006utility}
J.~Xu, W.~Wang, J.~Pei, X.~Wang, B.~Shi, and A.~W.-C. Fu, ``Utility-based
  anonymization using local recoding,'' in \emph{ACM SIGKDD}, 2006.

\bibitem{Ghinita}
G.~Ghinita, P.~Karras, P.~Kalnis, and N.~Mamoulis, ``Fast data anonymization
  with low information loss,'' in \emph{VLDB}, 2007.

\bibitem{IbmApt}
A.~Goldsteen, G.~Ezov, R.~Shmelkin, M.~Moffie, and A.~Farkash, ``Data
  minimization for {GDPR} compliance in machine learning models,'' \emph{AI and
  Ethics}, 2022.

\bibitem{GDPR3Attacks}
\BIBentryALTinterwordspacing
A.~. D. P.~W. Party, ``Opinion 05/2014 on anonymisation techniques,'' 2014.
  [Online]. Available:
  \url{https://ec.europa.eu/justice/article-29/documentation/opinion-recommendation/files/2014/wp216_en.pdf}
\BIBentrySTDinterwordspacing

\bibitem{piius}
\BIBentryALTinterwordspacing
U.~D. of~Labor, ``Guidance on the protection of personal identifiable
  information.'' [Online]. Available: \url{https://www.dol.gov/general/ppii}
\BIBentrySTDinterwordspacing

\bibitem{fung2010privacy}
B.~C. Fung, K.~Wang, R.~Chen, and P.~S. Yu, ``Privacy-preserving data
  publishing: A survey of recent developments,'' \emph{ACM Computing Surveys},
  2010.

\bibitem{han2013sloms}
J.~Han, F.~Luo, J.~Lu, and H.~Peng, ``Sloms: A privacy preserving data
  publishing method for multiple sensitive attributes microdata.''
  \emph{Journal of Software}, 2013.

\bibitem{anjum2018efficient}
A.~Anjum, N.~Ahmad, S.~U. Malik, S.~Zubair, and B.~Shahzad, ``An efficient
  approach for publishing microdata for multiple sensitive attributes,''
  \emph{The Journal of Supercomputing}, 2018.

\bibitem{balunovic2022fair}
M.~Balunovic, A.~Ruoss, and M.~Vechev, ``Fair normalizing flows,'' in
  \emph{ICLR}, 2022.

\bibitem{noauthor_prohibited_nodate}
\BIBentryALTinterwordspacing
``\BIBforeignlanguage{en}{Prohibited {Employment} {Policies}/{Practices}}.''
  [Online]. Available:
  \url{https://www.eeoc.gov/prohibited-employment-policiespractices}
\BIBentrySTDinterwordspacing

\bibitem{MIA}
R.~Shokri, M.~Stronati, C.~Song, and V.~Shmatikov, ``Membership inference
  attacks against machine learning models,'' in \emph{IEEE S\&P}, 2017.

\bibitem{MIASOK}
A.~Salem, G.~Cherubin, D.~Evans, B.~Kopf, A.~Paverd, A.~Suri, S.~Tople, and
  S.~Zanella-Beguelin, ``Sok: Let the privacy games begin! a unified treatment
  of data inference privacy in machine learning,'' in \emph{IEEE S\&P}, 2023.

\bibitem{9084352}
T.~Li, A.~K. Sahu, A.~Talwalkar, and V.~Smith, ``Federated learning:
  Challenges, methods, and future directions,'' \emph{IEEE Signal Processing
  Magazine}, 2020.

\bibitem{balunovic2021bayesian}
M.~Balunovi{\'c}, D.~I. Dimitrov, R.~Staab, and M.~Vechev, ``Bayesian framework
  for gradient leakage,'' \emph{ICLR}, 2022.

\bibitem{dwork2006differential}
C.~Dwork, ``Differential privacy,'' in \emph{ICALP}, 2006.

\bibitem{ye_local_2020}
Q.~Ye and H.~Hu, ``Local {Differential} {Privacy}: {Tools}, {Challenges}, and
  {Opportunities},'' in \emph{Web {Information} {Systems} {Engineering}}, 2020.

\bibitem{abadi2016deep}
M.~Abadi, A.~Chu, I.~Goodfellow, H.~B. McMahan, I.~Mironov, K.~Talwar, and
  L.~Zhang, ``Deep learning with differential privacy,'' in \emph{ACM CCS},
  2016.

\bibitem{fhe_gentry}
C.~Gentry, ``\BIBforeignlanguage{English}{A fully homomorphic encryption
  scheme},'' Ph.D. dissertation, 2009.

\bibitem{chillotti_new_nodate}
I.~Chillotti, M.~Joye, and P.~Paillier, ``\BIBforeignlanguage{en}{New
  {Challenges} for {Fully} {Homomorphic} {Encryption}}.''

\bibitem{hernandez_marcano_fully_2019}
N.~J. Hernandez~Marcano, M.~Moller, S.~Hansen, and R.~H. Jacobsen, ``On {Fully}
  {Homomorphic} {Encryption} for {Privacy}-{Preserving} {Deep} {Learning},'' in
  \emph{{IEEE} {Globecom} {Workshops}}, 2019.

\bibitem{lee_privacy-preserving_2022}
J.-W. Lee, H.~Kang, Y.~Lee, W.~Choi, J.~Eom, M.~Deryabin, E.~Lee, J.~Lee,
  D.~Yoo, Y.-S. Kim, and J.-S. No, ``Privacy-{Preserving} {Machine} {Learning}
  {With} {Fully} {Homomorphic} {Encryption} for {Deep} {Neural} {Network},''
  2022.

\bibitem{cramer2015secure}
R.~Cramer, I.~B. Damg{\aa}rd \emph{et~al.}, \emph{Secure multiparty
  computation}.\hskip 1em plus 0.5em minus 0.4em\relax Cambridge University
  Press, 2015.

\bibitem{knott2021crypten}
B.~Knott, S.~Venkataraman, A.~Hannun, S.~Sengupta, M.~Ibrahim, and L.~van~der
  Maaten, ``Crypten: Secure multi-party computation meets machine learning,''
  \emph{NeurIPS}, 2021.

\bibitem{juvekar2018gazelle}
C.~Juvekar, V.~Vaikuntanathan, and A.~Chandrakasan, ``Gazelle: A low latency
  framework for secure neural network inference,'' in \emph{USENIX Security},
  2018.

\bibitem{mishra2020delphi}
P.~Mishra, R.~Lehmkuhl, A.~Srinivasan, W.~Zheng, and R.~A. Popa, ``Delphi: A
  cryptographic inference system for neural networks,'' in \emph{Workshop on
  Privacy-Preserving Machine Learning in Practice}, 2020.

\bibitem{Sweene02}
L.~Sweeney, ``k-anonymity: {A} model for protecting privacy,'' 2002.

\bibitem{machana}
A.~Machanavajjhala, J.~Gehrke, D.~Kifer, and M.~Venkitasubramaniam,
  ``l-diversity: Privacy beyond k-anonymity,'' in \emph{IEEE ICDE}, 2006.

\bibitem{li2006t}
N.~Li, T.~Li, and S.~Venkatasubramanian, ``t-closeness: Privacy beyond
  k-anonymity and l-diversity,'' in \emph{IEEE ICDE}, 2006.

\bibitem{he2012permutation}
X.~He, Y.~Xiao, Y.~Li, Q.~Wang, W.~Wang, and B.~Shi, ``Permutation
  anonymization: Improving anatomy for privacy preservation in data
  publication,'' in \emph{New Frontiers in Applied Data Mining: PAKDD
  International Workshops}, 2012.

\bibitem{xu2018synthesizing}
L.~Xu and K.~Veeramachaneni, ``Synthesizing tabular data using generative
  adversarial networks,'' \emph{arXiv preprint arXiv:1811.11264}, 2018.

\bibitem{stadler2022synthetic}
T.~Stadler, B.~Oprisanu, and C.~Troncoso, ``Synthetic data--anonymisation
  groundhog day,'' in \emph{USENIX Security}, 2022.

\bibitem{chen_training_2020}
Y.~Chen, F.~Luo, T.~Li, T.~Xiang, Z.~Liu, and J.~Li, ``A training-integrity
  privacy-preserving federated learning scheme with trusted execution
  environment,'' \emph{Information Sciences}, vol. 522, 2020.

\bibitem{decentriq}
\BIBentryALTinterwordspacing
Decentriq, ``Future-proof {Data} {Clean} {Rooms} {\textbar} {Decentriq},''
  2023. [Online]. Available: \url{https://decentriq.com/}
\BIBentrySTDinterwordspacing

\bibitem{ModelInv}
M.~Fredrikson, S.~Jha, and T.~Ristenpart, ``Model inversion attacks that
  exploit confidence information and basic countermeasures,'' in \emph{{ACM
  CCS}}, 2015.

\bibitem{giomi2023privacysynthetic}
\BIBentryALTinterwordspacing
M.~Giomi, F.~Boenisch, C.~Wehmeyer, and B.~Tasnádi, ``A unified framework for
  quantifying privacy risk in synthetic data,'' 2022. [Online]. Available:
  \url{https://arxiv.org/abs/2211.10459}
\BIBentrySTDinterwordspacing

\bibitem{auxdata}
I.~E. Olatunji, J.~Rauch, M.~Katzensteiner, and M.~Khosla, ``A review of
  anonymization for healthcare data,'' \emph{CoRR}, 2021.

\bibitem{linkability}
A.~Narayanan and V.~Shmatikov, ``Robust de-anonymization of large sparse
  datasets,'' in \emph{{IEEE} S\&P}, 2008.

\bibitem{sideChannelFranziska}
F.~Boenisch, R.~Munz, M.~Tiepelt, S.~Hanisch, C.~Kuhn, and P.~Francis,
  ``Side-channel attacks on query-based data anonymization,'' in \emph{{ACM
  CCS}}, 2021.

\bibitem{deepSinglingOut}
A.~Cohen and K.~Nissim, ``Towards formalizing the gdpr's notion of singling
  out,'' \emph{Proc. Natl. Acad. Sci. {USA}}, 2020.

\bibitem{dingledine2006anonymity}
R.~Dingledine and N.~Mathewson, ``Anonymity loves company: Usability and the
  network effect.'' in \emph{WEIS}, 2006.

\bibitem{mono1}
J.~Sill, ``Monotonic networks,'' in \emph{NIPS}, 1997.

\bibitem{mono2}
X.~Liu, X.~Han, N.~Zhang, and Q.~Liu, ``Certified monotonic neural networks,''
  in \emph{NeurIPS}, 2020.

\bibitem{MadrasAdvTrain}
D.~Madras, E.~Creager, T.~Pitassi, and R.~S. Zemel, ``Learning adversarially
  fair and transferable representations,'' in \emph{ICML}, 2018.

\bibitem{Gans}
I.~J. Goodfellow, J.~Pouget{-}Abadie, M.~Mirza, B.~Xu, D.~Warde{-}Farley,
  S.~Ozair, A.~C. Courville, and Y.~Bengio, ``Generative adversarial nets,'' in
  \emph{NeurIPS}, 2014.

\bibitem{zhang_faht_2019}
W.~Zhang and E.~Ntoutsi, ``{FAHT}: {An} {Adaptive} {Fairness}-aware {Decision}
  {Tree} {Classifier},'' in \emph{IJCAI}, 2019.

\bibitem{RetiringAdult}
F.~Ding, M.~Hardt, J.~Miller, and L.~Schmidt, ``Retiring adult: New datasets
  for fair machine learning,'' in \emph{NeurIPS}, 2021.

\bibitem{noauthor_heritage_nodate}
\BIBentryALTinterwordspacing
Kaggle, ``\BIBforeignlanguage{en}{Heritage {Health} {Prize}},'' 2012. [Online].
  Available: \url{https://kaggle.com/competitions/hhp}
\BIBentrySTDinterwordspacing

\bibitem{xu_modeling_2019}
\BIBentryALTinterwordspacing
L.~Xu, M.~Skoularidou, A.~Cuesta-Infante, and K.~Veeramachaneni,
  ``\BIBforeignlanguage{en}{Modeling {Tabular} data using {Conditional}
  {GAN}},'' Oct. 2019, arXiv:1907.00503 [cs, stat]. [Online]. Available:
  \url{http://arxiv.org/abs/1907.00503}
\BIBentrySTDinterwordspacing

\bibitem{AbadiCGMMT016}
M.~Abadi, A.~Chu, I.~J. Goodfellow, H.~B. McMahan, I.~Mironov, K.~Talwar, and
  L.~Zhang, ``Deep learning with differential privacy,'' in \emph{{ACM CCS}},
  2016.

\bibitem{GiladDLLNW16}
R.~Gilad-Bachrach, N.~Dowlin, K.~Laine, K.~Lauter, M.~Naehrig, and J.~Wernsing,
  ``Cryptonets: Applying neural networks to encrypted data with high throughput
  and accuracy,'' in \emph{ICML}, 2016.

\bibitem{LiuJLA17}
J.~Liu, M.~Juuti, Y.~Lu, and N.~Asokan, ``Oblivious neural network predictions
  via minionn transformations,'' in \emph{{ACM CCS}}, 2017.

\bibitem{DathathriSCLLMMM19}
R.~Dathathri, O.~Saarikivi, H.~Chen, K.~Laine, K.~Lauter, S.~Maleki,
  M.~Musuvathi, and T.~Mytkowicz, ``{CHET: An Optimizing Compiler for
  Fully-Homomorphic Neural-Network Inferencing},'' in \emph{ACM PLDI}, 2019.

\bibitem{icouk}
\BIBentryALTinterwordspacing
U.~I. C.~O. UKICO, ``Guide to the general data protection regulations (gdpr),''
  2018. [Online]. Available:
  \url{https://ico.org.uk/for-organisations/guide-to-data-protection/guide-to-the-general-data-protection\\-regulation-gdpr/}
\BIBentrySTDinterwordspacing

\bibitem{finck2021reviving}
M.~Finck and A.~J. Biega, ``Reviving purpose limitation and data minimisation
  in data-driven systems,'' \emph{Technology and Regulation}, 2021.

\bibitem{Guo}
H.~Guo, B.~Chen, R.~Tang, W.~Zhang, Z.~Li, and X.~He, ``An embedding learning
  framework for numerical features in {CTR} prediction,'' in \emph{KDD}, 2021.

\bibitem{discret2}
Y.~Qu, B.~Fang, W.~Zhang, R.~Tang, M.~Niu, H.~Guo, Y.~Yu, and X.~He,
  ``Product-based neural networks for user response prediction over multi-field
  categorical data,'' \emph{{ACM} Trans. Inf. Syst.}, 2019.

\bibitem{binning1}
U.~M. Fayyad and K.~B. Irani, ``Multi-interval discretization of
  continuous-valued attributes for classification learning,'' in \emph{IJCAI},
  1993.

\bibitem{Palencia}
G.~Navas{-}Palencia, ``Optimal binning: mathematical programming formulation,''
  2020.

\bibitem{raste}
B.~Rastegarpanah, M.~Crovella, and K.~P. Gummadi, ``Fair inputs and fair
  outputs: The incompatibility of fairness in privacy and accuracy,'' in
  \emph{UMAP}, 2020.

\bibitem{Biega}
A.~J. Biega, P.~Potash, H.~D. III, F.~Diaz, and M.~Finck, ``Operationalizing
  the legal principle of data minimization for personalization,'' in \emph{ACM
  SIGIR}, 2020.

\bibitem{raste2}
B.~Rastegarpanah, K.~P. Gummadi, and M.~Crovella, ``Auditing black-box
  prediction models for data minimization compliance,'' in \emph{NeurIPS},
  2021.

\bibitem{zemel2013learning}
R.~S. Zemel, Y.~Wu, K.~Swersky, T.~Pitassi, and C.~Dwork, ``Learning fair
  representations,'' in \emph{ICML}, 2013.

\bibitem{gupta2021controllable}
U.~Gupta, A.~Ferber, B.~Dilkina, and G.~V. Steeg, ``Controllable guarantees for
  fair outcomes via contrastive information estimation,'' \emph{AAAI}, 2021.

\bibitem{louizos2016vae}
C.~Louizos, K.~Swersky, Y.~Li, M.~Welling, and R.~S. Zemel, ``The variational
  fair autoencoder,'' in \emph{ICLR}, 2016.

\bibitem{song2019controllable}
J.~Song, P.~Kalluri, A.~Grover, S.~Zhao, and S.~Ermon, ``Learning controllable
  fair representations,'' in \emph{AISTATS}, 2019.

\bibitem{aliferis10a}
C.~F. Aliferis, A.~Statnikov, I.~Tsamardinos, S.~Mani, and X.~D. Koutsoukos,
  ``Local causal and markov blanket induction for causal discovery and feature
  selection for classification part i: Algorithms and empirical evaluation,''
  \emph{JMLR}, 2010.

\bibitem{YuLLDL20}
K.~Yu, L.~Liu, J.~Li, W.~Ding, and T.~D. Le, ``Multi-source causal feature
  selection,'' \emph{{IEEE} TPAMI}, 2020.

\bibitem{tople}
S.~Tople, A.~Sharma, and A.~Nori, ``Alleviating privacy attacks via causal
  learning,'' in \emph{ICML}, 2020.

\bibitem{hasanfritz}
R.~Hasan and M.~Fritz, ``Understanding utility and privacy of demographic data
  in education technology by causal analysis and adversarial-censoring,''
  \emph{PoPETs}, vol. 2022.

\bibitem{loan}
\BIBentryALTinterwordspacing
``Lending club loan data.'' [Online]. Available:
  \url{https://www.kaggle.com/datasets/wordsforthewise/lending-club}
\BIBentrySTDinterwordspacing

\bibitem{uci}
\BIBentryALTinterwordspacing
M.~Kelly, R.~Longjohn, and K.~Nottingham, ``{The} {UCI} {Machine} {Learning}
  {Repository}.'' [Online]. Available: \url{https://archive.ics.uci.edu}
\BIBentrySTDinterwordspacing

\end{thebibliography}

\crefalias{section}{appendix}
\appendices
\renewcommand{\thesubsection}{\alph{subsection}}

\section{More Details on Adversaries} \label{app:adversaries}

\para{Masking in A1-A6 (Reconstruction)}
For any reconstruction adversary, let $h(\vz)_p$ denote the logits produced when reconstructing attribute $p$ from $\vz$. As $h$ is implemented as a neural network, the co-domain (range) of $h$ is the cardinality (number of classes) of $p$. However, as we assume that the adversary has access to the generalization function $g$, it can use this information to mask out all logits in $h(\vz)_p$ that correspond to classes in $\vx_p$ that g could not have mapped to $\vz_p$. This can significantly limit the co-domain for specific reconstructions, resulting in a stronger adversary. We apply the same masking procedure for all reconstruction adversaries. In particular, A6 intersects the masks corresponding to the different generalizations.

\para{Details on A7 (Linkability)}
Here we provide further detail on how we approximate $p(\vx_{A} \mid \vx_{B}=\vb)$ when knowing $\advMin$. For this, let $A, B \subseteq [1,...,d]$ be any two (not necessarily disjoint) subsets of our attributes.
Let $g_{A}(\vx) = g(\vx)_{A}$ denote the generalized attributes of $\vx$ that are also in $A$ and let $g_{A}(\va)$ denote the evaluation of $g$ on a vector $\va$ which only contains attributes in $A$ (this is well-defined as $g$ maps attributes independently of each other). We can now get an approximation of $p(\vx_{A} \mid \vx_{B}=\vb)$ by simply observing the relative sampling frequencies of $\vx_{A}$ conditioned on $\vx_{B}=\vb$ (over $\advMin$). In particular, we find that $p(x_A=\va\mid x_B=\vb) \approx \frac{|\{z\in S'_{min} \mid z_{A\cup B} = g_{A\cup B}(\va,\vb)\}|}{|\{z \in S'_{min} \mid z_{B} = g_B(\vb)\}|}$.

We note that this is only well-defined if $|\{z \in S'_{min} \mid z_{B} = g_B(\vb)\}| \geq 1$ which we assume to hold for our cases.

\para{Details on A8 (Singling-out)}
To more formally describe the A8 adversary, we must model $\advMin$ as a multiset. In particular, let $X_{\text{min}}$ denote the set of full granularity records used to create $\advMin$ ($g(X_{\text{min}}) = \advMin$). We now define the cardinality for any $z \in \advMin$ as $|z| = |\{x \in X_{\text{min}} \mid g(x) = z\}|$ and additionally $z^{\text{min}} = \argmin_{z\in \advMin} |z|.$ In case $z^{\text{min}} = 1$ we can single out the unique $\vx \in X_{\text{min}}$ for which $g(\vx) = z^{\text{min}}$ via $\Pi_{\vx}$, which describes the generalized attributes of $\vz^{\text{min}}$. 
In case $z^{\text{min}} > 1$, an adversary still can construct $\Pi_{z^{\text{min}}}$ to target $z^{\text{min}}$. Conditioned on $\Pi_{z^{\text{min}}}$, records $\vx$ which map to $z^{\text{min}}$ have a significantly smaller anonymity set making it easier for an adversary to single an individual.

\begin{figure}[t]
  \centering
  \hspace*{-1em}
  \begin{tabular}{cc}
  \includegraphics[width=0.24\textwidth]{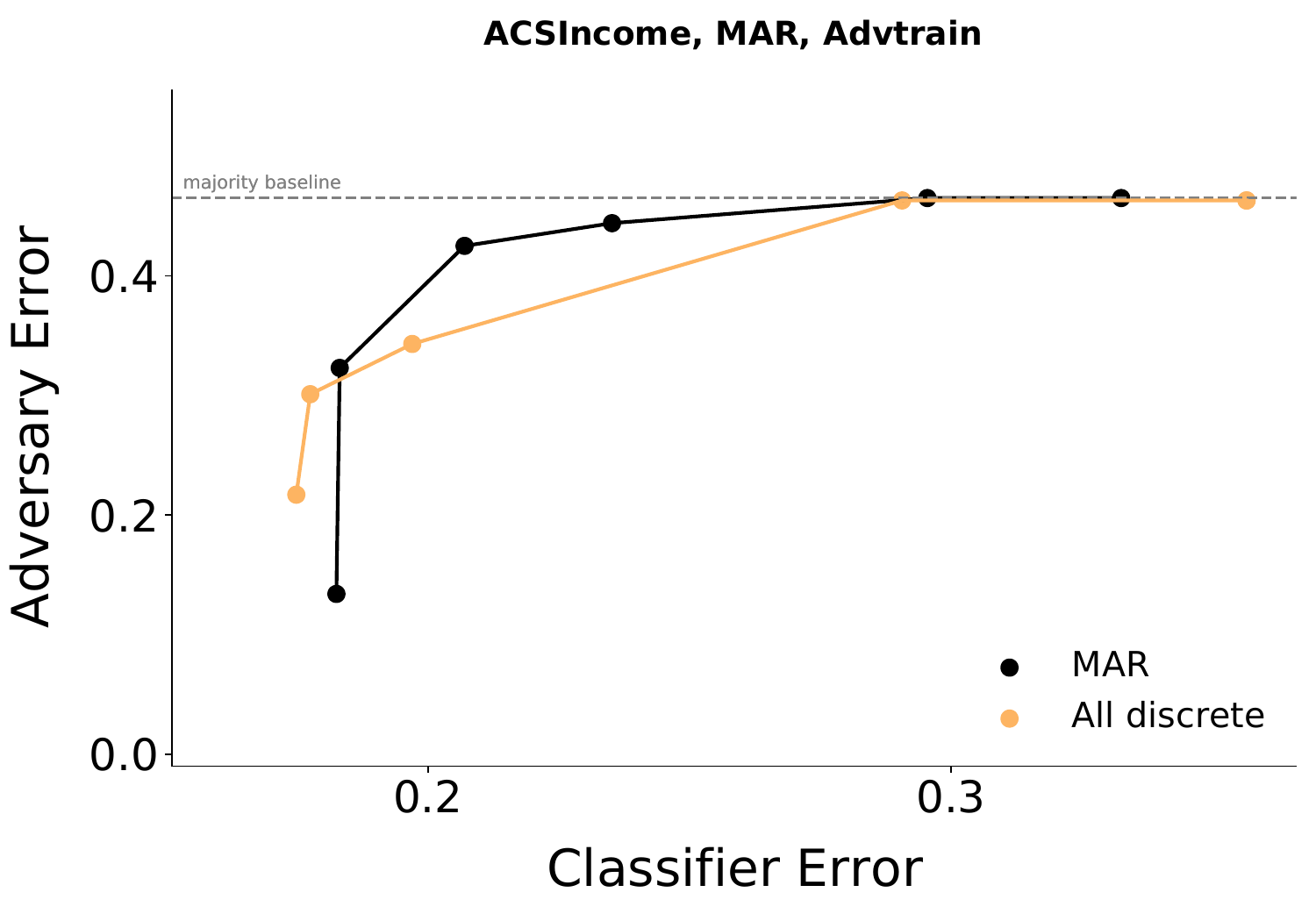}  & \includegraphics[width=0.24\textwidth]{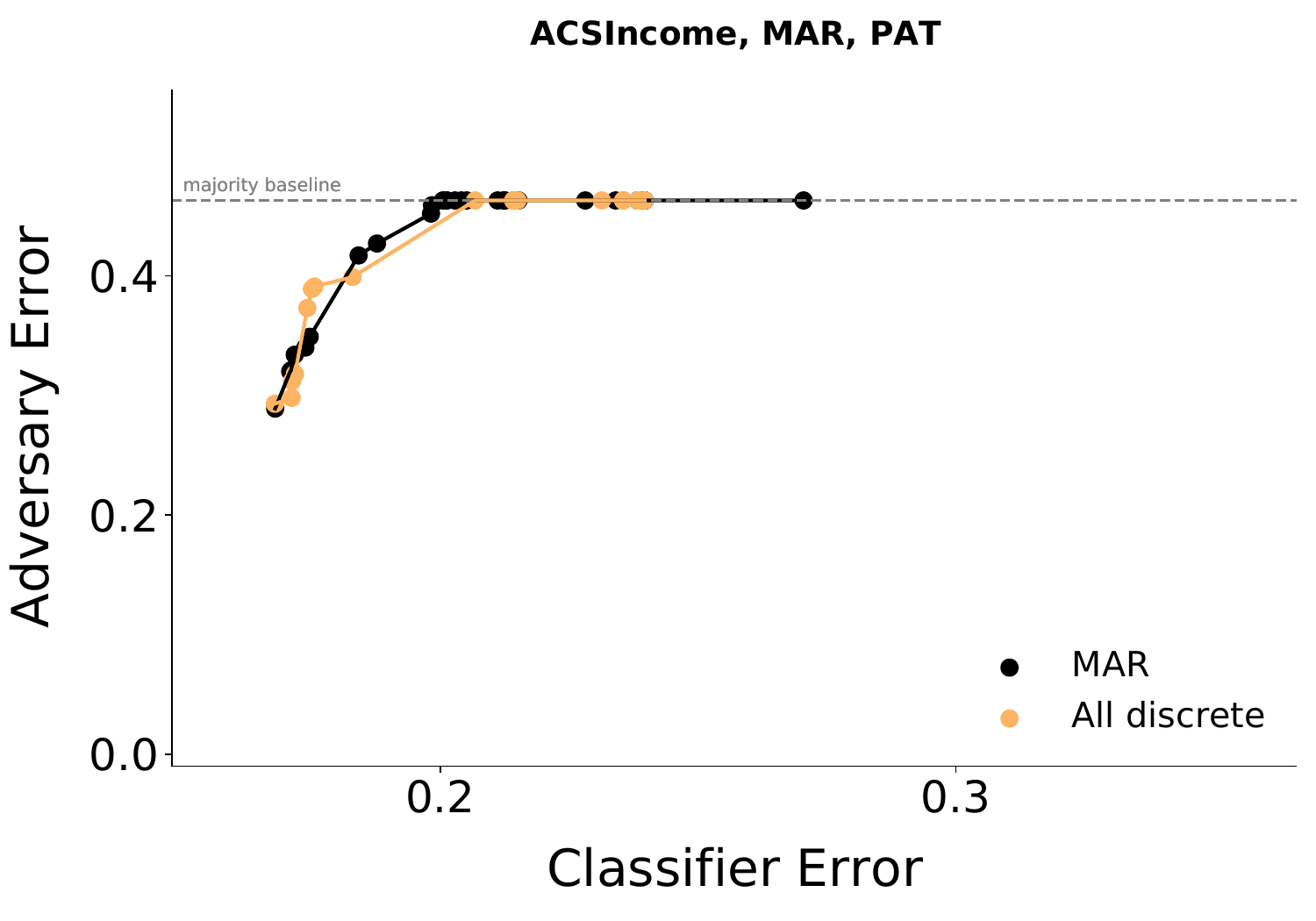} \\
  \includegraphics[width=0.24\textwidth]{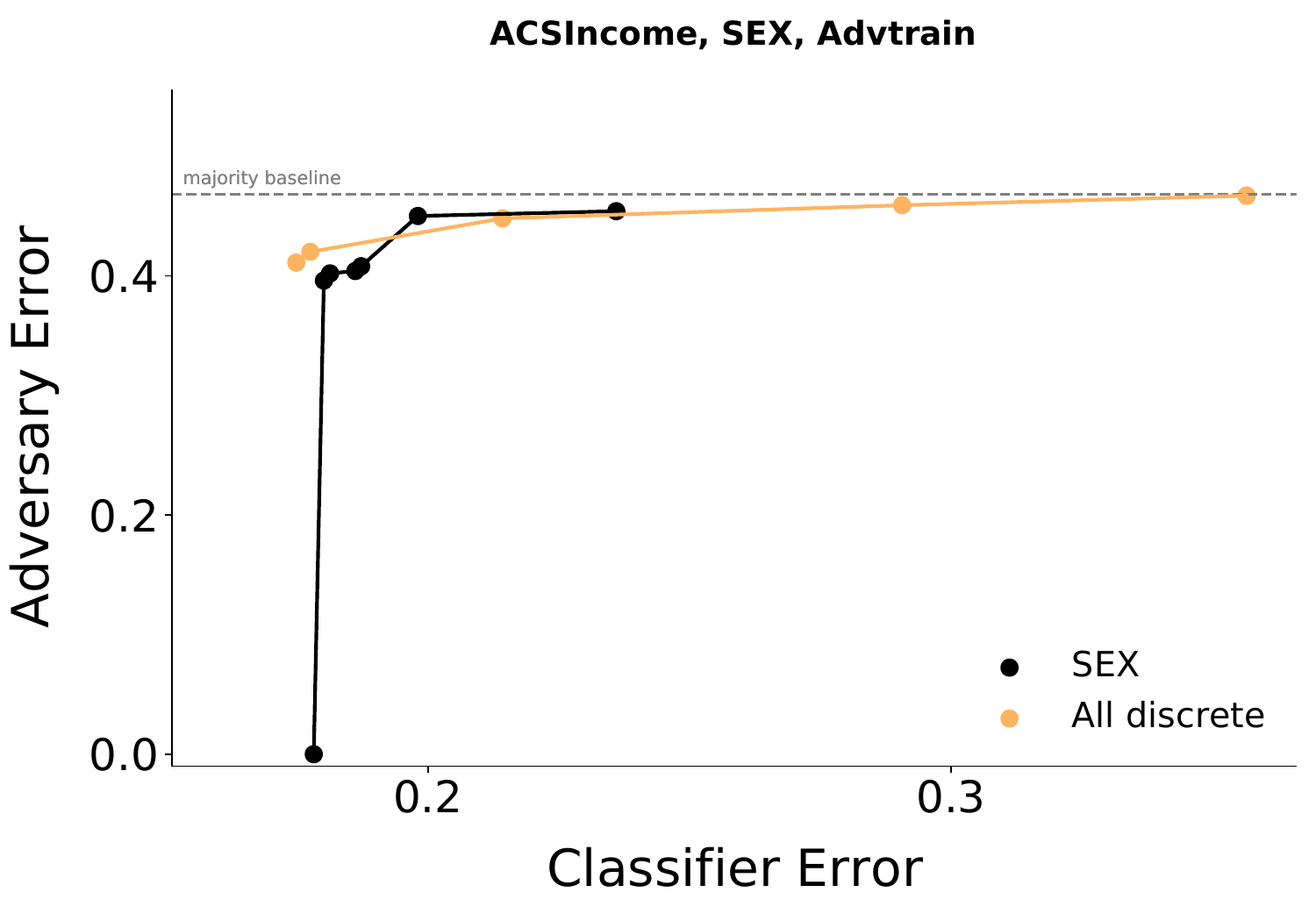} & \includegraphics[width=0.24\textwidth]{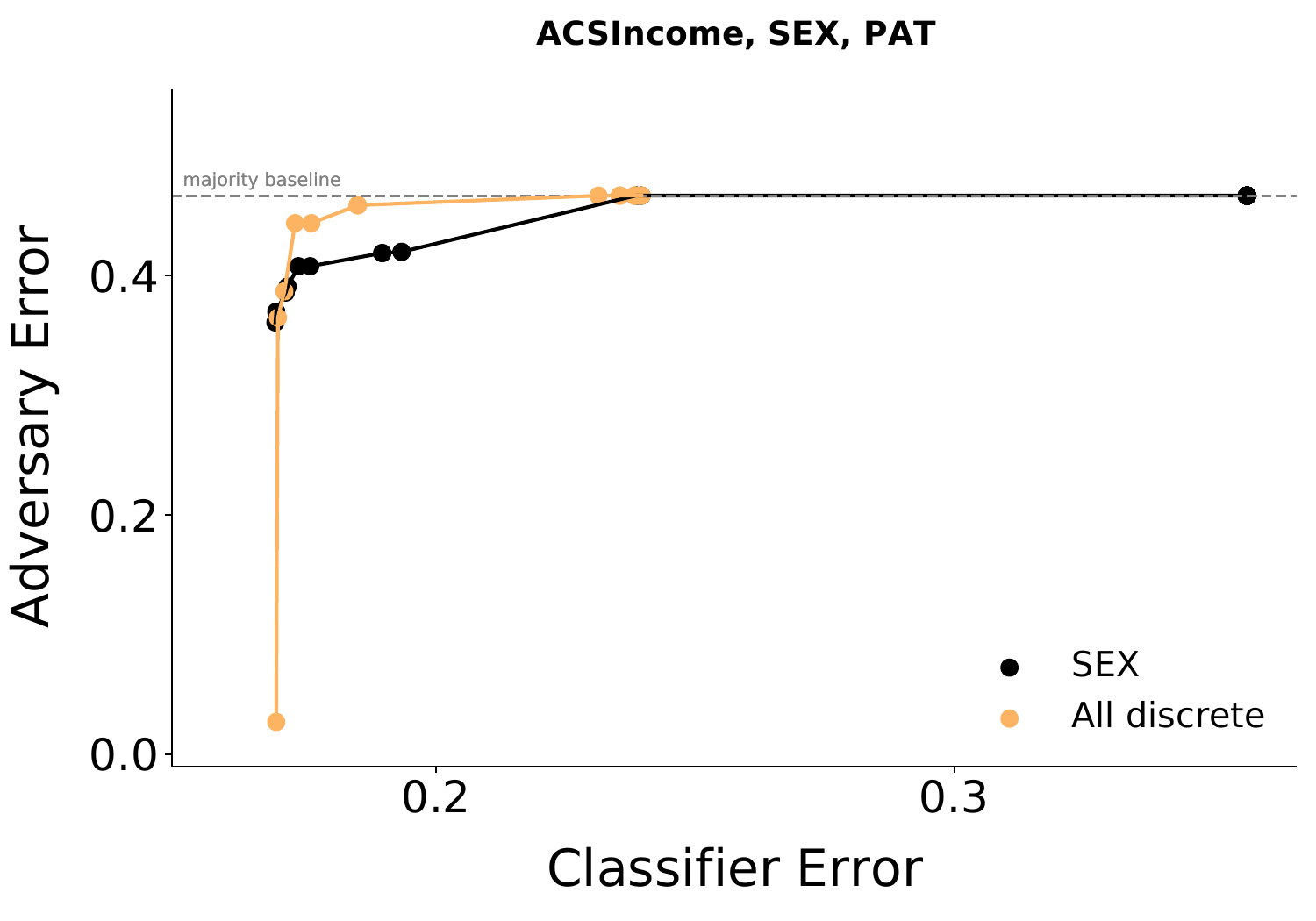}
  \end{tabular}
\caption{Reconstruction error of A1 adversary on a personal attribute (Married/SEX), when attacking generalizations from a minimizer (\emph{AdvTrain} or \emph{PAT}) trained to protect all discrete attributes (orange) or just the chosen attribute (black).}
\label{fig:abl_mean}
\vspace{-0.75em}
\end{figure}

\section{Protecting Several Personal Attributes} \label{app:meanvsmax}

One question that arises when designing minimizers is if protecting a large set of personal attributes is a good proxy for protecting a single personal attribute that may be of particular interest.
We investigate this with the following experiment, whose results are shown in~\cref{fig:abl_mean}. In each plot, we train a minimizer (\emph{AdvTrain} or \emph{PAT}) twice. 
One minimization is, as usual, trained to protect a larger set of personal attributes (orange line), while the other is trained to only protect a single attribute (\texttt{MAR} or \texttt{SEX} respectively). 
Both minimizations are then attacked by the A1 adversary, aiming to reconstruct only the single attribute.
As we can see in~\cref{fig:abl_mean}, the respective adversarial error curves are very close for both minimizations and target attributes. This gives a strong indication that using mean aggregation (over all personal attributes) when learning minimizers is a reasonable proxy for protecting specific personal attributes.

\section{Details Omitted from Experimental Evaluation} \label{app:details}

Here we supply all details omitted from \cref{sec:experimental}. 

\para{Detailed Descriptions of Minimizers} \label{app:details:minimizers}

We further formalize the baseline minimizers \emph{AdvTrain}, \emph{MutualInf}, and \emph{Iterative} (\cref{sec:baselines}).
For this we write the generalization $g_\psi: \mathcal{X} \to \mathcal{Z}$, the classifier $f_\theta: \mathcal{Z} \to \mathcal{Y}$ and the adversary $h_\phi: \mathcal{Z} \to \mathcal{X_S}$.
 
\para{Generalization $g_\psi$ as a Neural Network}
As shown in \cref{sec:baselines} the \emph{AdvTrain} and \emph{MutualInf} minimizers model the generalization $g_\psi$ as a set of $d$ independent neural networks $g_\psi^{(i)}$.
We will now give more detail on how $g_\psi$ is implemented both for discrete and continuous attributes. For this, let $\vx \in \mathcal{X}$ be an input $g_\psi$.

For discrete attributes $i$, let $g_\psi^{(i)}: \sR^{c_i} \rightarrow \sR^{k_i}$ be a network that receives a 1-hot encoding of the attribute value $x_i \in \{1, 2, ..., c_i\}$.
The output is the probability $p_j$ of generalizing $x_i$ to value $j \in \{1, 2, ..., k_i\}$ computed by applying the softmax with temperature $\tau$ to the unnormalized probabilities produced by $g_\psi^{(i)}$: $ p_j(x_i) = \frac{\exp(g_\psi^{(i)}(x_i)_j/\tau)}{\sum_{j'=1}^{k_i} \exp(g_\psi^{(i)}(x_i)_{j'}/\tau)}$.

For continuous attributes $i$, we assume them to be normalized to $[0,1]$. The network \mbox{$g_\psi^{(i)}: [0, 1] \rightarrow [0, 1]$} uses $W \odot W \geq 0$ as linear layer weights, together with tanh activations and batch normalization, to ensure a monotonic mapping.
Then, we divide $[0, 1]$ into $k_i$ equally-sized intervals and let $c_j = \frac{2j-1}{2k_i}$ be the center of the $j$-th interval.
We set the probability of generalizing $x_i$ to attribute value $j$ as: $p_j(x_i) = \frac{\exp(-(g_\psi^{(i)}(x_i)_j - c_j)^2/\tau)}{\sum_{j'=1}^{k_i} \exp(-(g_\psi^{(i)}(x_i)_{j'} - c_{j'})^2/\tau)}.$

During training, we progressively decrease the temperature $\tau$. Once the training is finished, we generalize each attribute to the value with the highest probability $p_j(x_i)$, essentially corresponding to the limit when $\tau \rightarrow 0$.
For \emph{AdvTrain} and \emph{MutualInf}, we set $k_i=k$ for all attributes while allowing the network to learn not to use some values.

\para{Minimization with \emph{MutualInf}}
The goal of the \emph{MutualInf} minimizer is to reduce the mutual information between the generalized $\vz$ and the original attributes $\vx$, written as $I(\vz, \vx) = H(\vz) - H(\vz | \vx)$.
Using the fact that each generalized attribute $z_i$ is computed independently using $x_i$ and applying Jensen's inequality, we can bound $H(z)$ via $\log P(\vz) = \log \E_x \left[ P(\vz \mid \vx) \right] = \log \E_x \left[ \prod_{i=1}^d P(z_i \mid x_i) \right] \geq \E_x \left[ \sum_{i=1}^d \log P(z_i \mid x_i) \right]$ and
$H(\vz) = \smallminus\E_\vz \left[ \log P(\vz) \right] \leq \smallminus\E_{\vz, \vx} \left[ \sum_{i=1}^d \log P(z_i \mid x_i) \right]$.

Similarly, for the conditional entropy, we can write:
\begin{align*}
  H(\vz|\vx) %
  &= -\E_\vx \E_{\vz \mid \vx} \left[ \sum_{i=1}^d \log P(z_i \mid x_i) \right].
\end{align*}

The upper bound for $H(\vz)$ can be approximated by independently sampling $\vz$ and $\vx$, while the conditional entropy can be approximated by first sampling $\vx$ and then $\vz$ conditionally on $\vx$.
Since we are using neural networks to model the generalization function \mbox{$P(z_i = j \mid x_i) = p_j(x_i)$}, where $p_j(x_i)$ depends on the parameters $\theta$ of the neural networks.
We can use this to jointly minimize the mutual information and the classification loss during training:
\begin{equation*}
  \min_{\theta, \psi} \E_{\vx, y} \left[ (1 - \lambda) {L}_{\text{clf}}(f_\theta(g_\psi(\vx)), y) + \lambda {L}_{\text{inf}}(g_\psi(\vx), \vx) \right],
\end{equation*}

where $\lambda$ is (as for \emph{AdvTrain}) a tradeoff factor between two optimization objectives, ${L}_{\text{clf}}$ denotes the classifier loss, and ${L}_{\text{inf}}$ denotes the mutual information objective.

\para{Minimization with \emph{Iterative}}
We now give more detail on the \emph{Iterative} minimizer which uses a heuristic procedure to generalize each attribute to a fixed number of buckets (equivalence classes), trying to improve the generalization while keeping the classification error below a threshold $T$.

Assume that the number of buckets $k$ is known for each attribute.
For discrete attributes, we fit a logistic regression $\langle w, x^{(oh)} \rangle + b$ predicting the target label, where $x^{(oh)}$ denotes the one-hot encoding of the training data $x$, and sort the array of possible values for $x_i$ w.r.t. the matching element of $w$, as a proxy for the impact of a value on classification (i.e., \emph{score}).
Intuitively, we want to map attribute values with similar scores to the same bucket.

We compute this mapping using dynamic programming with the state ($a$, $k'$). We aim to decide how to assign the first $a$ possible values for attribute $x_i$ into $k'$ groups such that the average of the variances of scores inside each group is as small as possible.
This is done by trying all possible values $b \leq a$ as starting positions of the last group and then taking $b$ to minimize the average variance of the solution that uses the group $[b,a]$ together with the solution used at $(b-1, k'-1)$. For continuous attributes, we directly split the range $[0,1]$ based on $k$-quantiles of the training set.

We now explain how the \emph{Iterative} minimizer determines the number of buckets $k$ for each attribute. We first sort all attributes by an estimate of the attribute's impact on the difference between the classification error and the adversarial error.
Let $\Delta_{\text{clf}}^{(i)}$ and $\Delta_{\text{adv}}^{(i)}$ respectively denote the increase in classification and adversarial error from a model which predicts using original attributes to the model which predicts without attribute $i$ (\ie it is fully generalized). We then sort all attributes by $\Delta_{\text{clf}}^{(i)} - \Delta_{\text{adv}}^{(i)}$ in increasing order and generalize the attributes sequentially.

We first set the number of buckets for each attribute to $k$ (a hyperparameter).
Then, we reduce the number of buckets as long as the classification error is below the threshold $T$ and then proceed to the next attribute.

\begin{figure*}[t]
  \centering
 \begin{tabular}{cccc}
  \includegraphics[width=0.22\textwidth]{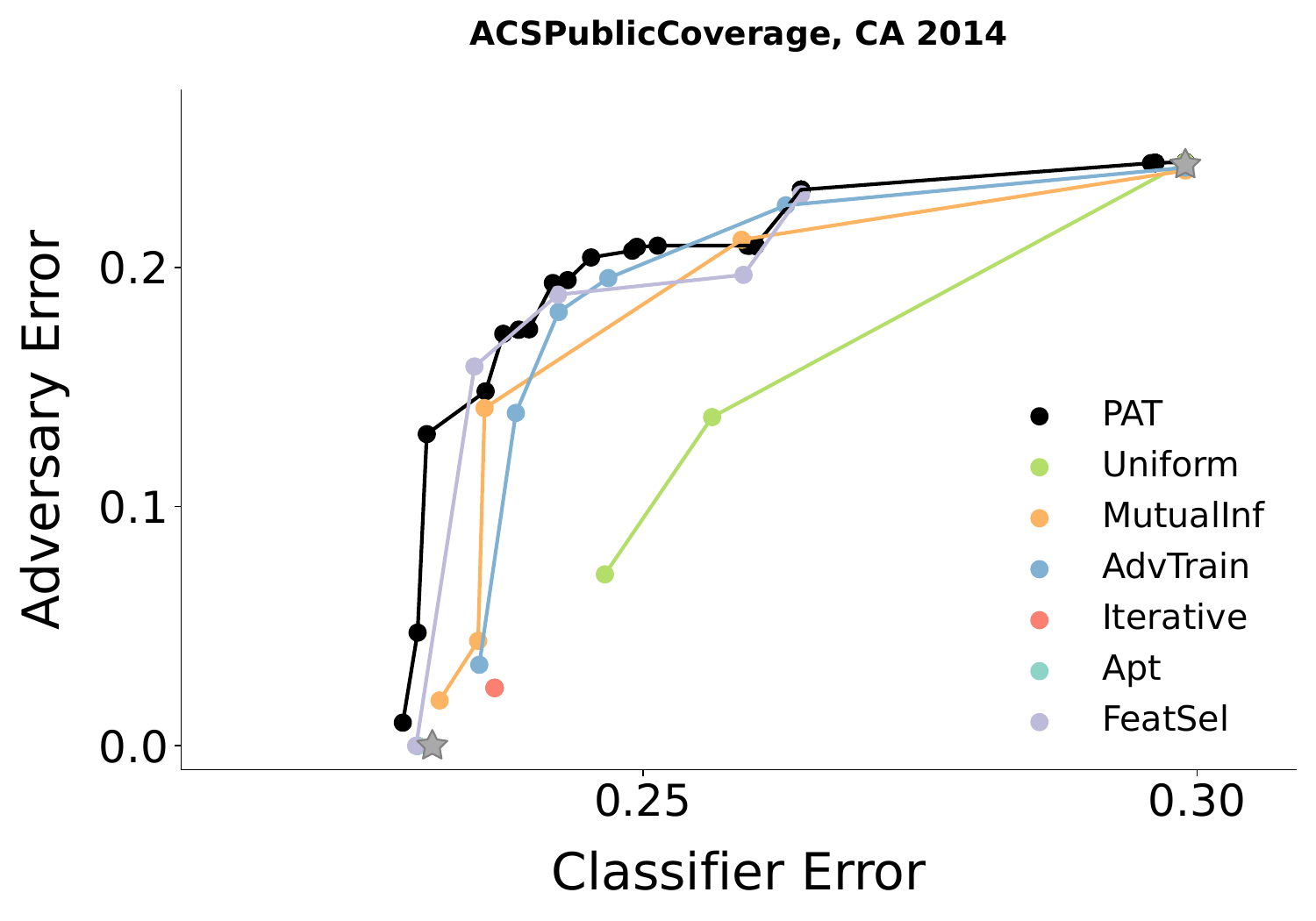} & 
  \includegraphics[width=0.22\textwidth]{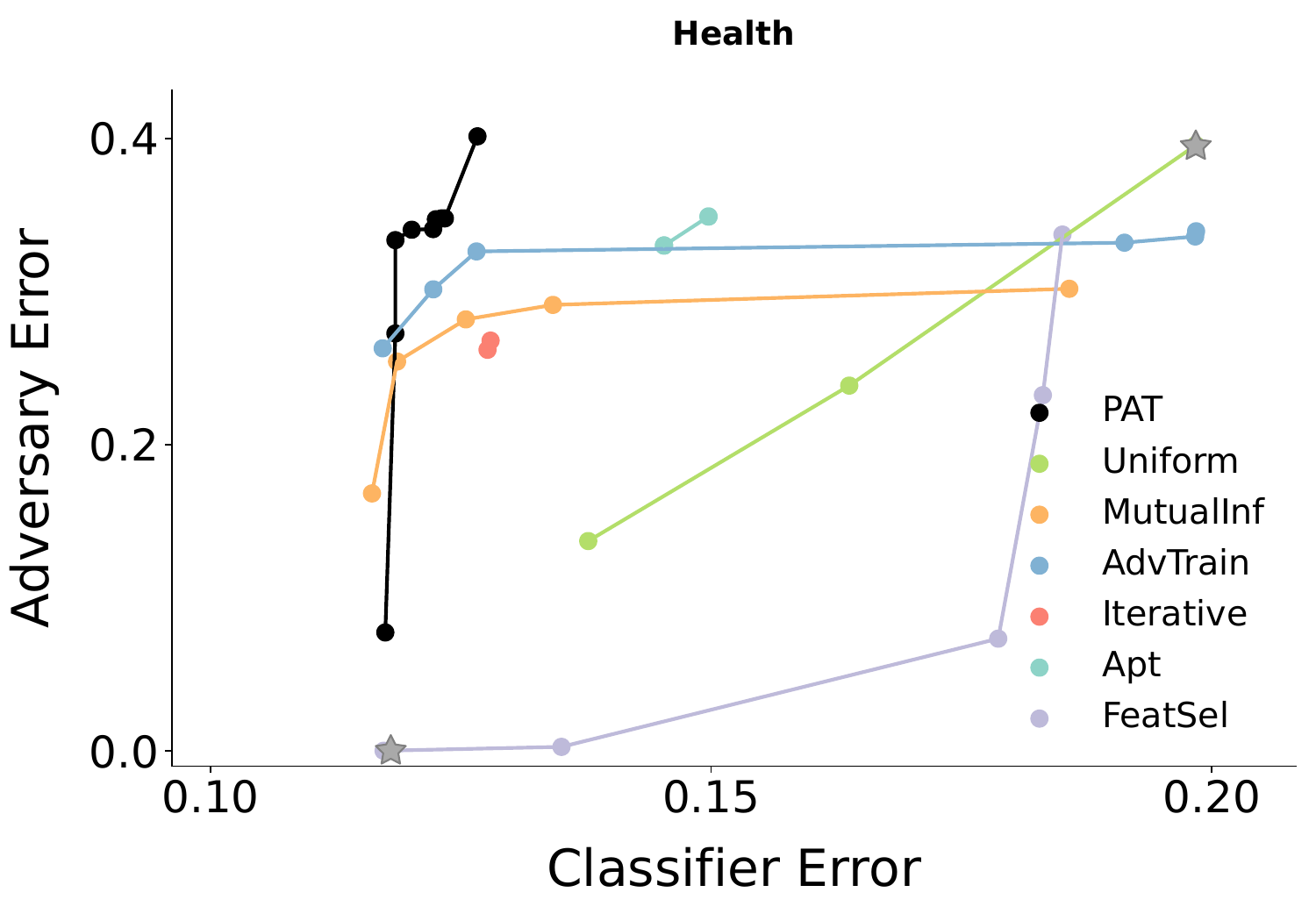} & 
  \includegraphics[width=0.22\textwidth]{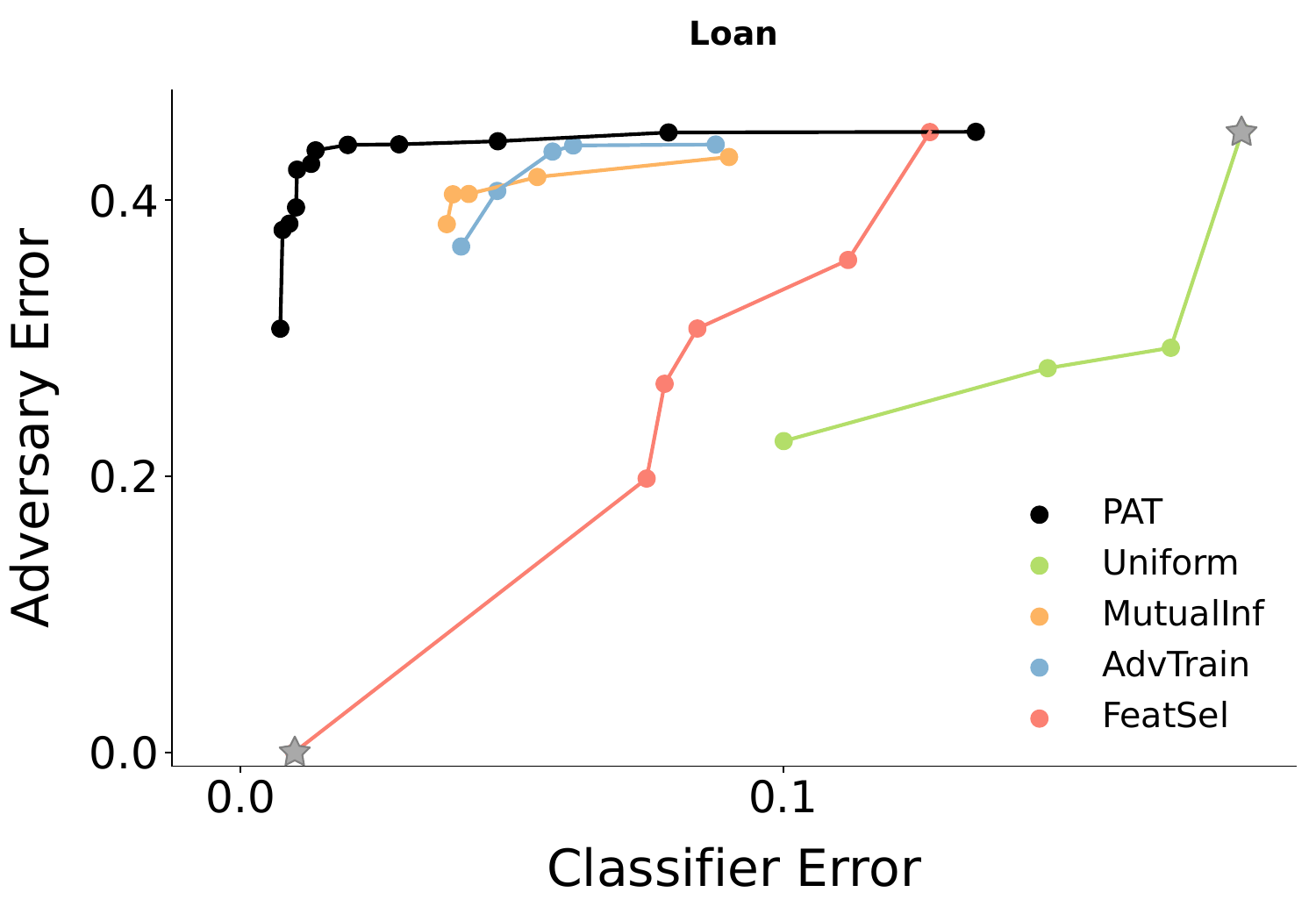} &  
  \includegraphics[width=0.22\textwidth]{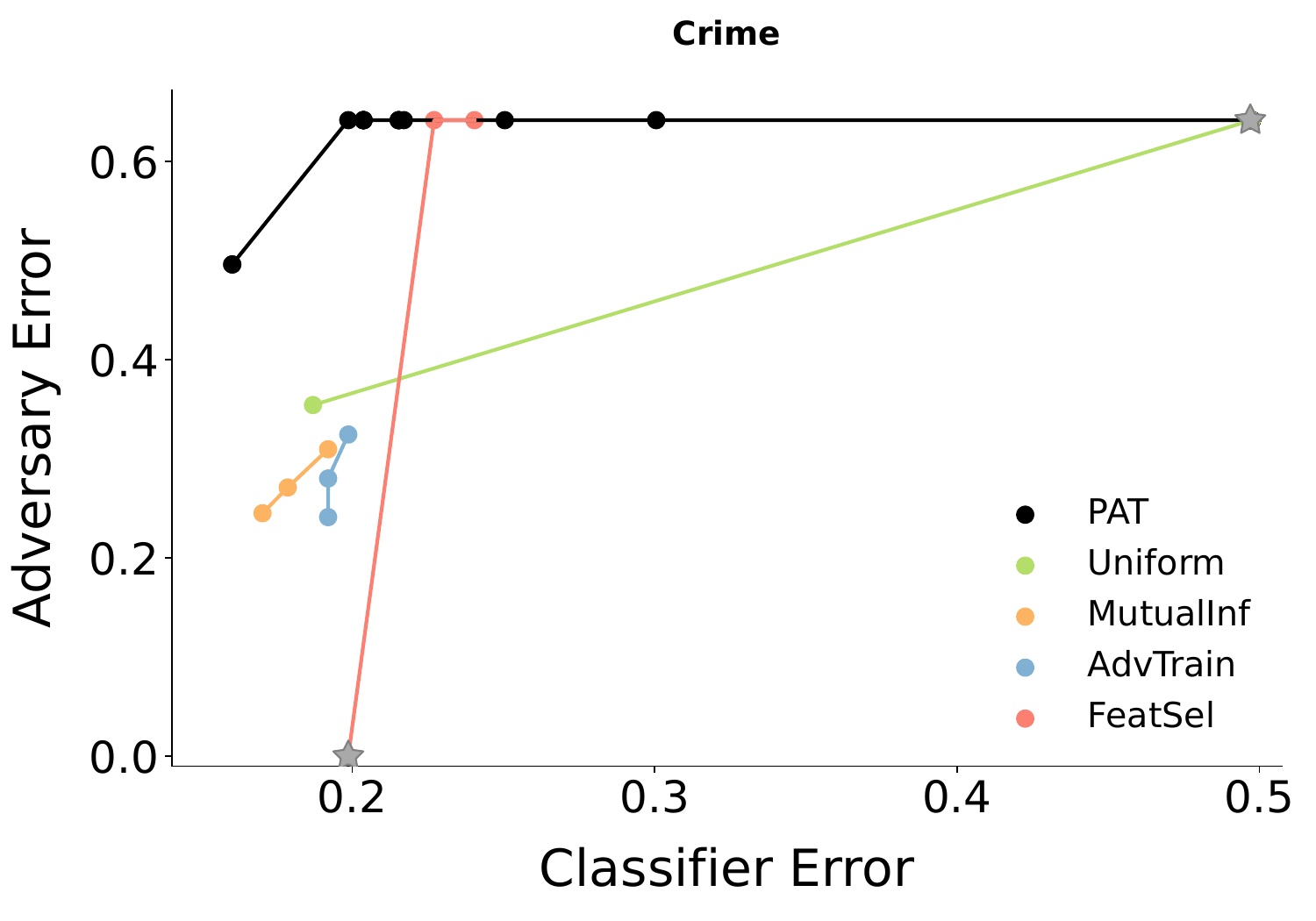}
 \end{tabular}
  \caption{Utility-privacy tradeoffs of candidate generalizations produced by minimizers on ACSPublicCoverage, Health, Loan and UCI Crime. Classifier and adversarial errors are reported on a held-out test set and selected on the validation set.}
  \label{fig:add_plots}
  \vspace{-0.5em}
\end{figure*}
 
\section{Dataset Details} \label{app:details:dataset}

\para{ACS}
In this work, we use several datasets and prediction tasks from the ACS suite, recently proposed by \citet{RetiringAdult}. The suite is derived from the American Community Survey (ACS) data, released by the US Census Bureau. All tasks used in \cref{sec:experimental} are described below. Each dataset offers slices by US state and year, which we fix to California (CA) and 2014, respectively. If not specified otherwise, we set all discrete attributes (see Appendix B in \citet{RetiringAdult} for a detailed list of all attributes) as personal.  

\emph{ACSEmployment}: The task is to predict if an adult is employed. There are 372,553 datapoints for CA in 2014 having 16 attributes (14 personal) each.

\emph{ACSIncome}: The task is to predict if the yearly income of a person is above \$50,000. There are 183,941 datapoints for CA in 2014 having 10 attributes (7 personal) each.

\emph{ACSPublicCoverage}: The task is to predict if a low-income individual not eligible for Medicare has public health insurance coverage. There are 152,676 datapoints  for CA in 2014 having 19 attributes (16 personal) each.

\para{Health} We further evaluate on the Heritage Health Dataset first proposed in \citet{noauthor_heritage_nodate}, trying to predict the Charlson Comorbidity Index of hospital patients. The health dataset contains 218,415 datapoints. We preprocess the health dataset as described in \citet{balunovic2022fair}, selecting two versions:

\emph{Health}: Contains all 101 attributes out of which we select the following as personal: PCG=CANCRM, PCG=COPD, PCG=METAB3, PCG=PRGNCY, Specialty=Internal, PG=EM, PG=SCS, PlaceSvc=Office.

\emph{Health, pruned}: We subsample the set of attributes to 
DrugCount\_total, DrugCount\_months, no\_Claims, no\_Providers, PayDelay\_total, PCG=COPD, PCG=METAB3, Specialty=Internal, PG=EM, PG=SCS, PlaceSvc=Office, AGE>60.
Out of these we consider the last 6 as personal.

\para{Loan} We additionally use the Loan dataset~\cite{loan}, an excerpt from the Lending Club loan data from 2015, as previously used by \citet{IbmApt}. Here we use 42 attributes of persons, with the goal of predicting loan status. We use all categorical attributes (term, grade, sub\_grade, emp\_length, home\_ownership, verification\_status, pymnt\_plan, purpose, initial\_list\_status, application\_type, hardship\_flag, disbursement\_method, issue\_d, addr\_state) as the personal attributes.

\para{UCI Crime} Finally, we use the Communities and Crime dataset from the UCI repository~\cite{uci}, which combines US socio-economic, law enforcement and crime data from several sources. It contains 128 attributes for each community, and the goal is to predict if the number of violent crimes per capita is above or below the median. We follow the preprocessing from \citet{balunovic2022fair}, resulting in 99 attributes of which we consider (pre-processed) race and state as personal.

\section{Additional Results} \label{app:details:moreresults}
In \cref{fig:add_plots} we show additional results of our main experiment, extending the ones given in \cref{fig:acs}.
We explore several new datasets (described in \cref{app:details:dataset}): ACSPublicCoverage from the ACS suite, full Health with 101 attributes, Loan and UCI Crime. 
In all cases we note the same results---PAT is the best minimizer overall, preserving utility while reducing the privacy risk. 

\begin{figure}[t]
  \centering
  \includegraphics[width=0.37\textwidth]{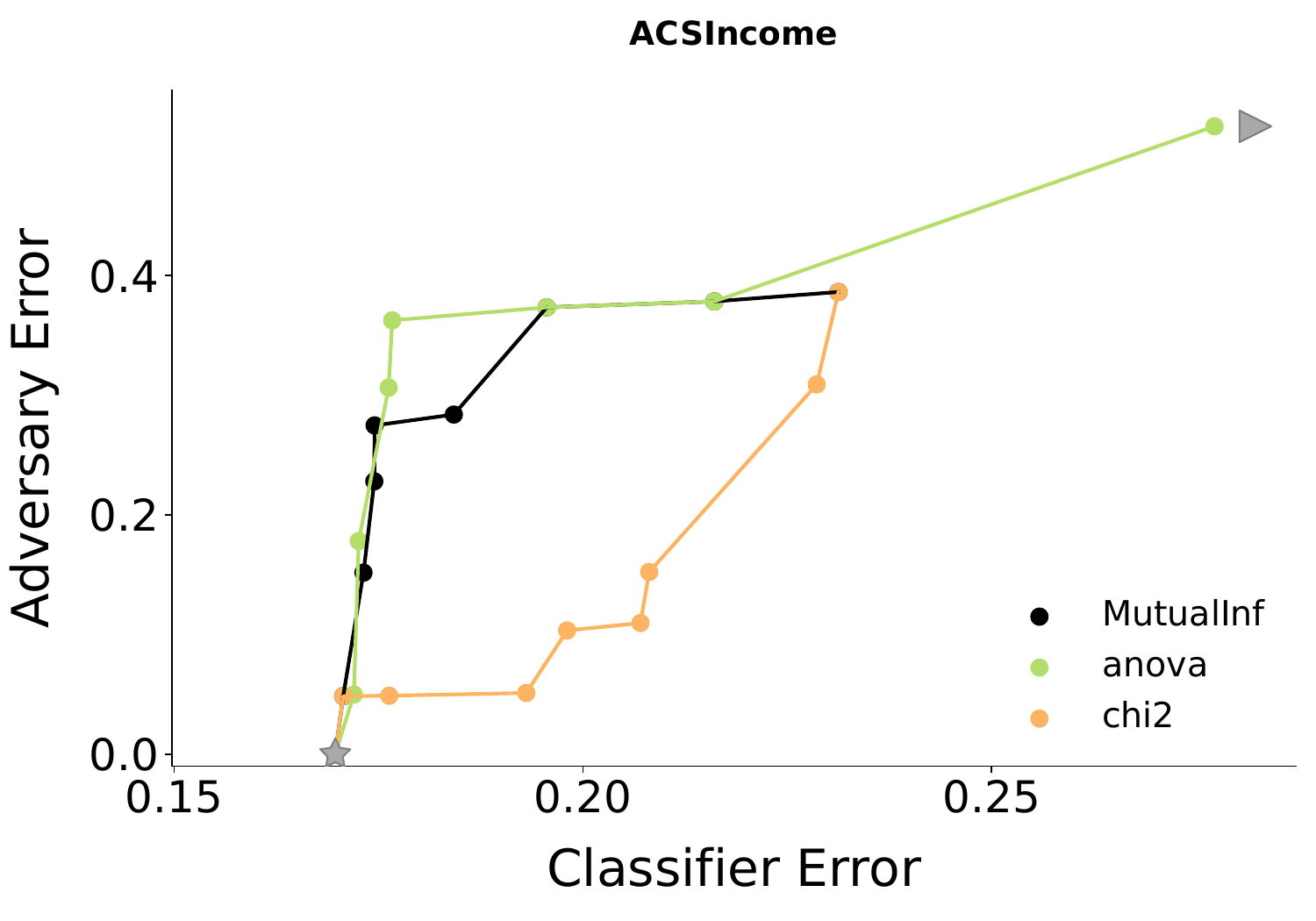}
  \caption{Comparison of feature selection methods on ACSIncome.}
  \label{fig:featsel}
  \vspace{-0.5em}
\end{figure} 
\begin{figure}[t]
  \centering
 \begin{tabular}{cc}
  \includegraphics[width=0.22\textwidth]{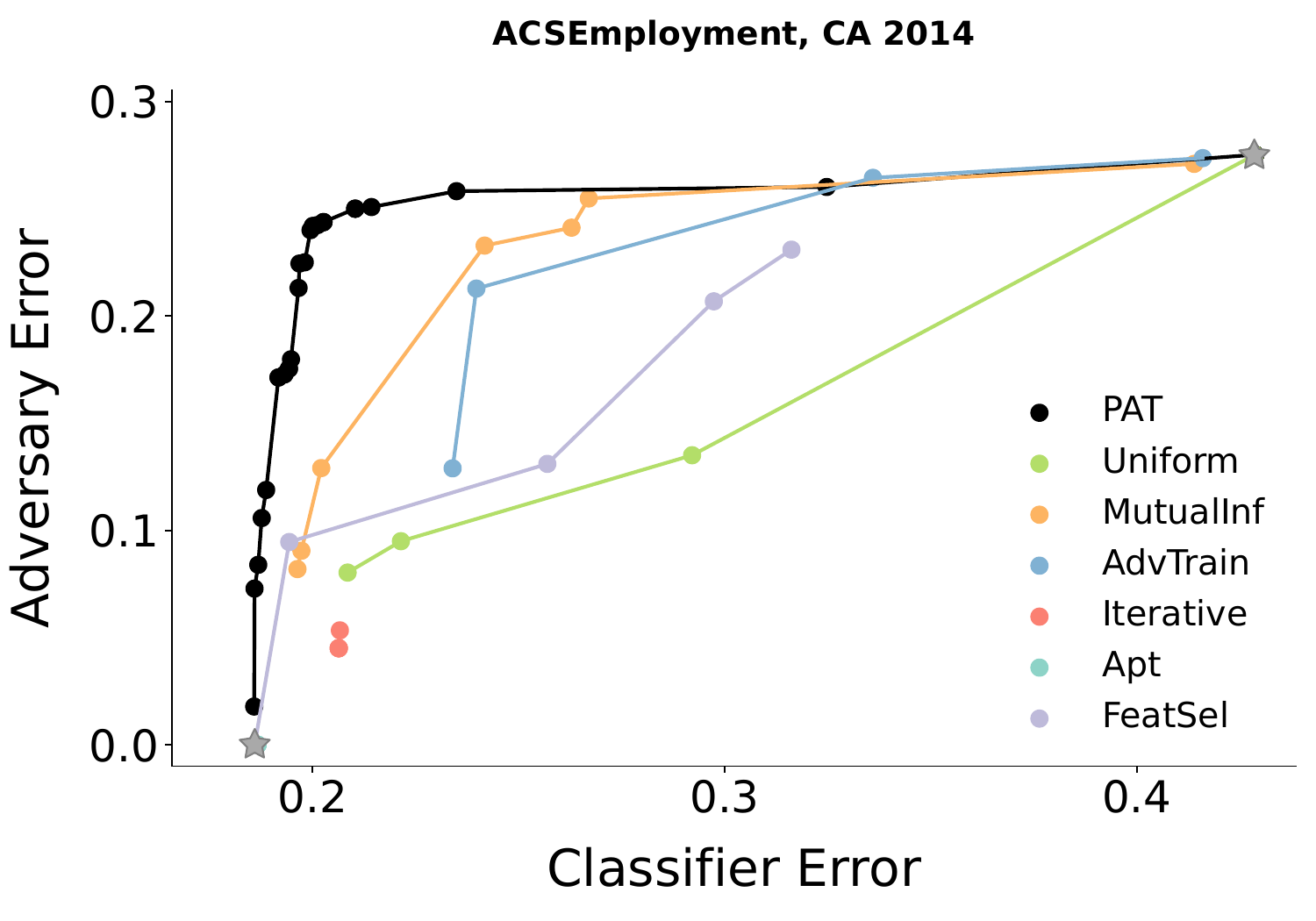} & \includegraphics[width=0.22\textwidth]{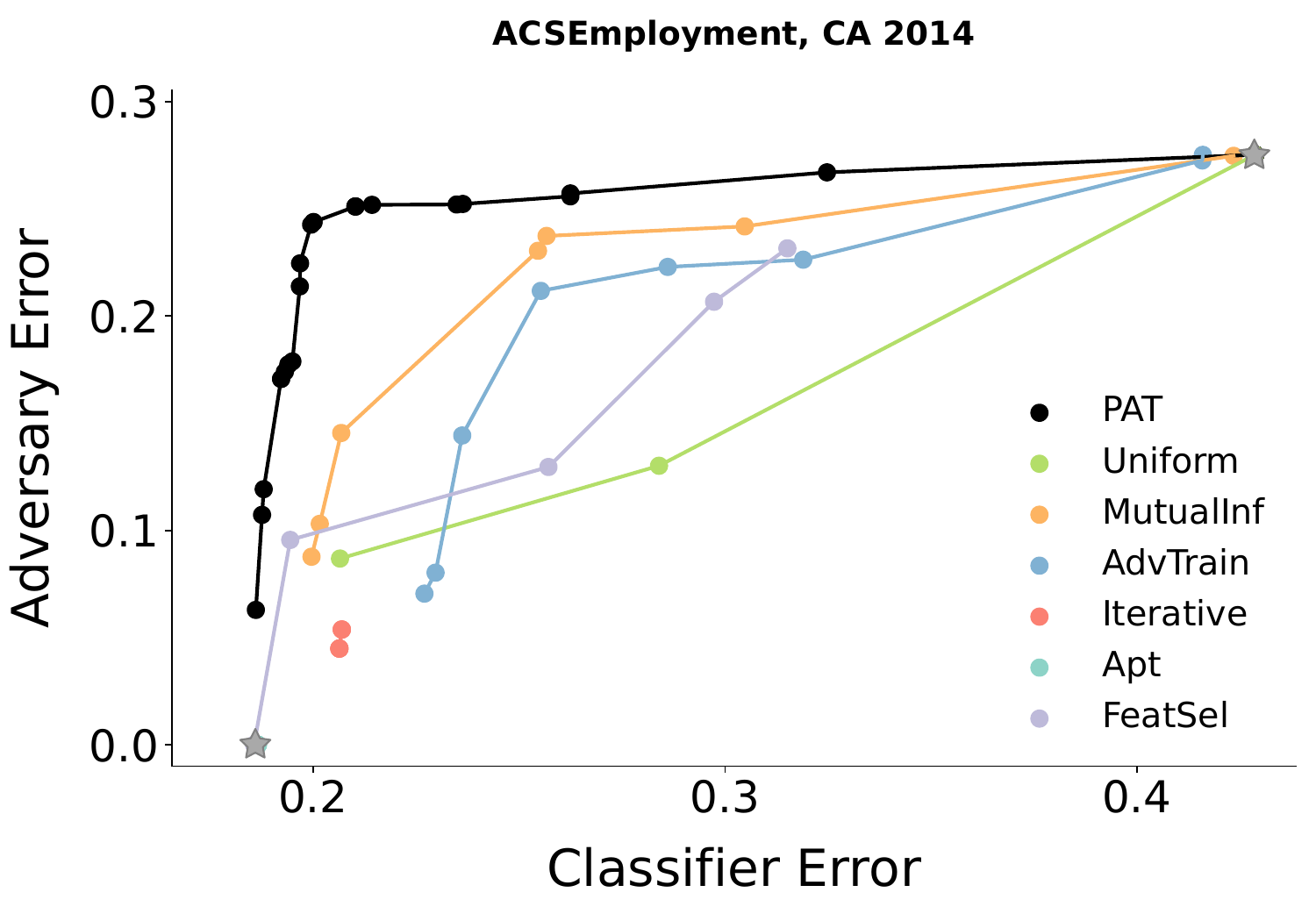}
 \end{tabular}
  \caption{Main runs with two different seeds on ACSEmployment. We observe similar behavior across all seeds.}
  \label{fig:seeds}
  \vspace{-0.8em}
\end{figure}

In~\cref{fig:featsel} we explore additional feature selection methods based on mutual information and $\chi^2$ values, and observe that they perform worse on ACSIncome compared to the ANOVA method we used in our main experiment as a feature selection baseline.
We additionally experimented with a variant of feature selection where we iteratively remove $1$ feature at a time (until $k$ remain), but observed that in all cases this resulted in the same final feature set. 

\section{Experimental Parameters} \label{app:details:params}
In all experiments, we use $60\%$ of the data as the training set (used to fit the minimizers), 10\% as held-out validation (used to select the Pareto front of generalizations and internal validation), and $30\%$ as the test set (used for final results). We repeat runs with 5 seeds observing the same behavior across all main experiments, as we show in \cref{fig:seeds}.

As \toolshort~requires each class to be in the training set at least once, we always (\ie for all minimizers) first select a set of samples that ensures this, filling the training set with randomly drawn from the remaining dataset.
The batch size is set to $256$. We model the classifier and the adversary as neural networks with one hidden layer of width $50$ and ReLU activations. For evaluation, classifier and adversary are trained for $20$ epochs with learning rate $0.01$ multiplied by $0.1$ after $10$ epochs. 
The value of the $L2$ regularization parameter is automatically tuned on the validation. 
Our framework performs several runs with various parameters before using the validation set to choose the Pareto front.

For the \emph{uniform baseline}, the only varying parameter is $k$. We perform runs for $k \in \{1,2,3,4,5\}$. For \emph{univariate feature selection}, we try all values $k \in \{2,4,8,10,20,all\}$.

We run \emph{Apt} and the \emph{Iterative} minimizer with $7$-$11$ values of the target classifier error $T$, chosen for each dataset to be between the two limit points (no and full generalization). 

For the \emph{Iterative} minimizer, we set the initial number of buckets $k=4$, as we did not observe further improvement with more buckets. For both \emph{Iterative} and \emph{Apt} we set a time-limit of $2h$ per run, $~5\times$ the next slowest minimizer.

For the neural minimizers (\emph{AdvTrain} and \emph{MutualInf}), we perform runs with $9$ values of $\lambda \in [0, 1]$, fixing the number of buckets to $k=5$. 
We use the same classifier/adversary architectures as used in the evaluation and perform $20$ epochs of training with starting learning rate of $0.01$, multiplied by $0.1$ every $5$ epochs. 
We use no L2 regularization and schedule the softmax temperature from $2.0$ to $0.5$ over the course of training. For \emph{AdvTrain} we set $N_{\text{inner}}=1$.

For PAT, we ran experiments with the maximum number of leaves $k^* \in \{2, 4, 6, 8, 10, 20, 50, 100, 200\}$, requiring that at least $100$ samples end up in every leaf. For each run, we selected the same 12 possible values for $\alpha$ in the range $[0,1]$. PAT required roughly $3$ seconds per run, significantly outperforming all other minimizers in terms of speed.

\section{Architectures \& Individual Attributes} \label{app:architectures}
\begin{figure}[t]
    \centering
    \includegraphics[width=0.37\textwidth]{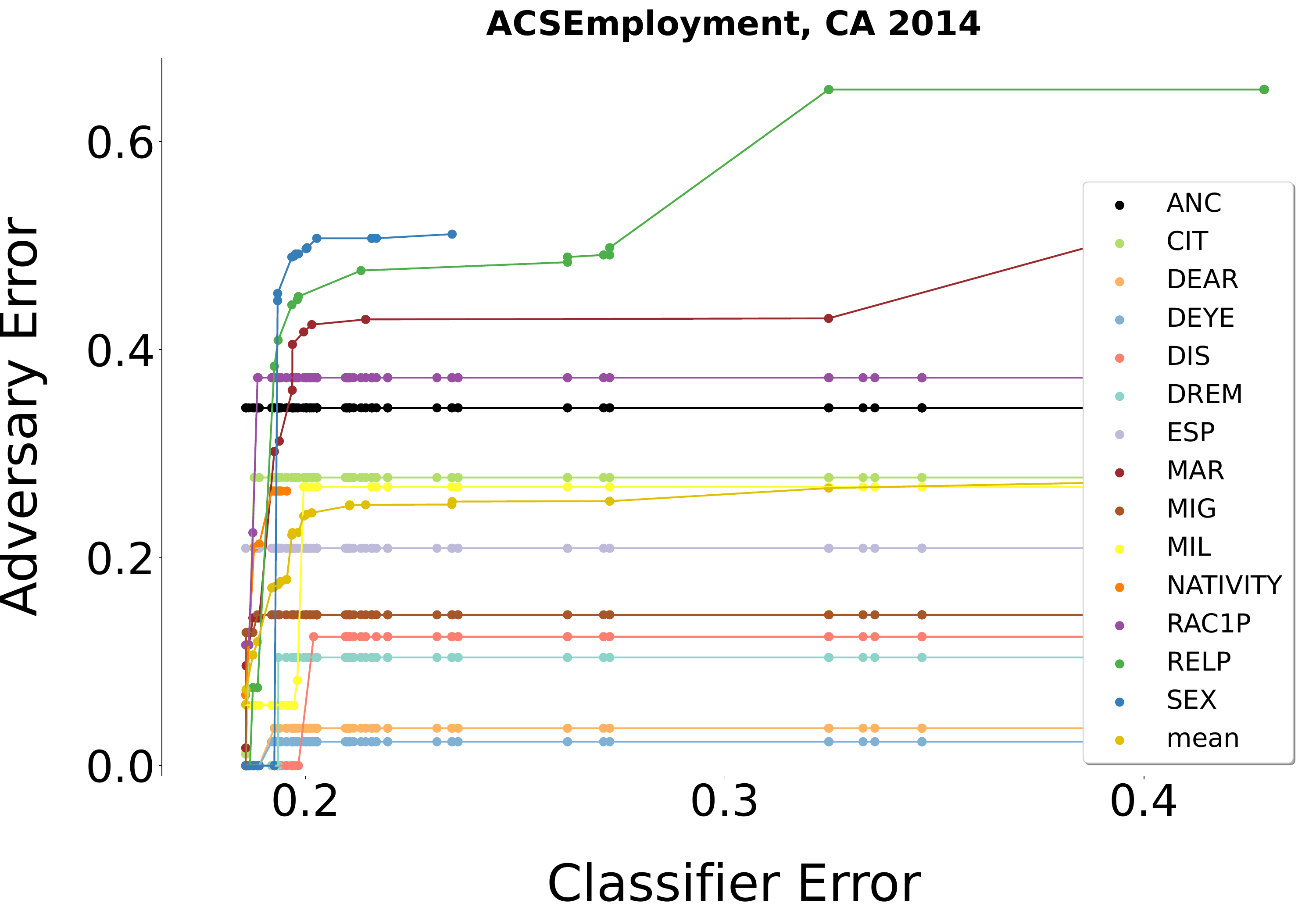}
    \caption{Utility-privacy tradeoff on ACSEmployment. For points on the Pareto front in \cref{sec:experimental} we report the adversarial error per attribute.}
    \label{fig:individual_plots}
    \vspace{-0.8em}
  \end{figure}

We also perform experiments on the architecture of our A1 adversary and downstream classifier. In particular, for the ACSIncome dataset, we investigate how the adversarial architecture used in \cref{sec:experimental} (referred to as MLP-2) is sufficient for the reconstruction of personal attributes. For this, we additionally introduce MLP-3, a 3 layer neural network with an intermediate layer width of 100 neurons, as well as MLP-5, a 5 layer neural network with an intermediate width of 256 neurons.
\cref{Tab:architectures} shows the adversarial reconstruction accuracy using the aforementioned architectures for the possible minimizer choices AdvTrain, Uniform, and PAT. We find that for AdvTrain and Uniform, using a slightly larger architecture can sometimes improve the reconstruction accuracy, while for PAT, MLP-2 consistently performs the best. This indicates that the results for PAT in \cref{sec:experimental} are reasonably close to the best achievable adversarial error rate.

\begin{table}[t]
    \centering
    \tabcolsep=0.05cm
    \caption{Classifier and adversary error for different choices of minimizers and adversarial network architectures. We highlighted cases where the error is at least $2\%$ smaller than on the baseline MLP-2 network.
    }
    \begin{tabular}{
      S[table-format=1.2]|
      S[table-format=1.2]|
      S[table-format=1.2] 
      S[table-format=1.2]|
      S[table-format=1.2]
      S[table-format=1.2]|
      S[table-format=1.2]
      S[table-format=1.2]
      S[table-format=1.2]
    }
    {\bfseries\splitcell{}}
     & {\bfseries\splitcell{Arch.}}
     & {\small\splitcell{Unif. \\ $k{=}3$\\}}
     & {\small\splitcell{Unif. \\ $k{=}5$\\}}
     & {\small\splitcell{Adv. \\ $\alpha {=} 0$\\}}
     & {\small\splitcell{Adv. \\ $\alpha {=} 0.5$\\}}
     & {\small\splitcell{PAT \\ $k^* {=} 4$ \\ $\alpha {=} 0.7$}}
     & {\small\splitcell{PAT \\ $k^* {=} 10$ \\ $\alpha {=} 0.7$}}
     & {\small\splitcell{PAT \\ $k^* {=} 50$ \\ $\alpha {=} 0.7$}}\\
    \midrule
    {\bfseries\splitcell{Classifier \\ Error}}
    & {\splitcell{MLP-2 \\ MLP-3 \\ MLP-5}}
    & {\splitcell{0.36 \\ 0.36 \\ 0.36}}
    & {\splitcell{0.22 \\ 0.21 \\ 0.21}}
    & {\splitcell{0.17 \\ 0.17 \\ 0.18}}
    & {\splitcell{0.22 \\ 0.22 \\ 0.22}}
    & {\splitcell{0.2 \\ 0.2 \\ 0.2}}
    & {\splitcell{0.19 \\ 0.19 \\ 0.19}}
    & {\splitcell{0.17 \\ 0.18 \\ 0.18}}\\
    \midrule
    {\bfseries\splitcell{Adversary \\ Error}}
    & {\splitcell{MLP-2 \\ MLP-3 \\ MLP-5}}
    & {\splitcell{0.52 \\ 0.52 \\ 0.52}}
    & {\splitcell{0.31 \\ \textbf{0.24} \\ \textbf{0.24}}}
    & {\splitcell{0.39 \\ \textbf{0.3} \\ \textbf{0.31}}}
    & {\splitcell{0.48 \\ 0.5 \\ \textbf{0.43}}}
    & {\splitcell{0.47 \\ 0.47 \\ 0.5}}
    & {\splitcell{0.46 \\ 0.46 \\ 0.47}}
    & {\splitcell{0.36 \\ 0.36 \\ 0.37}}\\
    \bottomrule
    \end{tabular}
    \label{Tab:architectures}
    \vspace{-0.5em}
\end{table}

\begin{figure}[t]
  \centering
  \includegraphics[width=0.37\textwidth]{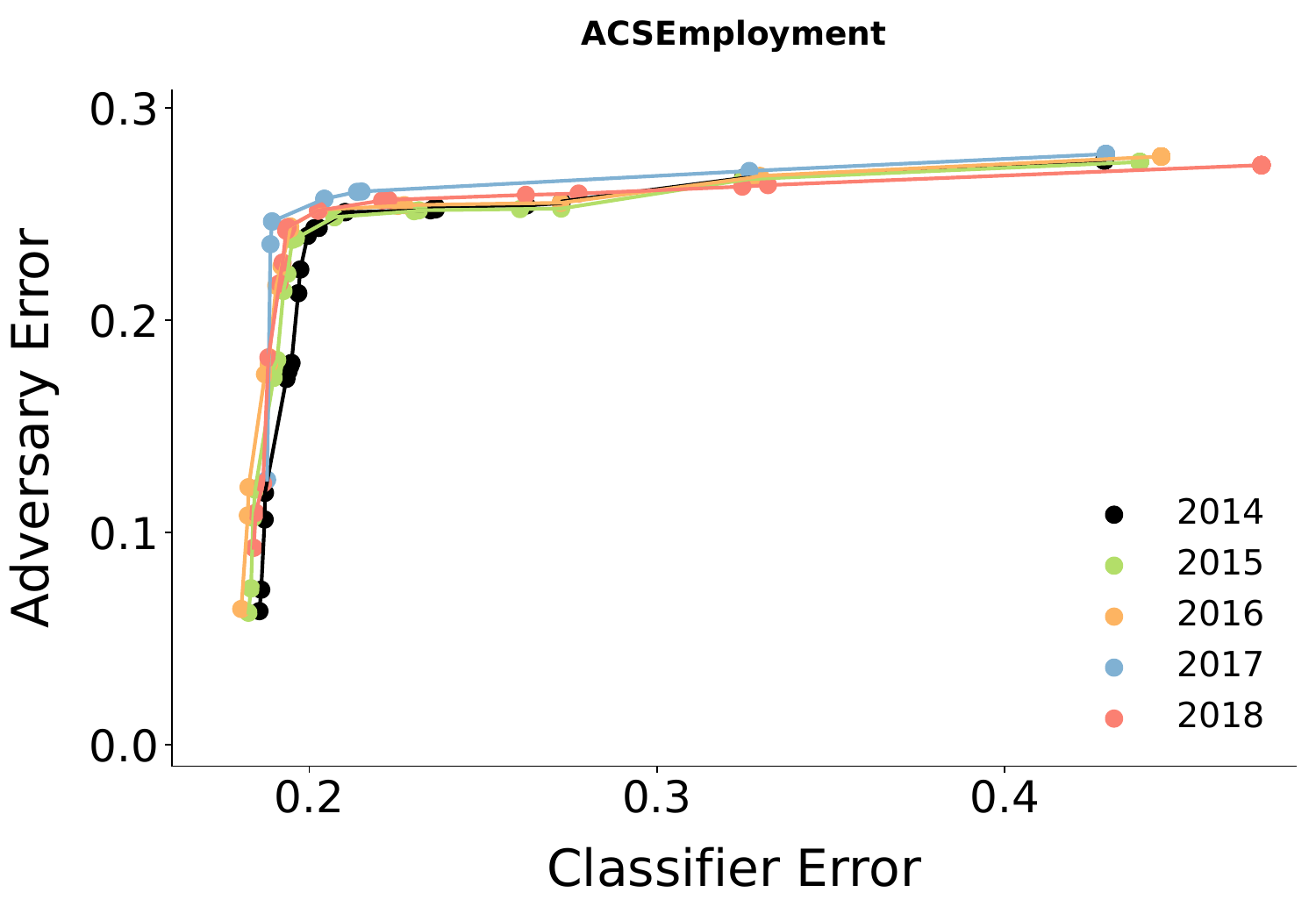}
  \caption{Robustness of PAT to distribution shift.}
  \label{fig:distshift}
  \vspace{-0.8em}
\end{figure}

Extending the experiment from~\cref{ssec:experimental:individual}, we present another individual attribute plot in~\cref{fig:individual_plots} for the ACSEmployment dataset. As ACSEmployment has more personal attributes (14) compared to ACSIncome,~\cref{fig:individual_plots} becomes harder to interpret, however when focussing on individual attributes, we can observe the same trends as in~\cref{ssec:experimental:individual}. In particular, we again find for all attributes strong inflection points, after which the adversarial error decreases drastically.

\section{Robustness to Distribution Shift} \label{app:distshift}

We test the robustness of PAT to temporal distribution shifts in the data. 
We fix the generalizations obtained on ACSEmployment data from 2014 (as used in our main experiments) and evaluate them under data from 2015, 2016, 2017 and 2018. 
The results are shown in~\cref{fig:distshift}---we notice no significant degradation. 
We further remark that such an evaluation can be used in practice to monitor model drift, and when significant drift is detected a new small dataset of updated data can be collected to retrain the minimizer.

\clearpage

\renewcommand{\thesubsection}{subsection}

\clearpage
 
\newpage %
\section{Meta-Review}

The following meta-review was prepared by the program committee for the 2024
IEEE Symposium on Security and Privacy (S\&P) as part of the review process as
detailed in the call for papers.

\subsection{Summary}
This paper proposes a data minimization workflow for machine learning tasks, focusing on collecting only the essential features and reducing the resolution of feature samples (e.g., buckets) to minimize privacy loss during potential data breaches. The authors assess its performance in eight adversarial contexts across five datasets.

\subsection{Scientific Contributions}
\begin{itemize}
\item Provides a Valuable Step Forward in an Established Field.
\end{itemize}

\subsection{Reasons for Acceptance}
\begin{enumerate}
\item This paper provides a valuable step forward in an established field. Developing machine learning models that adhere to data minimization principles has historically been a challenge. This research underscores the practical viability of data minimization methods.
\end{enumerate}

\subsection{Noteworthy Concerns} %
\begin{enumerate} %
\item The authors didn't provide a theoretical analysis detailing the precise level of privacy assurance the proposed method can achieve within a given utility error boundary.
\end{enumerate}

\section{Response to the Meta-Review} %

The meta-review notes that the work does not provide a theoretical analysis of the privacy-utility tradeoffs. As we point out in our future work section, obtaining a non-trivial and fully general theoretical result for vDM is a hard problem, with no clear solution within the scope of this paper. We agree that this can be a valuable extension of our work and encourage future efforts in this direction. We hope that the extensive vDM foundation set up by our work in terms of setting formalization, empirical risk estimation, baselines and PAT, can aid researchers in such follow-ups.

\end{document}